\documentclass[10pt]{article} 
\usepackage[preprint]{tmlr}

\usepackage{amsmath,amsfonts,bm}

\usepackage{tikz}
\usetikzlibrary{arrows,shapes,calc,matrix,fit,backgrounds}
\usepackage{pgfplots}
\usepackage{pgfplotstable}
\pgfplotsset{compat=1.13}
\usepgfplotslibrary{fillbetween,polar}
\usetikzlibrary{patterns}
\usepackage{pgfplotstable}
\usepackage{multirow}
\usepackage{amssymb}
\usepackage{adjustbox} 
\usepackage{enumitem}
\usepackage{booktabs}
\usepackage{capt-of}
\usepackage[table]{xcolor}
\usepackage{pifont}
\usepackage{ulem}

\usepackage{pgffor}
\usepackage{xintexpr}
\usepackage{graphicx}
\usepackage{soul}

\usepackage{url}
\definecolor{iccvblue}{rgb}{0.21,0.49,0.74}
\usepackage[
    pagebackref,
    breaklinks,
    colorlinks,
    citecolor=iccvblue
]{hyperref}
\usepackage{hyperref}
\usepackage[noabbrev,capitalize]{cleveref}

\definecolor{green}{HTML}{008015}

\definecolor{oursrow}{HTML}{E6F2F4}

\newcommand{\oursall}{ILGen-ALL}
\newcommand{\oursplus}{\oursall}
\newcommand{\oursgeneric}{ILGen-G}
\newcommand{\oursspecific}{ILGen-S}

\newcommand{\ie}{i.\,e.\ }
\newcommand{\eg}{e.\,g.\ }
\newcommand{\vs}{v.\,s.\ }

\def\month{07}  
\def\year{2026} 
\def\openreview{\url{https://openreview.net/forum?id=T3JgJXH3ZK}} 

\title{Instance-Level Generation for Representation Learning}

\author{\name Yankun Wu \email yankun@is.ids.osaka-u.ac.jp \\
      \addr The University of Osaka
      \AND
      \name Zakaria Laskar \email zakaria.laskar@iisertvm.ac.in \\
      \addr School of Data Science, IISER Thiruvananthapuram
      \AND
      \name Giorgos Kordopatis-Zilos \email kordogeo@fel.cvut.cz\\
      \addr VRG, FEE, Czech Technical University in Prague
      \AND
      \name Noa Garcia \email noagarcia@ids.osaka-u.ac.jp \\
      \addr The University of Osaka
      \AND
      \name Giorgos Tolias \email toliageo@fel.cvut.cz\\
      \addr VRG, FEE, Czech Technical University in Prague 
      }

\begin{document}

\maketitle

\begin{center}
\vspace{-15pt}
    \footnotesize
\begin{tabular}
{@{\hspace{3pt}}c@{\hspace{3pt}}c@{\hspace{3pt}}c@{\hspace{3pt}}c@{\hspace{10pt}}c@{\hspace{3pt}}c@{\hspace{3pt}}c@{\hspace{3pt}}c}
        generated object & foreground &  sample A &  sample B & generated object & foreground &  sample A &  sample B \\[5pt]        
    \includegraphics[width=50pt,height=50pt]{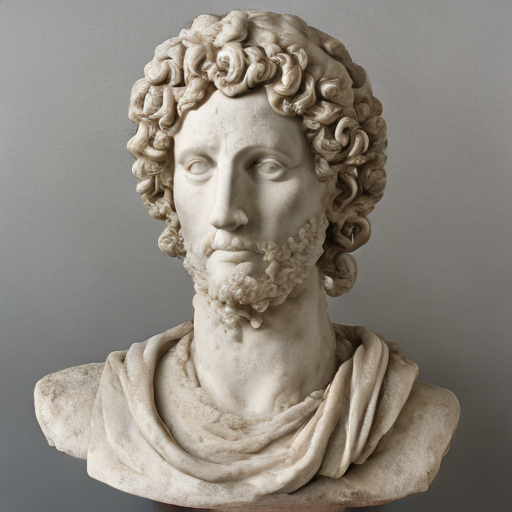} & 
    \includegraphics[width=50pt,height=50pt]{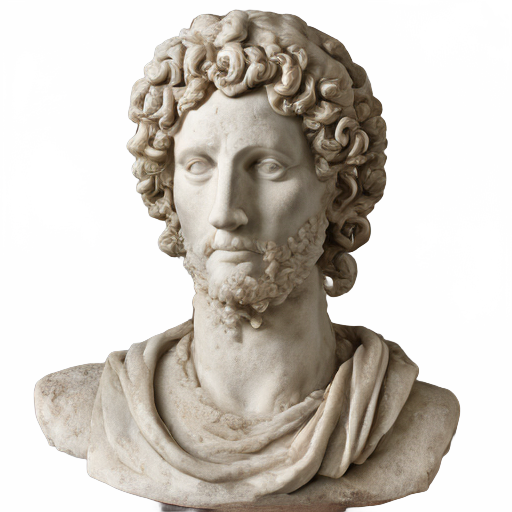} & 
    \includegraphics[width=50pt,height=50pt]{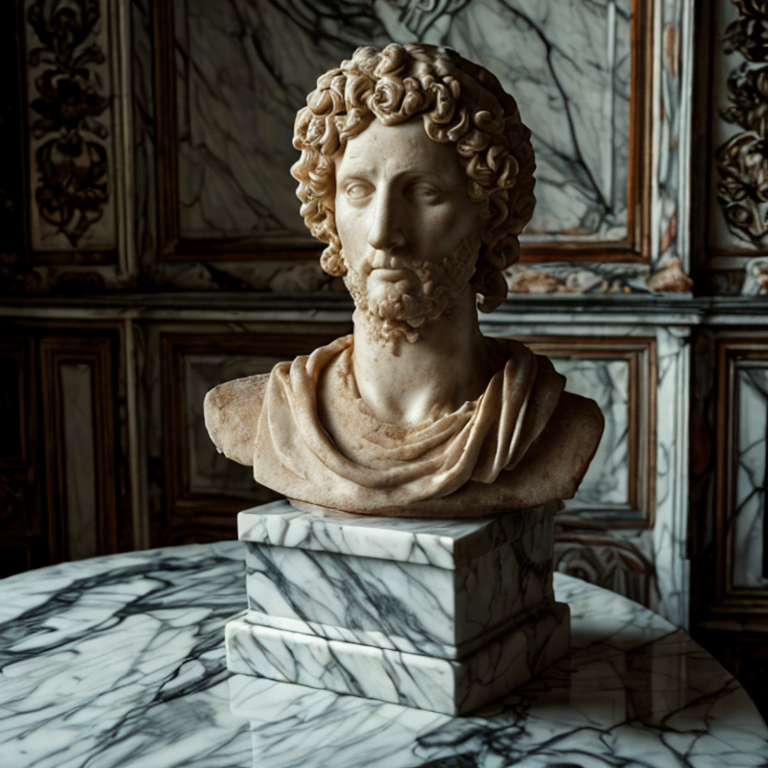} & 
    \includegraphics[width=50pt,height=50pt]{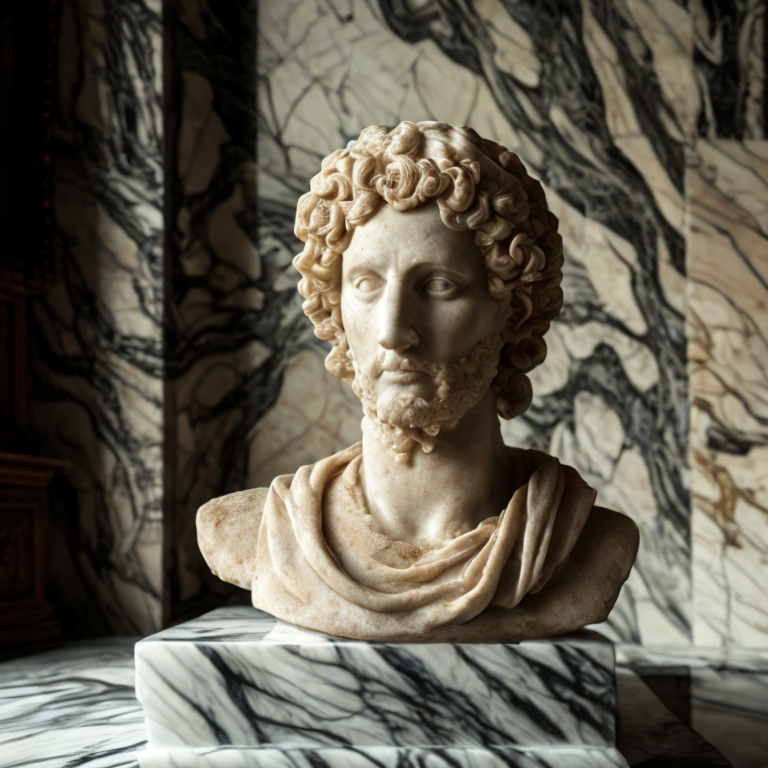} &
    \includegraphics[width=50pt,height=50pt]{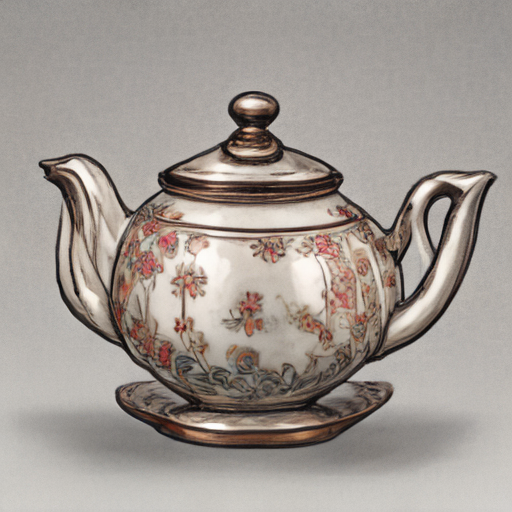} & 
    \includegraphics[width=50pt,height=50pt]{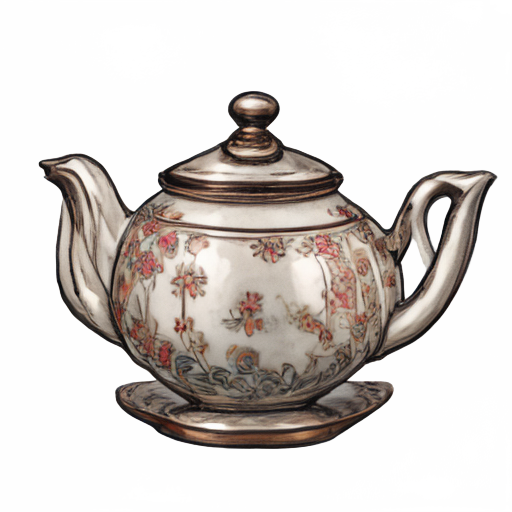} & 
    \includegraphics[width=50pt,height=50pt]{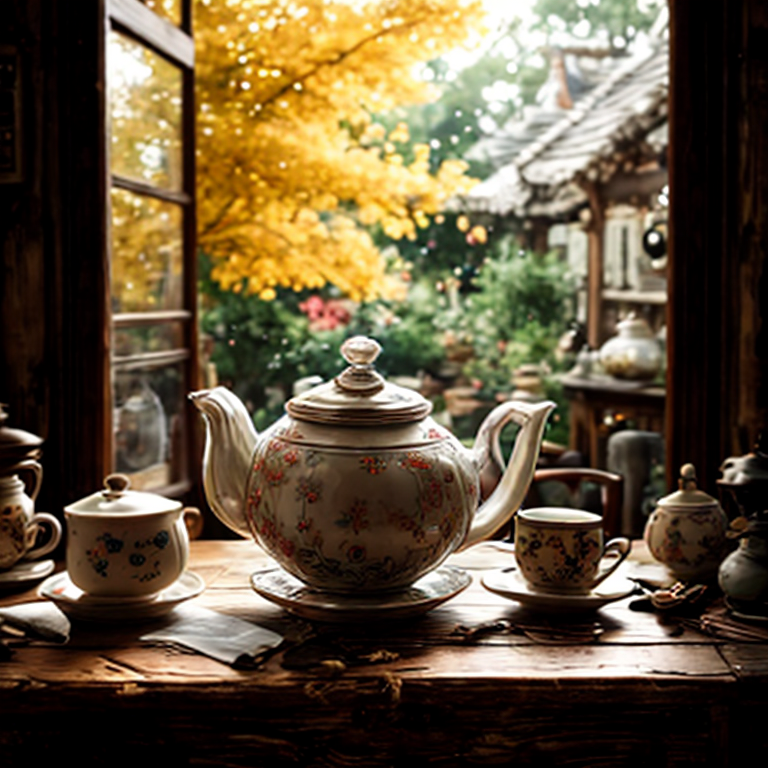} & 
    \includegraphics[width=50pt,height=50pt]{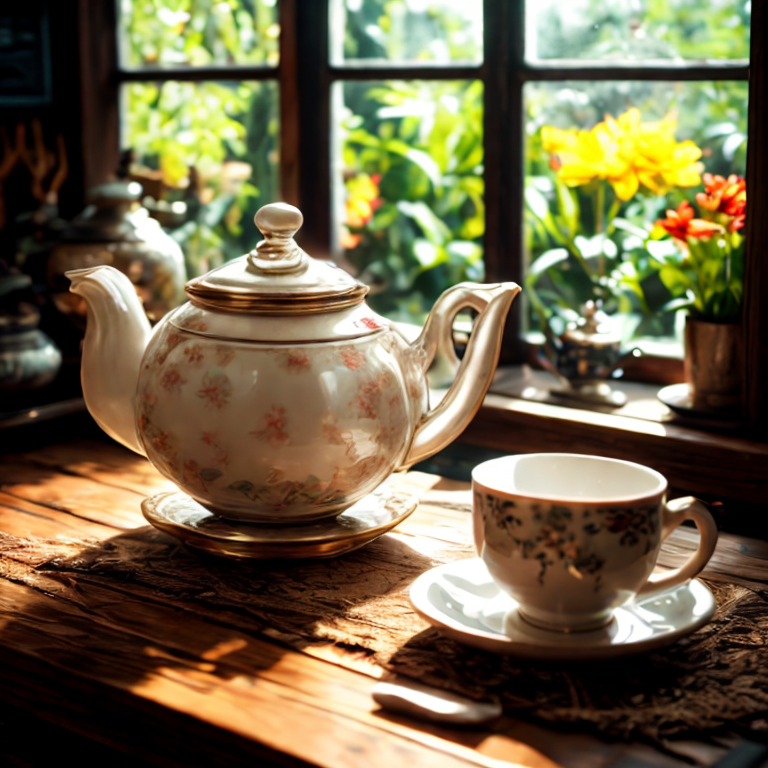}  \\[5pt]
    
    \includegraphics[width=50pt,height=50pt]{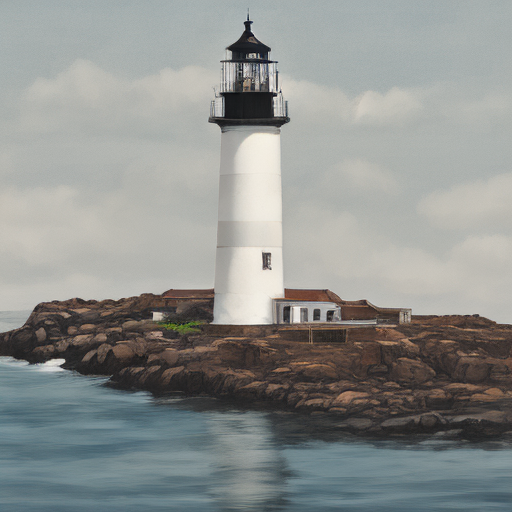} & 
    \includegraphics[width=50pt,height=50pt]{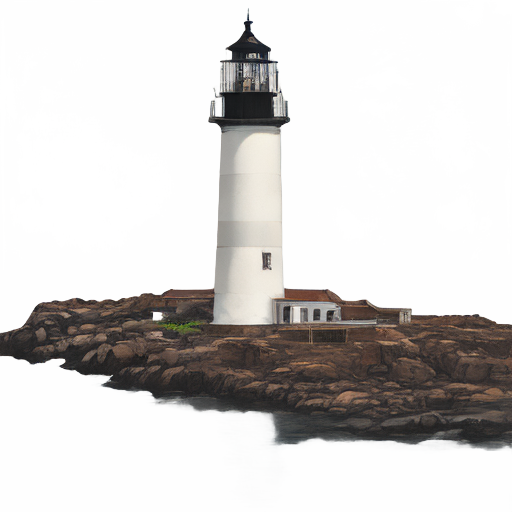} & 
    \includegraphics[width=50pt,height=50pt]{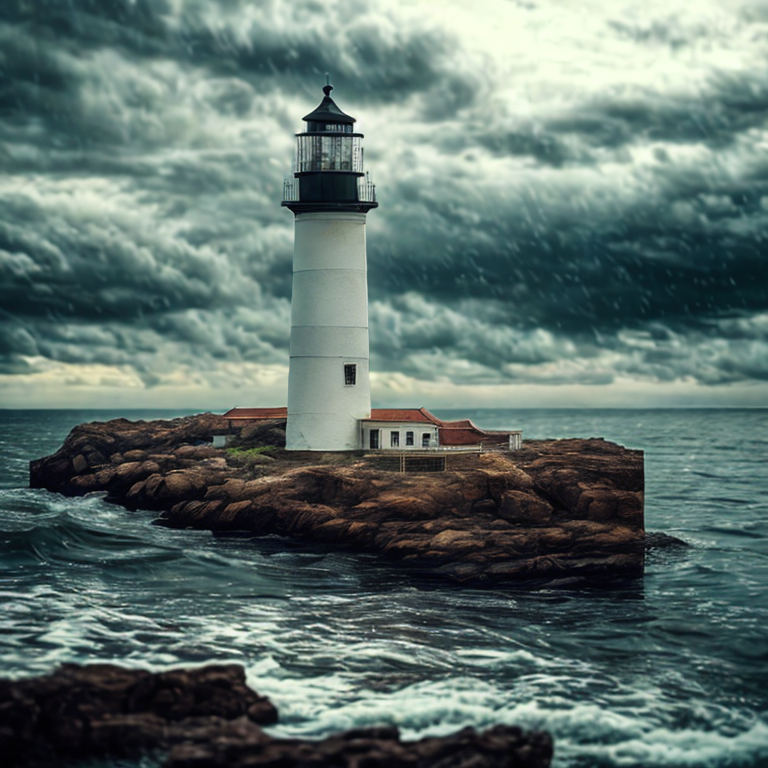} & 
    \includegraphics[width=50pt,height=50pt]{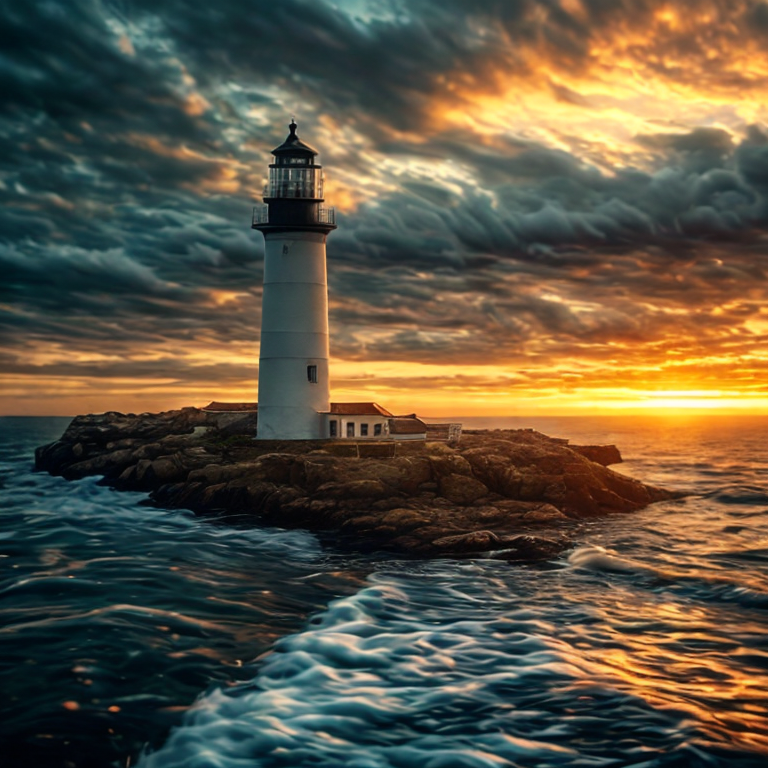} &
    \includegraphics[width=50pt,height=50pt]{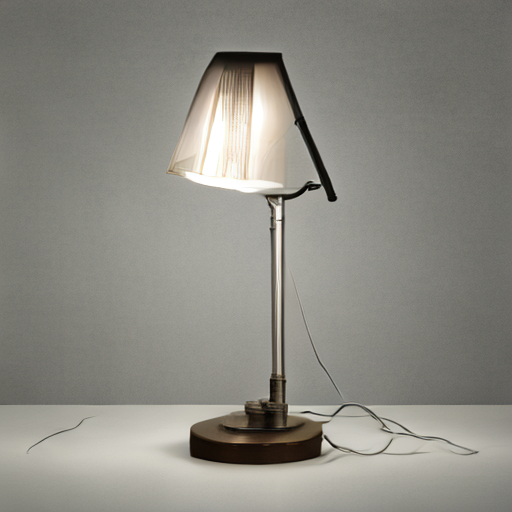} & 
    \includegraphics[width=50pt,height=50pt]{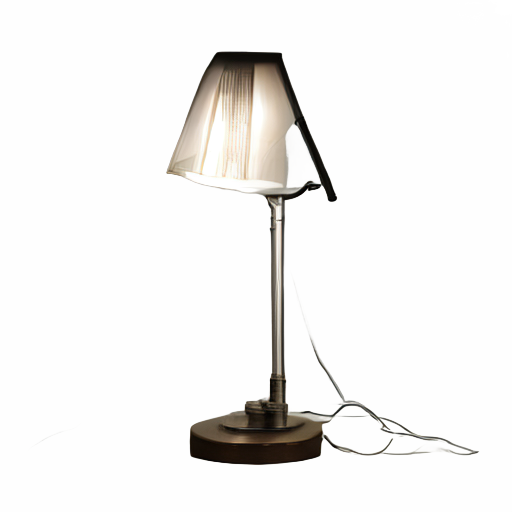} & 
    \includegraphics[width=50pt,height=50pt]{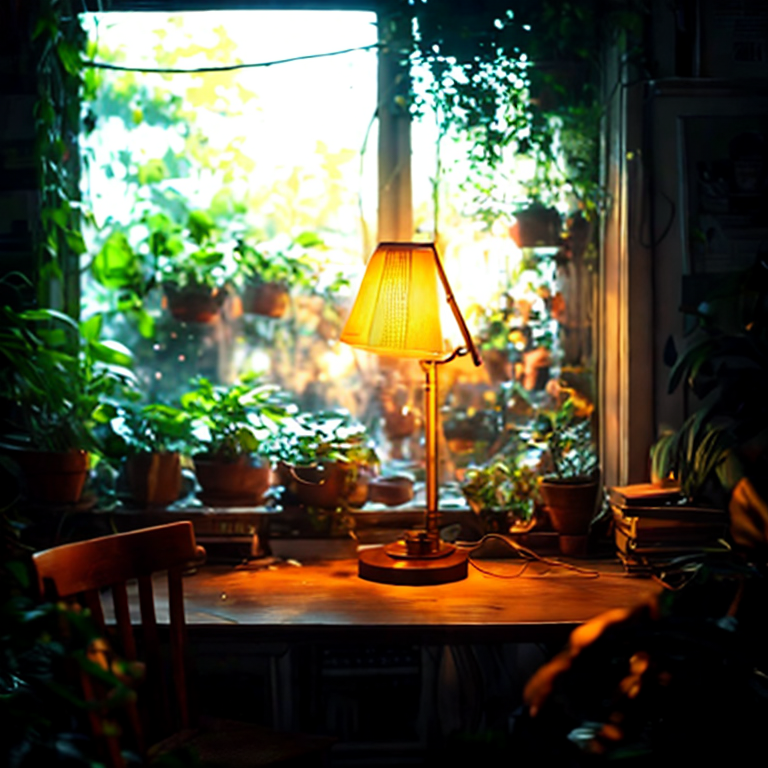} & 
    \includegraphics[width=50pt,height=50pt]{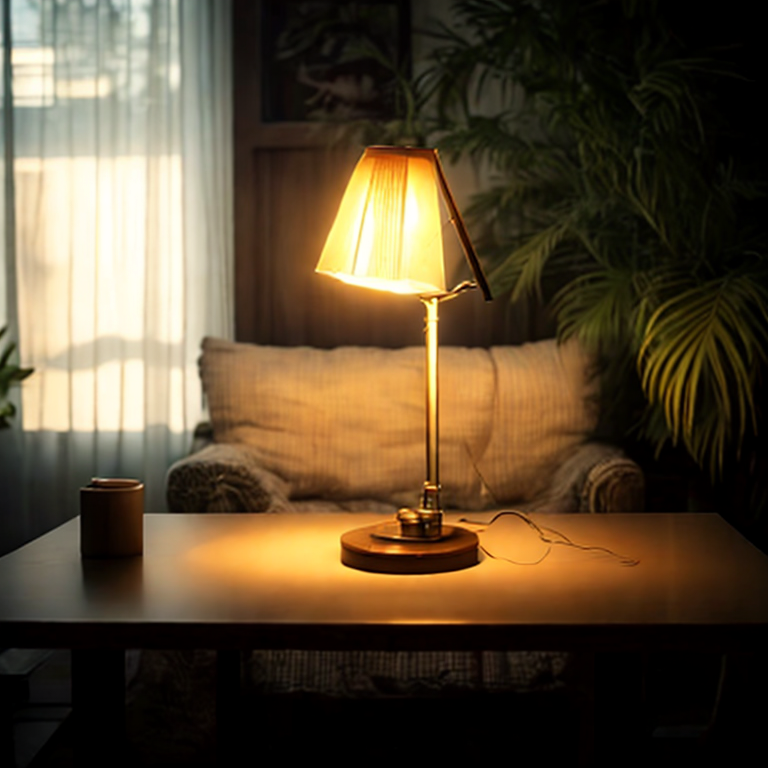} \\[5pt]

    \includegraphics[width=50pt,height=50pt]{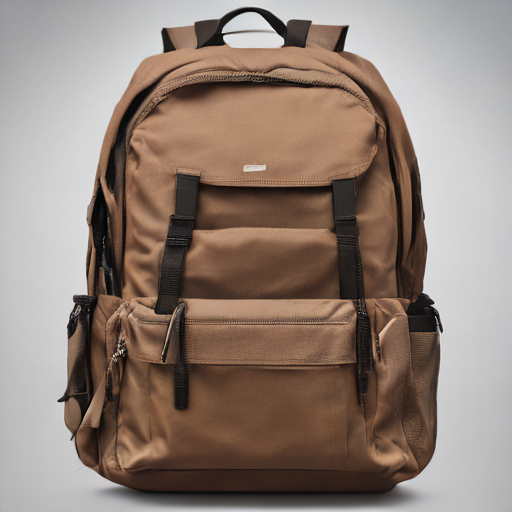} & 
    \includegraphics[width=50pt,height=50pt]{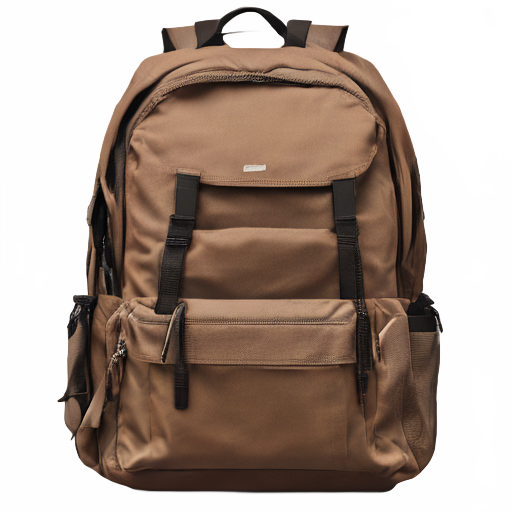} & 
    \includegraphics[width=50pt,height=50pt]{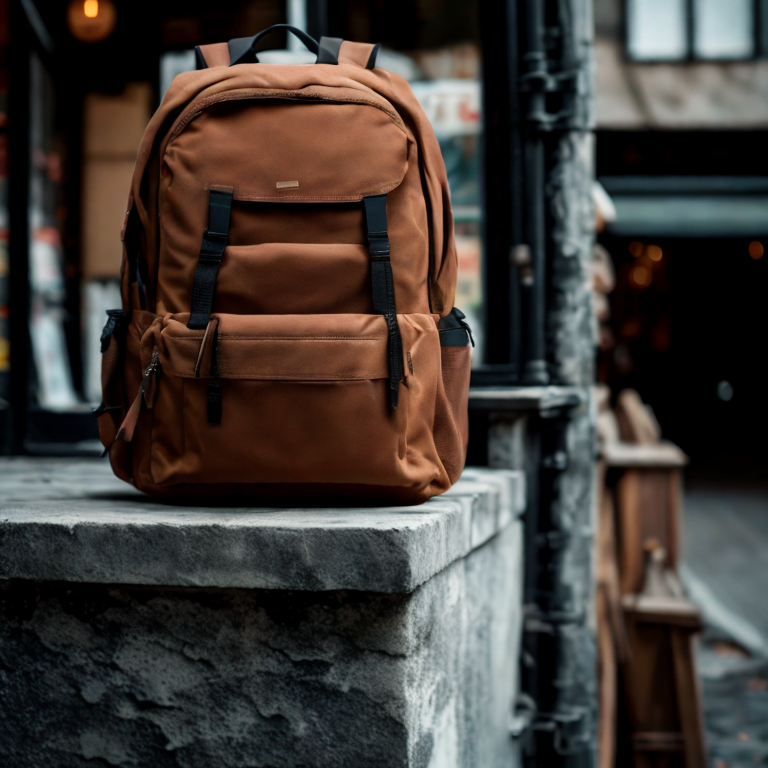} & 
    \includegraphics[width=50pt,height=50pt]{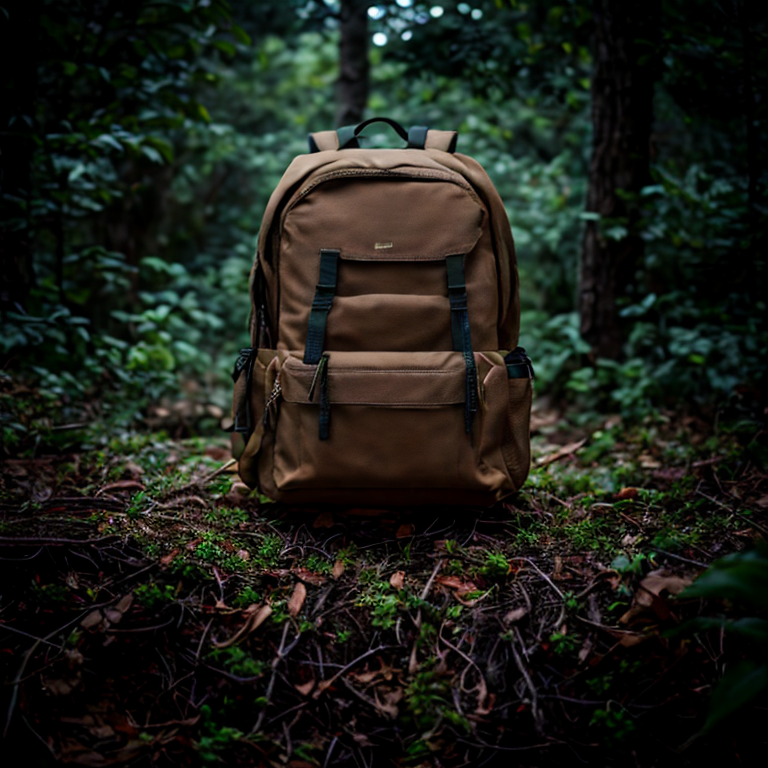} &
    \includegraphics[width=50pt,height=50pt]{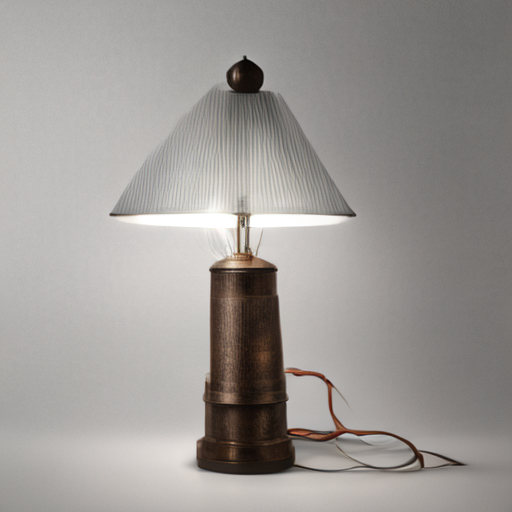} & 
    \includegraphics[width=50pt,height=50pt]{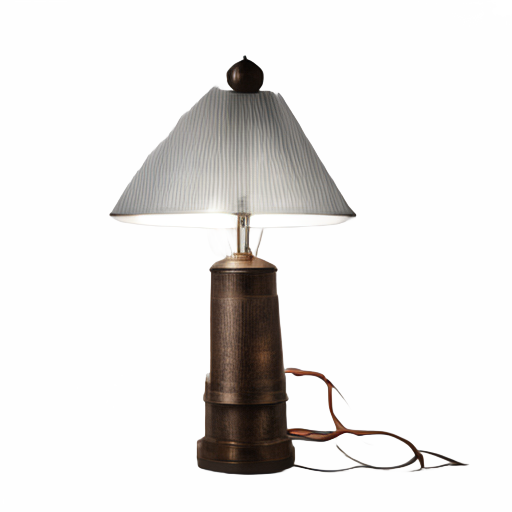} & 
    \includegraphics[width=50pt,height=50pt]{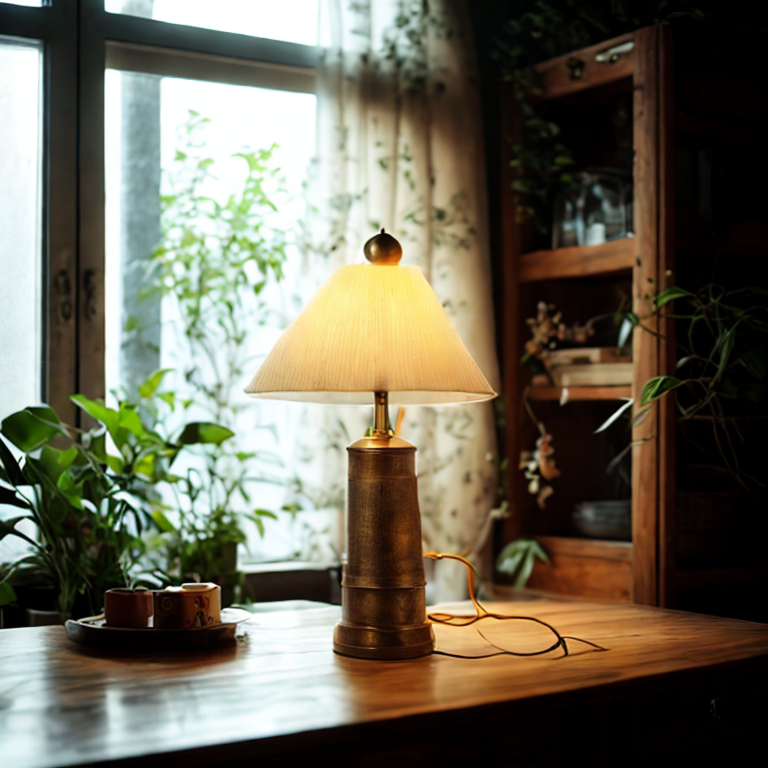} & 
    \includegraphics[width=50pt,height=50pt]{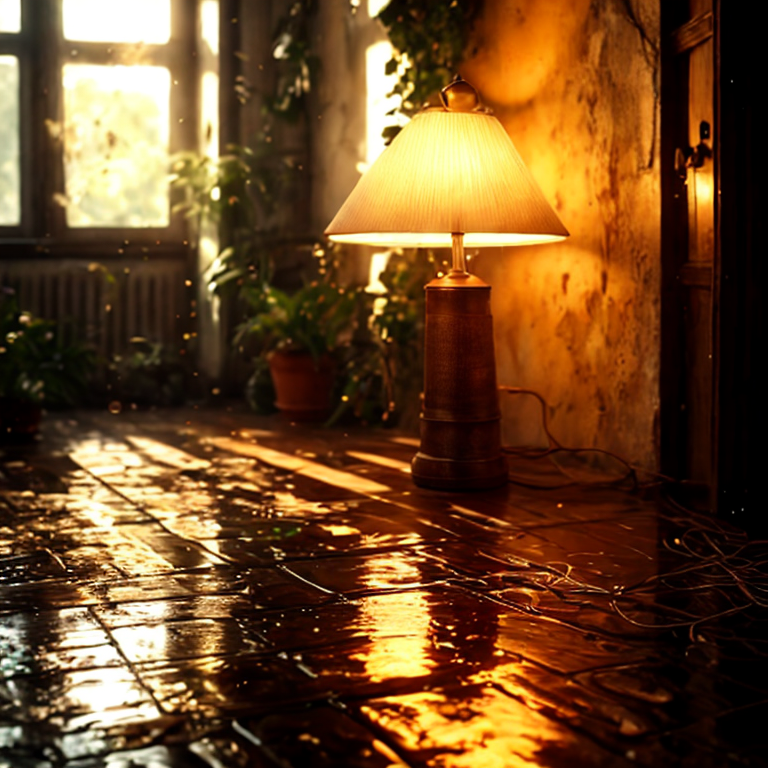}   \\[5pt]
\end{tabular}
\vspace{-15pt}
    \captionof{figure}{Examples of images generated for learning instance-level representations. Given an object generated by a generative diffusion model (column 1), the foreground is segmented (column 2) and different background variations are added (columns 3 \& 4), producing images of the same instance under diverse conditions.}
    \label{fig:fig1}
    \vspace{5pt}
\end{center}

\begin{abstract}
Instance-level recognition (ILR) focuses on identifying individual objects rather than broad categories, offering the highest granularity in image classification. However, this fine-grained nature makes creating large-scale annotated datasets challenging, limiting ILR’s real-world applicability across domains. To overcome this, we introduce a novel approach that synthetically generates diverse object instances from multiple domains under varied conditions and backgrounds, forming a large-scale training set. Unlike prior work on automatic data synthesis, our method is the first to address ILR-specific challenges without relying on any real images. Fine-tuning foundation vision models on the generated data significantly improves retrieval performance across seven ILR benchmarks spanning multiple domains. Our approach offers a new, efficient, and effective alternative to extensive data collection and curation, introducing a new ILR paradigm where the only input is the names of the target domains, unlocking a wide range of real-world applications. The code and pretrained models are publicly available at \url{https://github.com/yankungou/ILGen}.
\end{abstract}
    
\section{Introduction}
\label{sec:intro}
Object recognition and retrieval span multiple levels of granularity, from semantic-level labels~\citep{rds+15} to fine-grained categories~\citep{gmj+14,kjh+15}, and the most detailed form, \ie instance-level recognition (ILR)~\citep{ypsilantis2021met}. Unlike semantic recognition, which groups objects into broad classes, ILR identifies unique object instances, treating each real-world entity as its own category. This extreme granularity makes ILR particularly challenging.

ILR has applications in domains such as landmarks~\citep{wac+20,pci+07,pci+08}, artwork~\citep{ypsilantis2021met}, products~\citep{sxj+15,rp2k}, fashion~\citep{liu2016deepfashion}, and everyday objects~\citep{wj15,ilias}. However, large-scale training data remains a major bottleneck. Unlike semantic or fine-grained recognition, where class names help structure data and reduce false negatives, ILR requires exhaustive, instance-specific annotations, an expensive and labor-intensive process. Single-domain datasets rely on manually curated ground truth, while multi-domain datasets often lack dedicated training sets~\citep{wj15,ilias}. Collecting images of the same instance under different conditions further compounds the challenge, slowing progress.

To address this, we propose a novel pipeline that automatically generates images of unique objects under diverse conditions, enabling instance-level representation learning without manual data collection. The pipeline requires only the name of one or more domains, \eg ``everyday objects'' or ``artworks'', as input and outputs a representation model fine-tuned for those domains. A large language model (LLM)~\citep{hurst2024gpt} generates a list of relevant object categories, and a generative diffusion model (GDM)~\citep{sauer2025adversarial,rombach2022high} synthesizes images for those categories. We assume that generations from a given seed define an instance-level class, while different seeds correspond to distinct classes, and validate this assumption through experimental analysis and user study. To ensure diversity, we introduce background and lighting variations using ICLight~\citep{iclight}.

The generated instances (see \cref{fig:fig1}) are used to fine-tune a foundational vision encoder such as SigLIP~\citep{zhai2023sigmoid}. We adopt a metric learning approach~\citep{ptm22}, treating images of the same instance as positives and others as negatives, and optimize an information retrieval metric across large batches. The resulting representation improves over the base model across multiple ILR benchmarks, including artwork, landmark, and product datasets.

This is the first work to learn a single representation model that generalizes across diverse ILR domains while providing an effective alternative and complement to large-scale real data. While prior research explored synthetic training data~\citep{peng2015learning,fan2024scaling,tian2024stablerep}, our method is the first tailored specifically for ILR. The pipeline synergistically integrates LLMs and GDMs, leveraging rapid advances in both fields and remaining adaptable to future improvements.

\section{Related work}
\label{sec:relatedwork}

\paragraph{Instance-level representations}
Instance-level recognition requires image representations that capture fine-grained object details while distinguishing them from numerous semantically similar classes. Generic models like ResNet~\citep{hzr+16} and CLIP~\citep{rkh+21} struggle in this setting, as they prioritize high-level semantics over instance-specific features. A common solution is fine-tuning pre-trained backbones on domain-specific datasets—such as artwork~\citep{ypsilantis2021met}, landmarks~\citep{lsl+22,sck+23,cas20,ski+24}, or products~\citep{ptm22,ramzi2022hierarchical}—to enhance their ability to differentiate individual instances. Recent efforts focus on universal embeddings~\citep{ypsilantis2023towards} that cover jointly a whole range of domains and tasks. However, models still require fine-tuning with class-supervised learning to acquire the necessary discriminative properties, making the scarcity of high-quality labeled datasets a major challenge. Data augmentation techniques~\citep{ypsilantis2021met} help mitigate this issue by generating diverse variations of an instance from limited samples.
A method leveraging generative models for instance-level tasks~\citep{Sundaram25iclr} fine-tunes a model per instance, requiring a few real images as input. In contrast, we train a single model that generalizes across objects and domains without relying on real images.
Moreover, \cite{kalantidisweatherproofing} uses generative models to augment real images by synthesizing variations that typically distract outdoor visual localization, which is an instance-level task.
\looseness=-1

\paragraph{Training with synthetic images}
Synthetic data has been used in a variety of computer vision problems, such as object detection~\citep{peng2015learning,rozantsev2015rendering,georgakis2017synthesizing}, segmentation~\citep{chen2019learning,ros2016synthia}, autonomous driving~\citep{abu2018augmented}, object pose estimation~\citep{cai2022ove6d, labbe2020cosypose}, 3D-tasks~\citep{chang2015shapenet}, and recently for representation learning~\citep{tian2024stablerep, wu2023not}. An early practice is to cut the real objects and paste them onto backgrounds to generate synthetic images for instance or object detection~\citep{dwibedi2017cut, georgakis2017synthesizing}. However, challenges remain in reducing the boundary artifacts and achieving consistent lighting conditions between the object and background, as these problems often result in unrealistic composite images.
More recently, the main sources of synthetic images are computer graphics pipelines or rendering engines~\citep{mahmood2019amass}, generative adversarial networks (GAN)~\citep{besnier2020dataset,brock2018large}, and text-to-image GDM~\citep{fan2024scaling,sariyildiz2023fake}. Images generated through rendering engines often suffer from domain gap when compared to real-world test images, requiring domain adaptation techniques to mitigate the gap during training. In contrast, GAN and GDM produce more realistic images that do not typically require post-generation domain adaptation~\citep{wang2020self6d}. Text-to-image GDM, in particular, offers a higher degree of control in the image generation process, for example, changing the background of the target object using text prompts~\citep{mokady2023null,raj2023dreambooth3d,geng2024instructdiffusion,zhang2023adding}. This ability to control image features through text makes GDM particularly valuable for generating diverse images, which is crucial for representation learning~\citep{tian2024stablerep, wu2023not}. However, synthesizing images for instance-level task is not trivial, as it requires generating a synthetic object under various conditions while preserving its structure and texture. 

\paragraph{Metric learning for image retrieval}
Given a training dataset, the most common approach for training deep representation networks for image retrieval is supervised learning using categorical labels.
As a result, a large number of methods have proposed classification-based losses~\citep{zw18,dgx+19,tdt20,qss+19,kim2020proxy}.
Despite not directly optimizing the pairwise distance metric that is used at test time, such approaches achieve very good performance, especially when combined with propagating the representation across examples~\citep{evt+20,sel21,kpm+23}. 
Other methods directly optimize the distance metric with pairwise losses. 
These most often rely on hand-crafted loss functions, such as the most popular contrastive~\citep{hcl06}, and triplet loss~\citep{skp+15}, by postulating a correlation between such a training objective and the test time objective which is typically an information retrieval metric. 
Finding informative pairs and triplets~\citep{mbl20,rms+20,sxj+15,sohn16} appears to be very important.
As a natural follow-up, a few recent methods directly optimized differentiable approximations of retrieval metrics, such as average precision~\citep{rmp+20,hls18,rar+19,ramzi2021robust,ramzi2022hierarchical} and recall~\citep{ptm22}. 
In this work, we rely on recall@k~\citep{ptm22} as a loss function which is demonstrating top results on a variety of benchmarks in the literature and does not require hard negative mining.
Self-supervised~\citep{kkc+} methods exist as well and are shown effective, but are tested only on training data from the target distributions, which is not a realistic setup. 
A recent alternative to CLIP~\citep{rkh+21}, called Unicom~\citep{anunicom}, trains on LAION 400M \citep{schuhmann2021laion}, treats captions as weak annotations to perform text-based clustering, and reformulates the learning as a classification task. Their results show improvements in a set of different retrieval datasets, including instance-level ones.
Alternatively, we propose leveraging synthetic data to introduce an extensive collection of objects with diverse variations into the training dataset.

\section{Method}
\begin{figure*}[t]
\centering
\includegraphics[width=.95\linewidth]{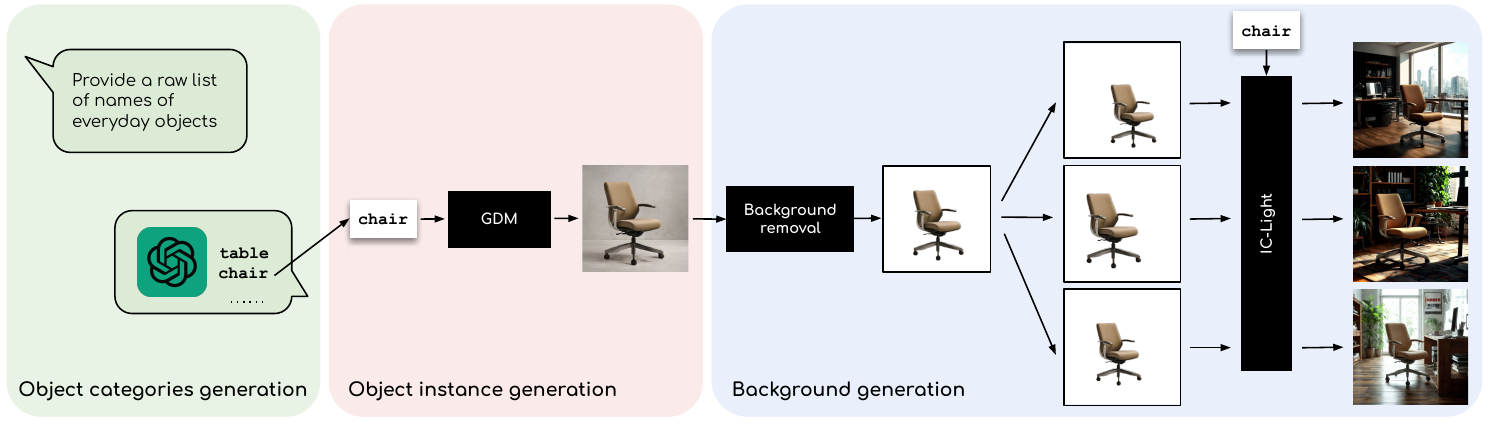}
\caption{Overview of instance-level training data generation.
A domain name or description is the only input, which is used to prompt an LLM to provide a list of object category names. 
Then, we generate examples of those categories using a GDM, remove the background, and synthesize lighting and background multiple times per generated example to create a diverse set of positive images for each instance.
} 
\label{fig:overview}
\end{figure*}

Here, we formulate the target task and describe the training data generation and representation learning. 
An overview of the proposed generation process is shown in \cref{fig:overview}.

\subsection{Task formulation}
The target task is instance-level image retrieval.
Given a query image, the goal is to retrieve all positive images from a database (db), \ie those that depict the same object instance as the query.
Images depicting different object instances, even if they belong to the same semantic category, are negatives and should not be retrieved.
This is an open-world task, testing on unseen objects from a variety of domains which may be seen or unseen during training.

We consider the efficient retrieval variant using global descriptors.  
Formally, an image $x$ is mapped to a $d$-dimensional global descriptor $\mathbf{z} = f_\theta(x) \in \mathbb{R}^d$.  
Retrieval is performed via nearest neighbor search in Euclidean space, ranking database descriptors based on their cosine similarity to the query.  
The encoder, parameterized by $\theta$, is optimized during training.
We focus on fine-tuning foundational models~\citep{zhai2023sigmoid} that already perform well by pretraining.  

\subsection{Instance-level training data generation}
We propose a pipeline that requires only the name, or a textual description, of a target domain as input, and automatically generates an image training set with instance-level labels.
The process consists of four stages:
(i) \textit{Objects categories generation} by prompting an LLM to provide a list of 
object category names;
(ii) \textit{Object instance generation} by prompting a GDM to generate object instances from each category;
(iii) \textit{Background generation} by synthesizing diverse backgrounds per instance;
(iv) \textit{Viewpoint variations} by augmenting the generated images with geometric transformations.
Each stage of the process is detailed below.

\paragraph{Object categories generation}
Object categories (\eg \textit{table}, \textit{chair}, \textit{clock}) are needed to prompt the GDM for image generation. 
We automatically obtain a list of object categories by prompting an LLM with minimal information about the domain of interest. 
In the general case in which we do not target a specific domain, the prompt we use is 
``\emph{Provide a raw list of names of everyday objects.}''
For specific domains, such as artwork, landmark, or product, we enrich the prompt with relevant information and hint with a few examples of object categories. 
Full details of the designed prompts are provided in the supplementary material. 
This approach yields a rich and diverse list of $C$ object categories.
Examples of category names generated for the general case are \textit{sofa}, \textit{desk}, while for the specific domains are \textit{bust}, \textit{castle}, and \textit{polaroid film}, for artwork, landmark, and product, respectively.

\paragraph{Object instance generation}
We prompt a GDM, in particular Stable Diffusion Turbo~\citep{sauer2025adversarial}, with an object category to generate $K$ images per category.
We assume that generating images with different random seeds produces variations that are distinct and recognizable as separate instances within the same category. 
Therefore, following an instance-level class definition, each of the $M$ generated images, where $M = C K$, is treated as a separate class in our training set.
To facilitate the follow-up step of background generation, we target a simple or uniform background. To achieve this, we add ``\emph{in a clean background}" to the prompt after the object category as in, ``\textit{a table in a clean background}." 
Examples in \cref{fig:sdimages} show that, even though the background removal process may fail in both cases, it is less likely to happen with the extended prompt, while the original prompt provides outputs with richer background.

\begin{figure*}[t]
\begin{center}    
\footnotesize
\newcommand{\figclean}[2]{\includegraphics[width=42pt,height=42pt]{fig/sd/#1/clean/#2.png}&\includegraphics[width=42pt,height=42pt]{fig/sd/#1/clean/#2_fg.png}}
\newcommand{\figunclean}[2]{\includegraphics[width=42pt,height=42pt]{fig/sd/#1/unclean/#2.png}&\includegraphics[width=42pt,height=42pt]{fig/sd/#1/unclean/#2_fg.png}}

\begin{tabular}{@{\hspace{0pt}}r@{\hspace{2pt}}c@{\hspace{2pt}}c@{\hspace{2pt}}c@{\hspace{2pt}}cr@{\hspace{2pt}}c@{\hspace{2pt}}c@{\hspace{2pt}}c@{\hspace{2pt}}c@{\hspace{0pt}}}
\raisebox{15pt}{bicycle} & \figclean{Bicycles}{sd_clean_Bicycles_2} & \figunclean{Bicycles}{sd_unclean_Bicycles_1} &\raisebox{15pt}{headphones} & \figclean{headphones}{sd_clean_headphones_1} & \figunclean{headphones}{sd_unclean_headphones_3}\\[5pt]
\raisebox{15pt}{luggage} & \figclean{luggage}{sd_clean_luggage_2} & \figunclean{luggage}{sd_unclean_luggage_0} & \raisebox{15pt}{women's jumpsuit} & \figclean{Womens-jumpsuit}{sd_clean_Womensjumpsuit_0} & \figunclean{Womens-jumpsuit}{sd_unclean_Womensjumpsuit_0}\\[5pt]
\raisebox{15pt}{temple} & \figclean{Temple}{sd_clean_Temple_0} & \figunclean{Temple}{sd_unclean_Temple_2} & \raisebox{15pt}{juicing machine} & \figclean{Juicing-machine}{sd_clean_Juicingmachine_4} & \figunclean{Juicing-machine}{sd_unclean_Juicingmachine_0}\\[5pt]
\raisebox{15pt}{toy car} & \figclean{toy-car}{sd_clean_toycar_1} & \figunclean{toy-car}{sd_unclean_toycar_1} &
\raisebox{15pt}{French empire clock} & \figclean{French-Empire-clock}{sd_clean_FrenchEmpireclock_1} & \figunclean{French-Empire-clock}{sd_unclean_FrenchEmpireclock_0}
\end{tabular}
\caption{Examples of object instances generated by GDM for specific categories. We show the category name, the generated image and the background removal process using ``\emph{in a clean background}'' (columns 1 \& 2) and without it (columns 3 \& 4).\label{fig:sdimages}}
\end{center}
\end{figure*}

\begin{figure*}[!h]
\centering
\hspace{-0.9cm}\includegraphics[width=0.91\linewidth]{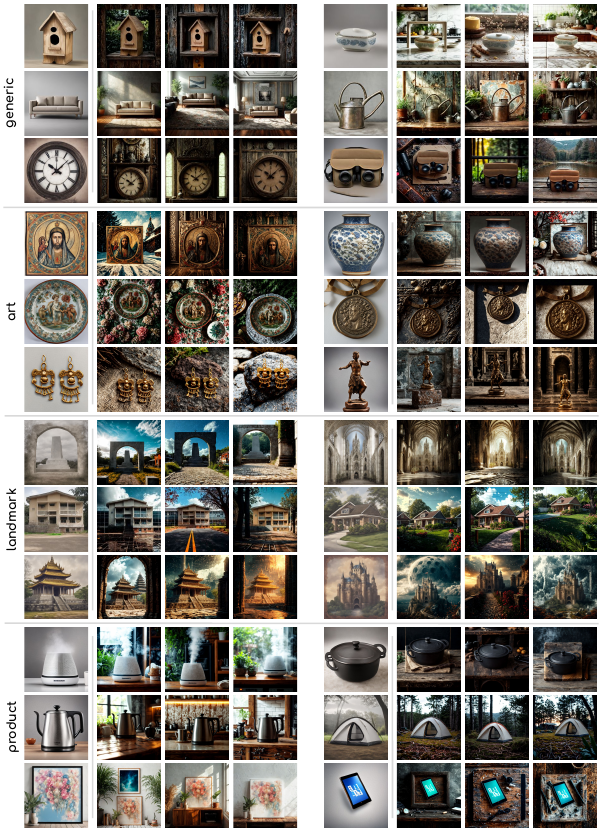}
\vspace{-9pt}
\caption{Examples of object instances generated by GDM (column 1), and the generated images that leave the object intact and add lighting and background that is well suited to the object (columns 2 $\sim$ 4).
\label{fig:iclight}}
\end{figure*}

\paragraph{Background generation} We create variations of an object instance by generating images with multiple, distinct backgrounds and lighting conditions. Given a generated instance in the previous step, we rely on ICLight \citep{iclight} to perform the relighting and add different backgrounds. This process is conducted in three parts:

\begin{enumerate}
\item \textbf{Background removal}.  Firstly, the background of the generated instance is removed to ensure that the input image only depicts the object of interest. 
We use RMBG v1.4\footnote{\url{https://huggingface.co/briaai/RMBG-1.4}}, 
which relies on the IS-Net model~\citep{qin2022highly}, which is a model trained on a large dataset of background-foreground masks. 
The RMBG tool outputs a soft alpha matte indicating foreground probability. The background is removed by alpha-blending the image with a constant background color, resulting in smooth object boundaries. 
\item \textbf{Size and position variation}. We additionally perform padding of random length/width and resize to the original resolution so that the object appears at different sizes and positions. 
Each left/right (top/bottom) pad is up to half of the image width/height. To preserve the aspect ratio and avoid distortion, we first sample the total horizontal padding, \ie, the sum of the left and right padding, compute the corresponding total vertical padding using a scaling factor, and adjust the top and bottom padding accordingly. The padding area is filled with the color of pixel (0, 0) from the image with the background removed.
\item \textbf{New background and lightning}. Then, the object category is used as a prompt to guide ICLight to generate an environment that is commonly appropriate for the specific object.
ICLight is designed to produce realistic lighting on the object and maintain consistent illumination between the object and the background. The semantic coherence between object and background, as well as the faithfulness of the background to the prompt, is largely inherited from the underlying pre-trained Stable Diffusion v1-5\footnote{\url{https://huggingface.co/stablediffusionapi/realistic-vision-v51}}. We apply ICLight with its default settings.
\end{enumerate}
We repeat the last two parts $N$ times per generated object instance with different seeds to generate multiple backgrounds. 
The $N$ images are all elements of the same class in our training set and the only members of this class.
\cref{fig:iclight} shows examples of generated lighting and background for a variety of object categories.

\paragraph{Viewpoint variations}
All images of a class depict the object under different backgrounds and similar viewpoints which only varies because of the padding of the previous step. We additionally rely on simple random geometric augmentations during training to further modify the object's geometry. 
This process resembles self-supervised learning with instance-discrimination~\citep{odm+23,chen2020simple}, where two positive examples are just two different random augmentations of the same input image. 
Nevertheless, there is an essential difference in our case, that the background and lighting significantly vary. Such a factor makes our training setting a unique of its kind. We additionally explore synthesizing viewpoints by rendering multiple angles of GDM-generated instances, then applying background generation for each rendered view. Details are presented in the supplementary.

\subsection{Representation learning}
\begin{figure}[t]
\begin{center}
\includegraphics[width=0.83\linewidth]{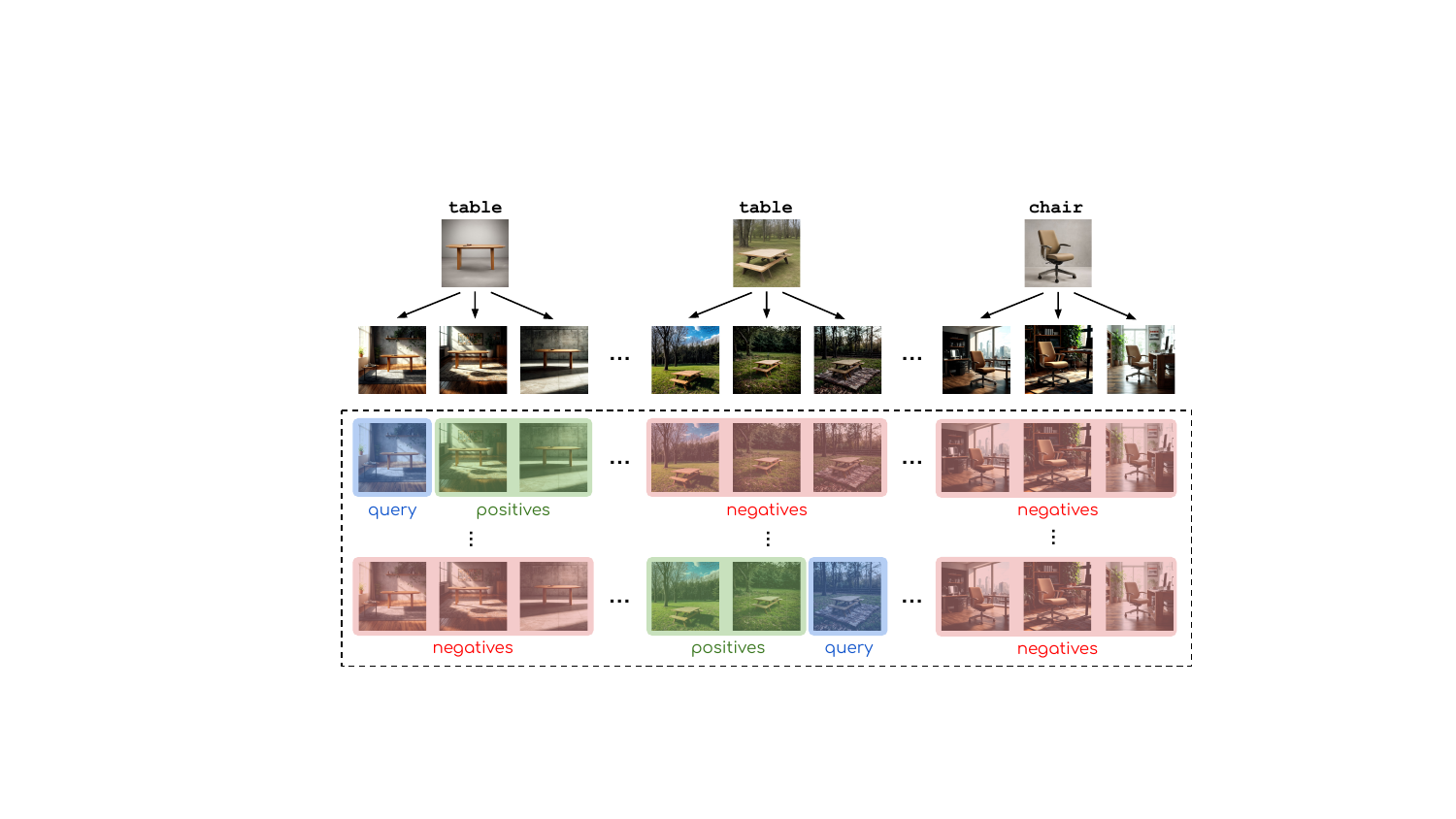}
\vspace{-10pt}
\caption{Training batch construction for instance-level representation learning. A batch simulates a retrieval task with a query (blue) and database of positive (green) and negative (red) images. Images are considered positive if they belong to the same class, otherwise they are negatives. An image encoder is trained with metric learning on this batch.}
\label{fig:batch}
\end{center}
\end{figure}

In total, our generated dataset contains $CKN$ training images, forming $CK$ classes coming from $C$ object categories. 
We construct training batches by sampling $B$ classes and all their corresponding images, resulting in $NB$  images per batch. 
During training, we adopt a query \vs database scheme:
one image from each of the $N$ images per class is randomly chosen as the query, while the remaining $NB-1$ images of the batch form the database, as shown in \cref{fig:batch}.

The similarity between the query and db images is computed in $\hat{\mathbf{y}} \in \mathbb{R}^{NB-1}$, while $\mathbf{y} \in \{0,1\}^{NB-1}$ denotes the labels of all db images with respect to the query, \ie positive or negatives based on their classes. 
We optimize an information retrieval metric as the loss function, in particular an approximation of recall at the top-$k$ ranks, based on $\hat{\mathbf{y}}$, and $\mathbf{y}$.
We train with the average of recall@k loss estimated for different values of $k$.
The approximation of recall is possible by formulating its estimation with the use of step functions, which, during training, are replaced with a sigmoid function.
The technical and implementation details can be found in the original paper~\citep{ptm22}.

\begin{table}[t]
    \centering
    \caption{Statistics of the generated training dataset. \oursgeneric{} and \oursspecific{} comprise only objects from the generic domain and one of the specific domains, respectively.  
    \oursplus{} comprises 50\% of objects from the generic domain ($10$K) and all objects from the three specific domains ($10$K), \ie $20$K objects in total.}
\small
\begin{tabular}{lrrr}
\toprule
\textbf{domain of objects}  & $C$ & $K$ & \textbf{instances} \\ 
\midrule
generic           & $2,000$  & $10$ & $20,000$ \\ 
art        & $200$  & $15$ & $3,000$ \\ 
landmark   & $50$ & $80$  & $4,000$ \\
product    & $200$ & $15$  & $3,000$ \\  
\bottomrule
\end{tabular}

\label{tab:gene_details}
\end{table}

\section{Experiments}
\subsection{Experimental details}

\paragraph{Data generation details} We use GPT-4o \citep{hurst2024gpt} as an LLM for the object categories generation. The LLM is prompted to generate two types of objects: generic and domain-specific. 
Generic objects consist of daily-life objects, while domain-specific objects are objects represented in the particular domains of our evaluation benchmarks. Details about the number of generated object categories are in \cref{tab:gene_details}. 
Since the generic domain broadly covers concepts spanning across several specific domains, we assign it a larger number of object categories ($C$) and use smaller numbers for specific domains given their narrower scope. To balance the total number of instances, we assign different numbers of instances per category ($K$) accordingly. See supplementary for ablation study on $C$ and $K$.
We set the number of inference steps to $1$ when generating instances from each object category using Stable Diffusion Turbo. 
Before applying ICLight to synthesize four distinct backgrounds, \ie $N=4$, we add random padding (up to 50\% of the image resolution) to the foreground-segmented instance, keeping the same aspect ratio. 

\paragraph{Training set variants} To evaluate the quality of our generated data, we compare the performance of the backbone models trained on our generated dataset, some of its variants and alternatives with real objects and/or images.

\begin{table}[t]
    \centering
    \caption{Details of evaluation datasets.}
    \small
\begin{tabular}{lrrrrr}
    \toprule
    \textbf{dataset} & \textbf{queries} & \textbf{database} & \textbf{domain} & \textbf{metric}  \\
    \midrule
    MET	\citep{ypsilantis2021met}         & $19.3$K   & $224.4$K     & artwork & mAP@$100$ \\
    R-Oxford \citep{rit+18}	 & $70$    & $5.0$K + $1$M    & landmark & mAP\\
    R-Paris	\citep{rit+18}     & $70$    & $6.3$K + $1$M     & landmark & mAP\\
    GLDv2 \citep{wac+20}	     & $1.1$K    & $761.8$K    & landmark & mAP@$100$\\
    SOP	\citep{sxj+15}         & $60.5$K & $60.5$K    & product & mAP@$100$ \\
    INSTRE \citep{wj15}	     & $1.3$K	 & $27.3$K     & multi & mAP\\
    mini-ILIAS \citep{ilias}	 & $1.2$K	 & $4.7$K + $5$M & multi & mAP@$1$K \\
    \bottomrule
\end{tabular}
\vspace{-10pt}

    \label{tab:datasets}
\end{table}

\begin{itemize}
    \item \textbf{Pretrained}: The original datasets which the backbones are pretrained on. SigLIP and CLIP are pretrained on web-based text-image datasets, WebLI~\citep{chenpali} and WIT~\citep{rkh+21}, respectively. ViT is pretrained on ImageNet~\citep{dsl+09}. The frozen backbones are evaluated. 
    \item \textbf{\oursplus{} - all domains}: Our generated dataset with $10$K objects from the generic domain and $10$K objects from the specific domains. 
    This dataset is used by default, unless otherwise stated. See \cref{tab:gene_details} for details. 
    \item \textbf{\oursgeneric{} - generic domain}: Our generated dataset with up to $20$K objects from the generic domain only.
    \item \textbf{\oursspecific{} - specific domain}: Our generated dataset with images from only one of the three specific domains.
    \item \textbf{\oursplus {} without background}: Our generated dataset without background generation.
    \item \textbf{Objaverse-background}: Objaverse 1.0~\citep{objaverse} is a large-scale 3D object dataset with $818$K 3D objects across various categories. We randomly select $20$K objects, render each 3D object into $16$ views~\citep{liusyncdreamer}, and choose the four views around the main one, resulting in a total of 80K images to match the statistics of our generated dataset. For each view, we add a background with the same generation process as in our method. This dataset allows us to compare with training on real objects rather than synthetic ones, but on synthesized images via rendering.
    \item \textbf{Real-S - specific domain}: To compare with training on real images that are manually annotated, we use the MET, GLDv2, and SOP training sets to obtain domain-specific models for artwork, landmark, and product, respectively.
    We follow the same dataset split as in \cite{ypsilantis2023towards}. To provide a direct comparison, we use the same number of instances as the corresponding domain-specific parts of our dataset, \ie $3$K, $4$K, and $3$K, respectively, and $4$ images per instance.
    \item \textbf{Real-ALL - all domains}: The above is extended to compose a dataset by merging the training sets of SOP, InShop, RP2k, GLDv2, and MET. We use all classes with at least $4$ images from the first three datasets that are small, and complement with enough classes equally from the other two datasets to reach $20$K instances. We sample $4$ images per class.
\end{itemize}

\paragraph{Training details} 
During training, we use random cropping, resizing, flipping, color jitter, and mapping to grayscale as image augmentations~\citep{hfw+20}.
We use a batch size of $1,600$ images ($B=400$, $N=4$) and optimize over $400$ queries, one per class. 
We use the vanilla version of the recall@k loss with its default hyper-parameters, $k=\{1,2,4,8\}$ for the recall@k loss with the two temperatures set to $0.01$ and $1.0$ as in the original work, and train until all classes have been loaded in a batch.
We use learning rate $10^{-5}$ and Adam optimizer~\citep{kb15} with a weight decay $10^{-6}$. Experiments are run on a single A100 or V100 GPU. The training process of \oursplus{} with SigLIP takes approximately $2.5$ hours on an A100 GPU.

\paragraph{Backbones} 
We use SigLIP ViT-L/16 \citep{zhai2023sigmoid}, CLIP ViT-L/14 \citep{rkh+21}, and ViT-B/16 \citep{dbk+21}, briefly referred to as SigLIP, CLIP, and ViT-B. Images are resized to $336 \times 336$, $384 \times 384$, and $224 \times 224$ pixels, respectively, according to their pretraining setup. We load the pre-trained models from timm\footnote{\url{https://timm.fast.ai/}} and treat the [CLS] token as the global descriptor. 

\begin{table*}[t]
    \centering
    \caption{Evaluation results using SigLIP with different training datasets, number of instances, and use of synthetic background (bg). \oursgeneric{} uses generic domain object categories, while \oursplus{} includes domain-specific objects. We train each setting for 3 seeds, and report the mean and standard deviation.}
\vspace{5pt}
\small

\newcommand{\stdv}[2]{#1{\scriptsize$\pm$#2}}

\scalebox{0.95}{
\begin{tabular}{@{}clrc c c@{\hspace{1pt}} cc c@{\hspace{1pt}} c c@{\hspace{1pt}} cc@{}}
\toprule

\multirow{2}{*}{ID} & \multirow{2}{*}{data} & \multirow{2}{*}{instance} & \multirow{2}{*}{avg} & \multicolumn{1}{c}{artwork} & & \multicolumn{2}{c}{landmark} & & \multicolumn{1}{c}{product} & & \multicolumn{2}{c}{multi} \\
\cline{5-5}
\cline{7-8}
\cline{10-10}
\cline{12-13}
 &  &  &  & MET & & ROP & GLD & & SOP & & INS & mIL \\

\midrule 
1 & pretrained & - & $47.5$ & $67.3$ & & $45.0$ & $15.7$ & & $55.4$ & & $80.6$ & $21.0$ \\

\midrule

2 & Objaverse-background & $20$K & \stdv{$50.9$}{0.5} & \stdv{$74.1$}{0.1} & & \stdv{$42.6$}{1.0} & \stdv{$15.9$}{0.5} & & \stdv{$57.6$}{0.4} & & \stdv{$86.8$}{0.5} & \stdv{$28.6$}{1.1} \\

\midrule
3 & \oursgeneric{} & $5$K & \stdv{$51.2$}{0.3} & \stdv{$72.8$}{0.6} & & \stdv{$46.5$}{0.3} & \stdv{$17.4$}{0.1} & & \stdv{$55.3$}{0.3} & & \stdv{$85.9$}{0.3} & \stdv{$29.4$}{1.6} \\

4 & \oursgeneric{} & $10$K & \stdv{$51.6$}{0.2} & \stdv{$72.7$}{0.1} & & \stdv{$46.3$}{0.2} & \stdv{$17.7$}{0.2} & & \stdv{$55.4$}{0.1} & & \stdv{$87.1$}{0.0} & \stdv{$30.2$}{1.0} \\

5 & \oursgeneric{} & $20$K & \stdv{$50.8$}{0.2} & \stdv{$72.4$}{0.2} & & \stdv{$46.4$}{0.4} & \stdv{$17.4$}{0.3} & & \stdv{$55.9$}{0.1} & & \stdv{$85.4$}{0.1} & \stdv{$27.4$}{1.1} \\
\midrule

6 & \oursplus{} w/o bg & $20$K & \stdv{$49.5$}{0.3} & \stdv{$73.0$}{0.6} & & \stdv{$46.8$}{1.2} & \stdv{$17.4$}{0.2} & & \stdv{$\textbf{61.1}$}{0.5} & & \stdv{$77.3$}{0.6} & \stdv{$21.5$}{1.0} \\

7 & \oursplus{} & $20$K & \stdv{$\textbf{52.7}$}{0.3} & \stdv{$\textbf{75.1}$}{0.2} & & \stdv{$\textbf{48.6}$}{0.3} & \stdv{$\textbf{18.7}$}{0.4} & & \stdv{$55.6$}{0.3} & & \stdv{$\textbf{87.5}$}{0.4} & \stdv{$\textbf{30.6}$}{0.6} \\

\bottomrule
\end{tabular}
}

\label{tab:all_results}
\end{table*}

\paragraph{Evaluation benchmarks}
We use a set of standard and diverse ILR retrieval and classification datasets for evaluation. ILR datasets are comprised of queries, a database in which the same instances as queries exist as positives, and occasionally, a distractor set of irrelevant images. 
Details are provided in \cref{tab:datasets} and the dataset list is as follows:
\begin{itemize}
    \item \textbf{Artwork domain}: The MET dataset~\citep{ypsilantis2021met} comprises a database of catalog photos from the Metropolitan Museum of Art and query images taken by visitors inside the museum. To adapt the benchmark for retrieval, we retain only queries with at least one positive match in the database, \ie we discard the distractor queries, and keep only the first positive per query in the database asserting visual overlap between the two images.
    \item \textbf{Landmark domain}: R-Oxford \citep{rit+18}, R-Paris \citep{rit+18}, and GLDv2 \citep{wac+20} are the most widely used datasets in this domain. 
    For R-Oxford and R-Paris, we report results on the Medium and Hard evaluation split with 1M distractors, and following standard practice, we report average performance across the two datasets, denoted as ROP. 
    \item \textbf{Product domain}: SOP \citep{sxj+15} whose images are crawled from e-commerce websites.
    \item \textbf{Multi-domain}: We use INSTRE \citep{wj15} and ILIAS \citep{ilias} which include a variety of objects from multiple domains such as daily objects, landmarks, etc. We use the mini version of ILIAS with 5M distractor images.
\end{itemize}

\subsection{Results for different training sets}
\cref{tab:all_results} shows the main results for SigLIP after training on a variety of datasets.

\paragraph{Impact of synthetic data} 
\oursplus{} (ID7) provides consistent improvement compared to the pretrained (ID1) model on all datasets, with an average improvement equal to $5.2$.
Compared to Objaverse, which uses images rendered from 3D objects rather than automatically generated, \oursplus{} performs better on most datasets, especially on SOP.
This suggests that our method, which relies solely on synthesized objects, learns representations that are at least as effective as those learned on rendered objects.

\begin{figure}[t]
\centering
\definecolor{appleblue}{RGB}{0,122,255}
\begin{tabular}{@{\hspace{0pt}}c@{\hspace{0pt}}c@{\hspace{0pt}}c@{\hspace{0pt}}}
\begin{tikzpicture}
    \small
    \begin{axis}[
        width=0.3\linewidth,
        height=0.3\linewidth,
        xmin=-1,
        xmax=101,
        ymin=-1,
        ymax=101,
        title={\small mini-ILIAS},
        grid=both,
        grid style={color=lightgray!60, dash pattern=on 2pt off 2pt},
        xtick={0, 20, 40, 60, 80, 100},
        ytick={0, 20, 40, 60, 80, 100},
        title style = {yshift = -5pt},
        xlabel = {\oursplus{} (mAP@$1$K)},
        ylabel = {Pretrained (mAP@$1$K)},
        xlabel style={yshift=2pt},
        ylabel style={yshift=-3pt},
        xticklabel style=
    {/pgf/number format/1000 sep=,anchor=north,font=\small},
        legend pos=north west,
        legend style={cells={anchor=east}, font=\small, fill opacity=0.7, row sep=-1pt},
    ]
    \addplot[only marks, mark=*, opacity=0.7, mark size=1.2, color=appleblue] 
    table[x=ours, y=pretrained] {./fig/ap_ilias_siglip_resize_large.dat};
    \end{axis}
\end{tikzpicture}
&
\begin{tikzpicture}
    \small
    \begin{axis}[
        width=0.3\linewidth,
        height=0.3\linewidth,
        xmin=-1,
        xmax=101,
        ymin=-1,
        ymax=101,
        title={\small INSTRE},
        grid=both,
        grid style={color=lightgray!60, dash pattern=on 2pt off 2pt},
        xtick={0, 20, 40, 60, 80, 100},
        ytick={0, 20, 40, 60, 80, 100},
        title style = {yshift = -5pt},
        xlabel = {\oursplus{} (mAP)},
        ylabel = {Pretrained (mAP)},
        xlabel style={yshift=2pt},
        ylabel style={yshift=-3pt},
        xticklabel style=
    {/pgf/number format/1000 sep=,anchor=north,font=\small},
        legend pos=north west,
        legend style={cells={anchor=east}, font=\small, fill opacity=0.7, row sep=-1pt},
    ]
    \addplot[only marks, mark=*, opacity=0.7, mark size=1.2, color=appleblue] 
    table[x=ours, y=pretrained] {./fig/ap_instre_siglip_resize_large.dat};
    \end{axis}
\end{tikzpicture}
&
\begin{tikzpicture}
    \small
    \begin{axis}[
        width=0.3\linewidth,
        height=0.3\linewidth,
        xmin=-1,
        xmax=101,
        ymin=-1,
        ymax=101,
        title={\small MET},
        grid=both,
        grid style={color=lightgray!60, dash pattern=on 2pt off 2pt},
        xtick={0, 20, 40, 60, 80, 100},
        ytick={0, 20, 40, 60, 80, 100},
        title style = {yshift = -5pt},
        xlabel = {\oursplus{} (mAP@$100$)},
        ylabel = {Pretrained (mAP@$100$)},
        xlabel style={yshift=2pt},
        ylabel style={yshift=-3pt},
        xticklabel style=
    {/pgf/number format/1000 sep=,anchor=north,font=\small},
        legend pos=north west,
        legend style={cells={anchor=east}, font=\small, fill opacity=0.7, row sep=-1pt},
    ]
    \addplot[only marks, mark=*, opacity=0.7, mark size=1.2, color=appleblue] 
    table[x=ours, y=pretrained] {./fig/ap_met_siglip_resize_large.dat};
    \end{axis}
\end{tikzpicture}
\end{tabular}
\vspace{-7pt}
\caption{Average Precision (AP) per query for the pretrained backbone (y-axis) and the backbone fine-tuned on \oursplus{} (ID7) (x-axis). 
Each point represents a query in the evaluation dataset. Points below the diagonal indicate a query with improved performance when fine-tuned on \oursplus{}. Results using SigLIP.
\label{fig:scatter}}
\vspace{-12pt}
\end{figure}

\begin{table}[t]
    \centering
    \caption{Comparison between training on real-labeled images and training our synthetic images on four different domains using SigLIP.} 
\small 
\begin{tabular}
{@{}clcc c cc c c c cc@{}}
    \toprule
     \multirow{2}{*}{ID} & \multirow{2}{*}{dataset} & \multirow{2}{*}{avg} & \multicolumn{1}{c}{artwork} & & \multicolumn{2}{c}{landmark} & & \multicolumn{1}{c}{product} & & \multicolumn{2}{c}{multi} \\
     \cline{4-4}
     \cline{6-7}
     \cline{9-9}
     \cline{11-12}

     & & & MET & & ROP & GLD & & SOP & & INS & mIL \\
    \midrule 
    $1$ & pretrained & $47.5$ & $67.3$ & & $45.0$ & $15.7$ & & $55.4$ & & $80.6$ & $21.0$ \\

    \midrule 
    $8$ & Real-S (artwork)   & $49.9$ & $\textbf{75.2}$ & & $46.8$ & $17.18$ & & $\textbf{57.0}$ & & $80.9$ & $22.4$ \\
    $9$ & \oursspecific{} (artwork) & $\textbf{51.2}$ & $73.7$ & & $\textbf{47.0}$ & $\textbf{17.24}$ & & $55.6$ & & $\textbf{85.4}$ & $\textbf{28.3}$ \\
    \midrule 
    $10$ & Real-S (landmark)  & $50.0$ & $69.6$ & & $\textbf{55.0}$ & $\textbf{19.8}$ & & $\textbf{56.7}$ & & $78.6$ & $20.2$ \\
    $11$ & \oursspecific{} (landmark)  & $\textbf{51.0}$ & $\textbf{72.5}$ & & $50.7$ & $19.7$ & & $54.6$ & & $\textbf{84.4}$ & $\textbf{24.2}$ \\
    \midrule 
    $12$ & Real-S (product)  & $48.3$ & $63.8$ & & $46.1$ & $16.9$ & & $\textbf{60.3}$ & & $80.7$ & $21.8$ \\
    $13$ & \oursspecific{} (product) & $\textbf{50.5}$ & $\textbf{71.6}$ & & $\textbf{46.4}$ & $\textbf{17.0}$ & & $55.9$ &&  $\textbf{85.0}$ & $\textbf{27.1}$ \\
    \midrule 
    $14$ & Real-ALL  & $51.4$ & $69.3$ & & $\textbf{55.3}$ & $\textbf{19.7}$ & & $\textbf{71.8}$ & & $72.7$ & $19.3$ \\
    7 & \oursplus{} & $\textbf{52.7}$ & $\textbf{75.1}$ & & $48.6$ & $18.7$ & & $55.6$ & & $\textbf{87.5}$ & $\textbf{30.6}$ \\
    \bottomrule
\end{tabular}

\vspace{-5pt}
\label{tab:synth_vs_real}
\end{table}

\paragraph{Number of instances}
We evaluate SigLIP backbone trained on the generic-domain version of the dataset, \oursgeneric{}, with different numbers of instances: $5$K, $10$K, and $20$K (corresponding to ID3, ID4, and ID5 in \cref{tab:all_results}). 
Even with the smallest set of $5$K generic instances (ID3), performance on all the benchmarks is better than the pre-trained backbone (ID1) except SOP where performance does not change. When the number of instances increases to $10$K (ID4), the average performance increases further, but saturates for the largest set (ID5).

\paragraph{Diverse \vs clean background}
Training on \oursplus{} with clean background (ID6) improves the performance on most datasets compared to the pretrained backbone. However, performance drops on INSTRE and the improvement is small on mini-ILIAS, which are two datasets with high background clutter. 
Synthesizing realistic and diverse backgrounds (ID7) leads to a substantial improvement on most datasets compared to clean background (ID6).
SOP forms an exception, where having clean background is the variant that brings a noticeable improvement, which is related to the commonly clean background in this test set.

\paragraph{Domain of the instances} 
Complementing \oursgeneric{}-$10$K (ID4) with $10$K images from domain-specific objects (ID7) is much better on average than complementing it with $10$K generic objects (ID5). Such a choice strengthens performance on all test sets across domains except SOP
Therefore, leveraging synthetic images in a diverse set of targeted domains, our method has the potential to effectively address data scarcity and obtain universal representation models.

\paragraph{Improvement per query}
In \cref{fig:scatter}, we compare the performance of the pretrained and the fine-tuned SigLIP on \oursplus{} (ID7) on a query basis. 
Training on the dataset of the proposed method improves the performance on the majority of queries and over the whole range of performance values with the pretrained model, even for many highly performing queries of INSTRE.

\paragraph{Comparison to real manually labeled images}
We train SigLIP on both real-labeled and our synthetic images with recall@k loss under the same setting and present results in \cref{tab:synth_vs_real}. 
We make the following observations. 
Training with our synthetic images yields better overall performance compared to real-labeled images. Although training with real images from a single domain achieves better performance within the specific domain, our synthetic images have better performance across other domains except for product. Notably, results on multi-domain (INSTRE and mini-ILIAS) reveal that our synthetic images are the best in all cases, indicating the strength of our approach to cover a large range of domains. Performance when testing on ROP is always better when training on real images, possibly indicating shortcomings of the generative models for large objects with many details.

\begin{table*}[t]
\centering
\caption{Training on a mixture of real and synthetic data (ID S25-30) performs better than training solely on real (ID14) or synthetic data (ID7).}

\begin{tabular}{@{}cccc cc c@{\hspace{1pt}} cc c@{\hspace{1pt}} c c@{\hspace{1pt}} cc@{}}
\toprule

\multirow{2}{*}{ID} & \multicolumn{2}{c}{training instances} & & \multirow{2}{*}{avg} & \multicolumn{1}{c}{artwork} & & \multicolumn{2}{c}{landmark} & & \multicolumn{1}{c}{product} & & \multicolumn{2}{c}{multi} \\
\cline{2-3}
\cline{6-6}
\cline{8-9}
\cline{11-11}
\cline{13-14}
     
 & \oursplus{} & real & & & MET & & ROP & GLD & & SOP & & INS & mIL \\
\midrule 

$1$ & - & - & & $47.5$ & $67.3$ & & $45.0$ & $15.7$ & & $55.4$ & & $80.6$ & $21.0$ \\

\hline 
$14$ & - & $20,000$ & & $51.4$ & $69.3$ & & $55.3$ & $19.7$ & & $71.8$ & & $72.7$ & $19.3$ \\
$7$ & $20,000$ & - & & $52.7$ & $75.1$ & & $48.6$ & $18.7$ & & $55.6$ & & $87.5$ & $30.6$ \\
\hline

S25 & $20,000$ & $2,000$ & & $53.5$ & $76.3$ & & $51.0$ & $19.4$ & & $58.3$ & & $86.3$ & $29.7$ \\
S26 & $20,000$ & $4,000$ & & $53.9$ & $76.8$ & & $51.1$ & $19.4$ & & $60.8$ & & $86.2$ & $29.4$ \\
S27 & $20,000$ & $8,000$ & & $54.2$ & $75.6$ & & $53.7$ & $18.7$ & & $64.7$ & & $84.5$ & $27.7$ \\
S28 & $20,000$ & $12,000$ & & $55.3$ & $75.2$ & & $55.0$ & $20.2$ & & $66.7$ & & $85.5$ & $29.2$ \\
S29 & $20,000$ & $16,000$ & & $54.0$ & $73.3$ & & $54.1$ & $18.8$ & & $69.2$ & & $82.7$ & $25.9$ \\
S30 & $20,000$ & $20,000$ & & $55.0$ & $75.3$ & & $56.5$ & $20.4$ & & $70.9$ & & $81.4$ & $25.4$ \\

\bottomrule
\end{tabular}
\label{tab:sup:mix_real_syn}
\end{table*}

\paragraph{Training with both real and synthetic data}
\cref{tab:sup:mix_real_syn} presents results for progressively adding real data on top of synthetic data (ID-S25 to ID-S29), up to using all the real data (ID-S30). 
Overall, in terms of average performance across all datasets, the mixture of real and synthetic training data performs better than using only  real or only synthetic data. 
This holds for all proportions of mixing the data, with better overall performance for balanced (ID-S30)  or nearly balanced (ID-S28) mixing. 
In domains where training data is rich and diverse the mixed data performs slightly worse than using only real images; products being the only such case. 
This experiment indicates that synthetic data serves, not only as a replacement, but also as an effective complement to real data for instance-level learning across various domains.

\begin{table}[t]
    \centering
    \caption{Evaluation results on different backbones. Representations learned on synthetic data using \oursplus{} outperform the pretrained representations on all datasets, except ViT on SOP.} 
\small
\begin{tabular}{@{}llcc c@{\hspace{1pt}} cc c@{\hspace{1pt}} c c@{\hspace{1pt}} cc@{}}
\toprule

\multirow{2}{*}{model} & \multirow{2}{*}{data} & \multirow{2}{*}{avg} & \multicolumn{1}{c}{artwork} & & \multicolumn{2}{c}{landmark} & & \multicolumn{1}{c}{product} & & \multicolumn{2}{c}{multi} \\
\cline{4-4}
\cline{6-7}
\cline{9-9}
\cline{11-12}
     
 & & & MET & & ROP & GLD & & SOP & & INS & mIL \\
\midrule 

\multirow{2}{*}{SigLIP} & pretrained & $47.5$ & $67.3$ & & $45.0$ & $15.7$ & & $55.4$ & & $80.6$ & $21.0$ \\
 & \oursplus{} & $\textbf{52.7}$ & $\textbf{75.1}$ & & $\textbf{48.6}$ & $\textbf{18.7}$ & & $\textbf{55.6}$ & & $\textbf{87.5}$ & $\textbf{30.6}$ \\
\midrule

\multirow{2}{*}{CLIP} & pretrained & $37.5$ & $47.1$ & & $40.0$ & $10.5$ & & $41.8$ & & $75.1$ & $10.4$ \\
 &  \oursplus{} & $\textbf{46.8}$ & $\textbf{69.6}$ & & $\textbf{43.7}$ & $\textbf{16.8}$ & & $\textbf{45.5}$ & & $\textbf{81.7}$ & $\textbf{23.8}$ \\
\midrule

\multirow{2}{*}{ViT-B} & pretrained & $25.7$ & $34.2$ & & $24.6$ & $5.7$ & & $\textbf{43.7}$ & & $41.9$ & $4.0$ \\
&  \oursplus{} & $\textbf{34.3}$ & $\textbf{50.8}$ & & $\textbf{29.8}$ & $\textbf{9.1}$ & & $40.8$ & & $\textbf{65.1}$ & $\textbf{10.1}$ \\

\bottomrule
\end{tabular}

\vspace{0pt}
\label{tab:backbone_results}
\end{table}

\begin{table}[t]
    \centering
    \caption{Evaluation results by training SigLIP on \oursplus{} using different loss function.} 
\small
\begin{tabular}{@{}lcc c@{\hspace{1pt}} cc c@{\hspace{1pt}} c c@{\hspace{1pt}} cc@{}}
    \toprule
     \multirow{2}{*}{loss} & \multirow{2}{*}{avg} & \multicolumn{1}{c}{artwork} & & \multicolumn{2}{c}{landmark} & & \multicolumn{1}{c}{product} & & \multicolumn{2}{c}{multi} \\
     \cline{3-3}
     \cline{5-6}
     \cline{8-8}
     \cline{10-11}

     & & MET & & ROP & GLD & & SOP & & INS & mIL \\
    \midrule 
    pretrained & $47.5$ & $67.3$ & & $45.0$ & $15.7$ & & $55.4$ & & $80.6$ & $21.0$ \\
    \midrule 
    recall@k ~\citep{ptm22} & $\textbf{52.7}$ & $\textbf{75.10}$ & & $\textbf{48.56}$ & $18.7$ & & $\textbf{55.6}$ & & $87.5$ & $30.6$ \\
    infoNCE ~\citep{chen2020simple}  & $52.2$ & $75.06$ & & $48.55$ & $\textbf{18.8}$ & & $54.2$ & & $86.0$ & $30.7$ \\
    contrastive ~\citep{chl05} & $50.6$ & $62.8$ & & $46.0$ & $16.1$ & & $53.8$ & & $86.4$ & $\textbf{38.4}$ \\
    softmax margin ~\citep{wwz+18}  & $51.5$ & $70.1$ & & $47.3$ & $18.4$ & & $55.4$ & & $\textbf{88.0}$ & $29.8$ \\
    \bottomrule
\end{tabular}

\label{tab:loss_results}
\end{table}

\begin{table*}[t]
  \centering
    \caption{Ablation study on training data (S1-S2), LLM (S3-S5), GDM (S6-S7), and background generation (S8-S9). ID1 (pretrained) and ID7 (\oursplus{}) were presented in \cref{tab:all_results}. Each ablation modifies only one \colorbox{oursrow}{component} compared to ID7. \textit{Pos} refers to the number of training images per instance class. \textit{Steps} are the inference steps during image generation. SD Turbo uses $1$ step by default. SD refers to Stable Diffusion.}
\vspace{10pt}
\footnotesize
\centering
\scalebox{0.89}{
\begin{tabular}{c l@{\hspace{6pt}}c@{\hspace{2pt}} cl@{\hspace{6pt}}l@{\hspace{4pt}} c@{\hspace{6pt}}l@{\hspace{6pt}}c@{\hspace{2pt}}  c@{\hspace{6pt}}l@{\hspace{6pt}}c@{\hspace{2pt}} c@{\hspace{8pt}}c@{\hspace{8pt}}c@{\hspace{8pt}}c@{\hspace{8pt}}c@{\hspace{8pt}}c@{\hspace{8pt}}c@{\hspace{8pt}}c@{\hspace{8pt}}c@{\hspace{8pt}}c@{}}
\toprule

\multirow{2}{*}{\textbf{ID}} & \multicolumn{2}{c}{\textbf{data}} & & \multicolumn{2}{c}{\textbf{LLM}} & & \multicolumn{2}{c}{\textbf{GDM}} & & \multicolumn{2}{c}{\textbf{background}} & & \multicolumn{7}{c}{\textbf{results}} \\
 \cline{2-3}
 \cline{5-6}
 \cline{8-9}
 \cline{11-12}
 \cline{14-20}

& dataset & pos & & prompt & model & & model & steps & & model & padding & & avg & MET & ROP & GLD & SOP & INS & mIL \\
\midrule 

1 & pretrained & - & & - & - & & - & - & & - & - & & $47.5$ & $67.3$ & $45.0$ & $15.7$ & $55.4$ & $80.6$ & $21.0$ \\
\midrule

S1 & generated & \cellcolor{oursrow}$3$ & & designed & GPT-4o & & SD Turbo & $1$ & & ICLight & \ding{52} & & $51.5$ & $73.7$ & $47.8$ & $18.3$ & $56.2$ & $85.6$ & $27.7$ \\
S2 & generated & \cellcolor{oursrow}$2$ & & designed & GPT-4o & & SD Turbo & $1$ & & ICLight & \ding{52} & & $50.3$ & $71.3$ & $46.6$ & $17.8$ & $55.3$ & $85.5$ & $25.2$ \\

\hline
S3 & generated & $4$ & & \cellcolor{oursrow}template & GPT-4o & & SD Turbo & $1$ & & ICLight & \ding{52} & & $52.6$ & $74.9$ & $48.0$ & $18.6$ & $56.5$ & $86.2$ & $31.6$ \\
S4 & generated & $4$ & & designed & \cellcolor{oursrow}DeepSeek & & SD Turbo & $1$ & & ICLight & \ding{52} & & $52.6$ & $75.3$ & $47.0$ & $18.2$ & $55.0$ & $88.0$ & $32.1$ \\
S5 & generated & $4$ & & designed & \cellcolor{oursrow}Claude & & SD Turbo & $1$ & & ICLight & \ding{52} & & $52.5$ & $74.6$ & $48.8$ & $18.3$ & $55.5$ & $87.5$ & $30.6$ \\

\hline
S6 & generated & $4$ & & designed & GPT-4o & & \cellcolor{oursrow}SD v2.0 & $50$ & & ICLight & \ding{52} & & $51.8$ & $74.1$ & $47.6$ & $18.2$ & $56.6$ & $86.9$ & $27.6$ \\
S7 & generated & $4$ & & designed & GPT-4o & & SD Turbo & \cellcolor{oursrow}$5$ & & ICLight & \ding{52} & & $53.0$ & $74.7$ & $49.1$ & $18.2$ & $56.6$ & $88.3$ & $31.0$ \\

\hline
S8 & generated & $4$ & & designed & GPT-4o & & SD Turbo & $1$ & & \cellcolor{oursrow}SD v2.0 & \ding{52} & & $47.1$ & $70.8$ & $48.8$ & $17.8$ & $54.6$ & $75.4$ & $15.2$ \\ 
S9 & generated & $4$ & & designed & GPT-4o & & SD Turbo & $1$ & & ICLight & \cellcolor{oursrow}\ding{55} & & $51.5$ & $75.1$ & $49.9$ & $19.1$ & $57.4$ & $82.9$ & $24.7$ \\

\midrule
7 & generated & $4$ & & designed & GPT-4o & & SD Turbo & $1$ & & ICLight & \ding{52} & & $52.7$ & $75.1$ & $48.6$ & $18.7$ & $55.6$ & $87.5$ & $30.6$ \\

\bottomrule
\end{tabular}
}
\label{tab:ablation}
\vspace{5pt}
\end{table*}

\vspace{-4pt}
\subsection{Ablations and more results}

\paragraph{Backbones}
In \cref{tab:backbone_results} we present results for fine-tuning two additional backbones. Performance improvements are similar to those of SigLIP, demonstrating the general applicability of our method.
\vspace{-4pt}

\paragraph{Different loss function}
We train SigLIP using infoNCE loss~\citep{chen2020simple}, contrastive loss~\citep{chl05}, and softmax margin loss~\citep{wwz+18}, which are widely used in representation learning.
The generated training set is shown to be effective with a diverse set of losses, while the recall@k loss remains the best overall choice.
Implementation details for this experiment are as follows. For infoNCE loss, we follow the same strategy for batch construction and training parameters as for the recall@k loss. Each image acts as one query, and its positive and negative pairs depend on the instance-level class label. We use a 0.05 temperature during training.
Regarding the contrastive loss, we treat each image as a query and randomly sample one positive among images with the same instance-level class label. For the negatives, we mine the hardest one in the dataset based on the cosine similarity of the image descriptors. We use a margin of $1$. The learning rate is $10^{-7}$, and the batch size is $8$.
For softmax margin loss, we follow the training process proposed in UnED \citep{ypsilantis2021met}, the backbone remains frozen for the first two epochs, and only the classifier is trained with a learning rate of $10^{-3}$. In the following epochs, the network is trained end-to-end with a $10^{-6}$ learning rate. Since this is a classification loss, no specific curation of the batches is necessary. We use a batch size of $16$. The scale and margin parameters are set at $16$ and $0$, respectively, as in UnED.
The results are present in \cref{tab:loss_results}.

\begin{figure}[!h]
\centering
\small
\newcommand{\topone}[2]{\includegraphics[width=50pt,height=50pt]{fig/contamination/ilgen_#1_top0_ilgen.jpg}&\includegraphics[width=50pt,height=50pt]{fig/contamination/ilgen_#1_top0_#1.jpg}}
\newcommand{\toptwo}[2]{\includegraphics[width=50pt,height=50pt]{fig/contamination/ilgen_#1_top1_ilgen.jpg}&\includegraphics[width=50pt,height=50pt]{fig/contamination/ilgen_#1_top1_#1.jpg}}

\begin{tabular}{@{\hspace{0pt}}c@{\hspace{2pt}}c@{\hspace{2pt}} c@{\hspace{13pt}} c@{\hspace{2pt}}c@{\hspace{2pt}} c@{\hspace{13pt}} c@{\hspace{2pt}}c@{\hspace{2pt}} c@{\hspace{13pt}} c@{\hspace{2pt}}c@{\hspace{2pt}}}

\multicolumn{2}{c}{similarity: $0.87$} & & \multicolumn{2}{c}{similarity: $0.83$} & & \multicolumn{2}{c}{similarity: $0.79$} & & \multicolumn{2}{c}{similarity: $0.77$}\\
\topone{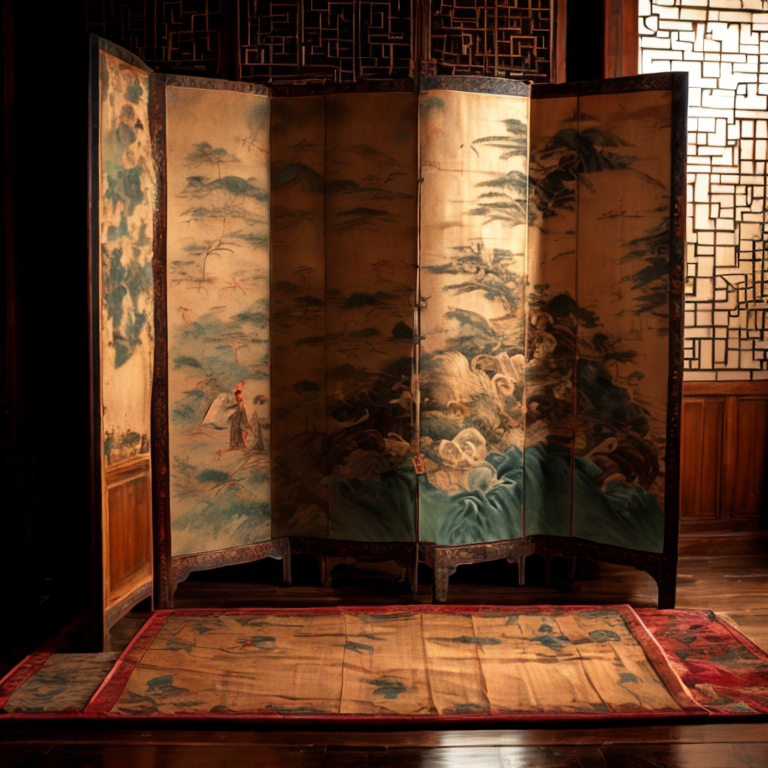}{} & & \topone{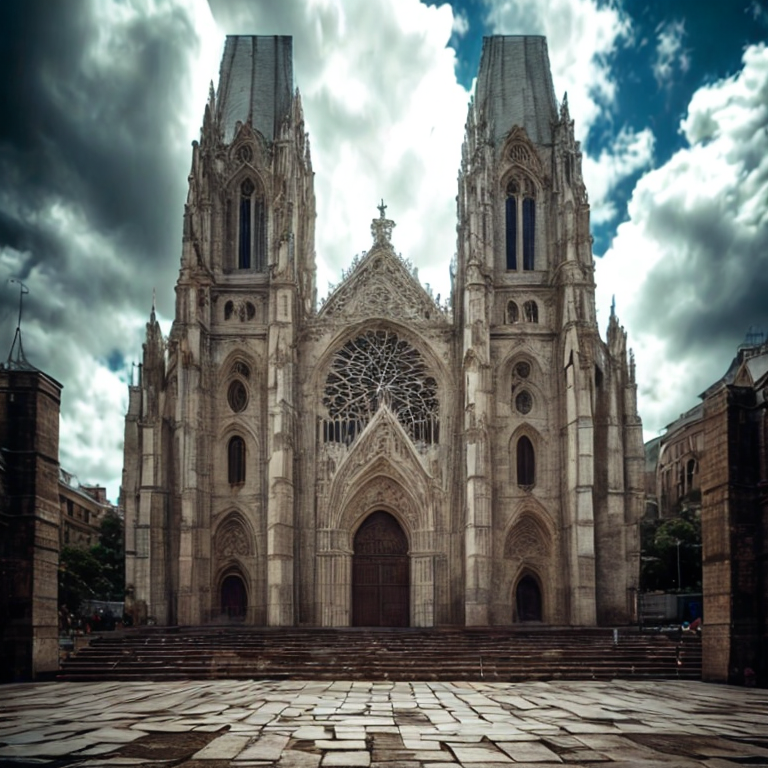}{} & & \topone{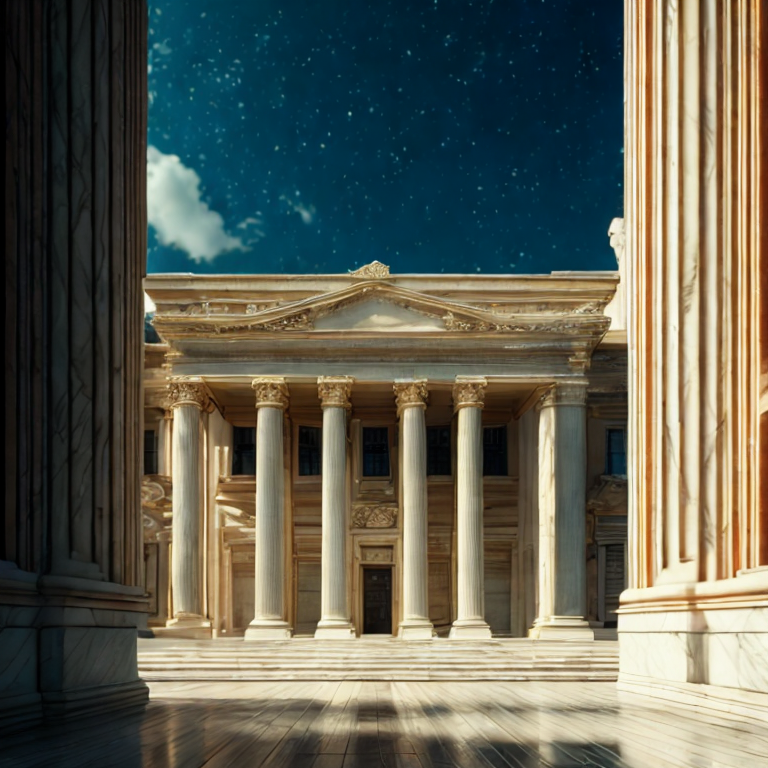}{} & & \topone{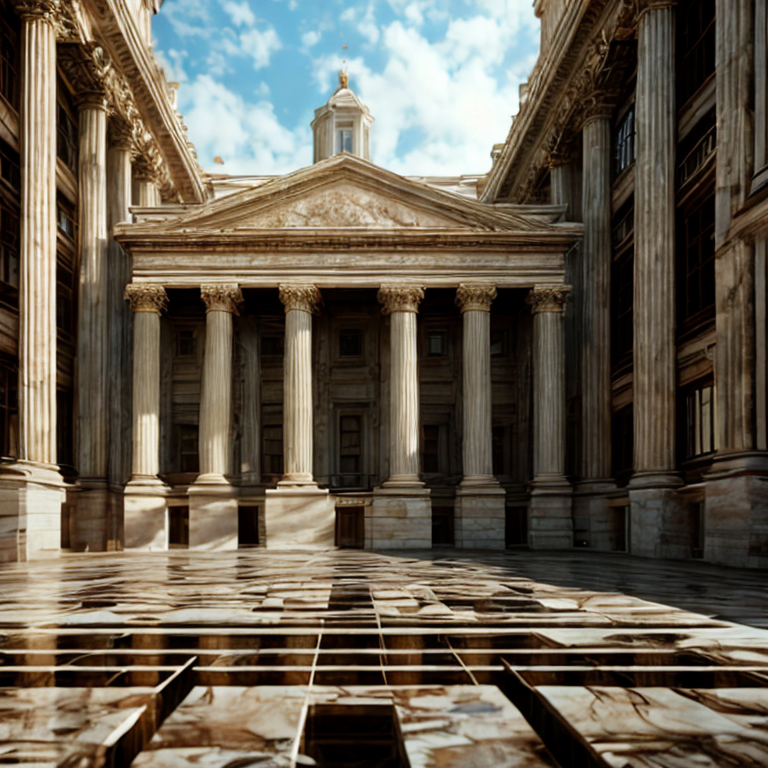}{}\\
\oursplus{} & MET & & \oursplus{} & GLDv2 & & \oursplus{} & R-Oxford & & \oursplus{} & R-Paris \\[4pt]

\multicolumn{2}{c}{similarity: $0.73$} & & \multicolumn{2}{c}{similarity: $0.82$} & & \multicolumn{2}{c}{similarity: $0.74$} & & \multicolumn{2}{c}{similarity: $0.73$} \\
\topone{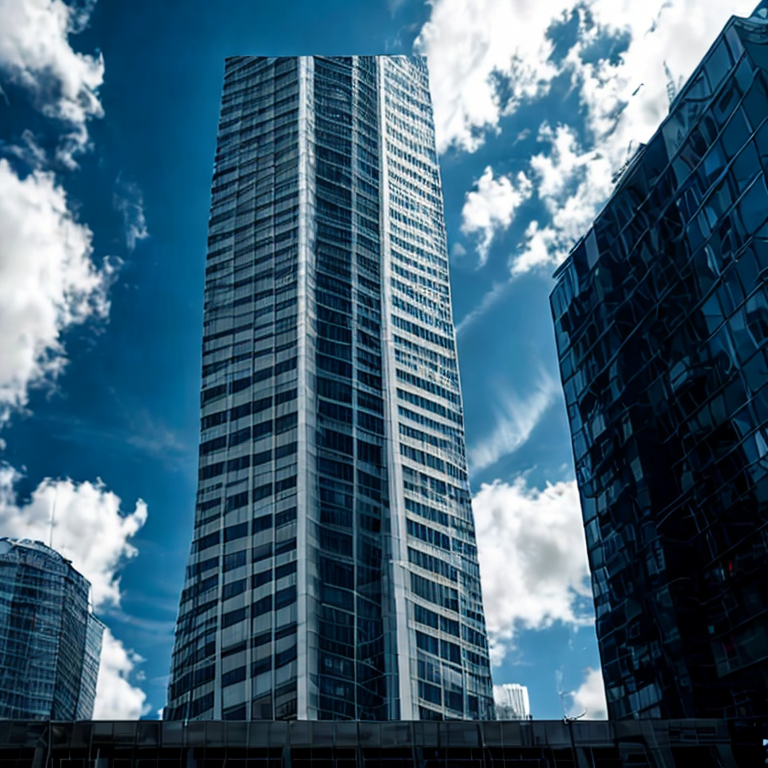}{} & & \topone{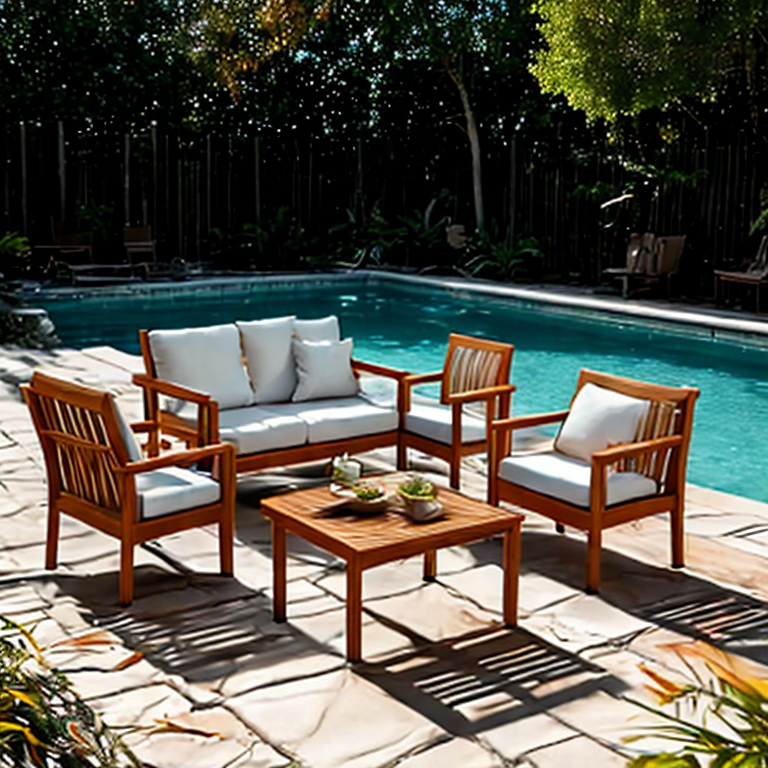}{} & & \topone{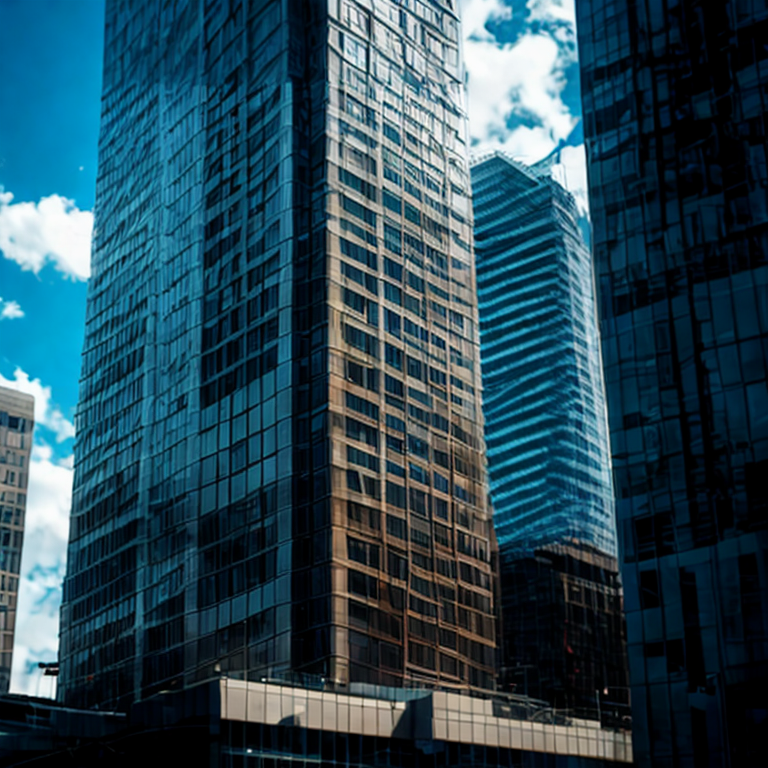}{} & & \toptwo{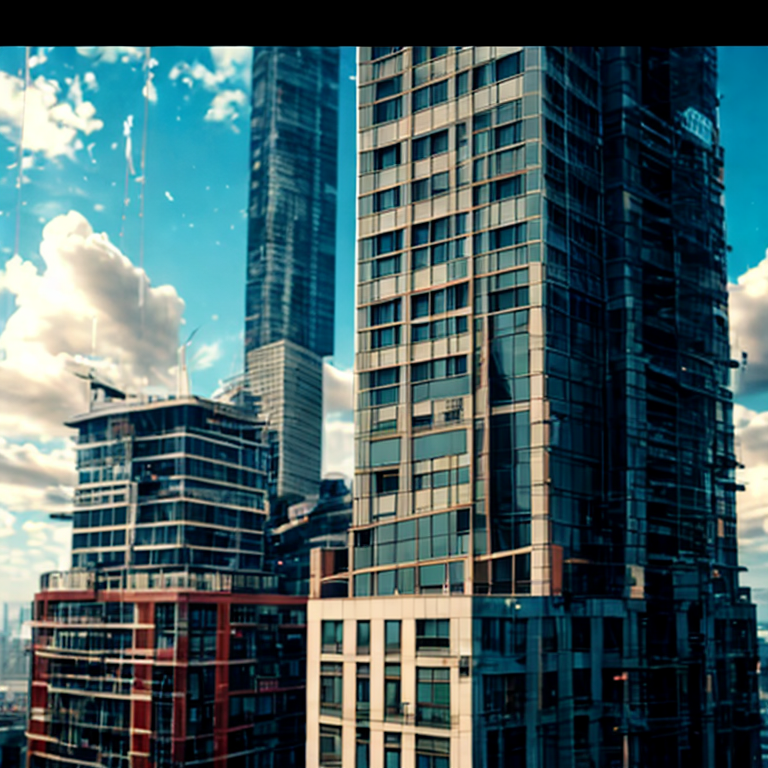}{}\\
\oursplus{} & INSTRE & & \oursplus{} & SOP & & \oursplus{} & mini-ILIAS & & \oursplus{} & mini-ILIAS \\[4pt]

\end{tabular}

\caption{Pairs of \oursplus{} and test sets with the highest similarity score. While these pairs share some common appearance, they do not indicate data leakage from an ILR point of view.}
 \label{fig:contamination}
\end{figure}

\paragraph{Training images per class}
\cref{tab:ablation} shows the performance with different numbers of images per instance-level class during training (ID-S1 and ID-S2). We decrease the number of images per class $N$ in the training set to $3$ and $2$. The trained models achieve an average performance of $51.5$ and $50.3$, respectively, which is a considerable drop compared to the main variant that achieves $52.7$. 
\vspace{-4pt}

\paragraph{LLM models and prompts}
To examine the effect of the prompts and LLMs, we evaluate variants from ID-S3 to ID-S5 in \cref{tab:ablation}. In ID-S3, we use a fixed prompt template across all the generic and specific domains with GPT-4o (see the supplementary material). 
In ID-S4 and ID-S5, we use our designed prompts with two other LLMs, DeepSeek-V3 and Claude 3.7 Sonnet, respectively. The similar results suggest that our method is robust regardless of the LLM or prompt type.
\vspace{-4pt}

\paragraph{GDM}
We apply different GDMs and higher-quality images to study how instance generation quality affects performance, as shown in ID-S6 and ID-S7 in \cref{tab:ablation}. In ID-S6, we change SD Turbo to SD v2.0, resulting in worse performance, likely due to more intricate backgrounds that hinder accurate foreground segmentation. We use $50$ inference steps following the default setting. 
In ID-S7, we increase the inference steps of SD Turbo from the default $1$ to $5$, aiming to generate higher-quality images. Although the visual quality is better, there was no overall significant performance improvement. Additional details are in the supplementary material.

\paragraph{Background generation}
As shown in \cref{tab:ablation}, changing ICLight to SD v2.0 for background generation (ID-S8) leads to worse performance even than the pretrained model (ID1). This is due to poor identity preservation, while ICLight is tailored to this task. When we switch off padding (ID-S9), which is our way of varying object size and position, the average performance drops by 1.2, demonstrating that even such a simple viewpoint variation has a positive impact.

\begin{figure}[t]
\centering
\small
\newcommand{\topone}[2]{\includegraphics[width=50pt,height=50pt]{fig/suppl_sscd/ilgen_#1_top0_ilgen.jpg}&\includegraphics[width=50pt,height=50pt]{fig/suppl_sscd/ilgen_#1_top0_#1.jpg}}
\newcommand{\toptwo}[2]{\includegraphics[width=50pt,height=50pt]{fig/suppl_sscd/ilgen_#1_top1_ilgen.jpg}&\includegraphics[width=50pt,height=50pt]{fig/suppl_sscd/ilgen_#1_top1_#1.jpg}}

\begin{tabular}{@{\hspace{0pt}}c@{\hspace{2pt}}c@{\hspace{2pt}} c@{\hspace{13pt}} c@{\hspace{2pt}}c@{\hspace{2pt}} c@{\hspace{13pt}} c@{\hspace{2pt}}c@{\hspace{2pt}} c@{\hspace{13pt}} c@{\hspace{2pt}}c@{\hspace{2pt}}}

\multicolumn{2}{c}{similarity (SSCD): $0.33$} & & \multicolumn{2}{c}{similarity (SSCD): $0.25$} & & \multicolumn{2}{c}{similarity (SSCD): $0.20$} & & \multicolumn{2}{c}{similarity (SSCD): $0.27$}\\
\topone{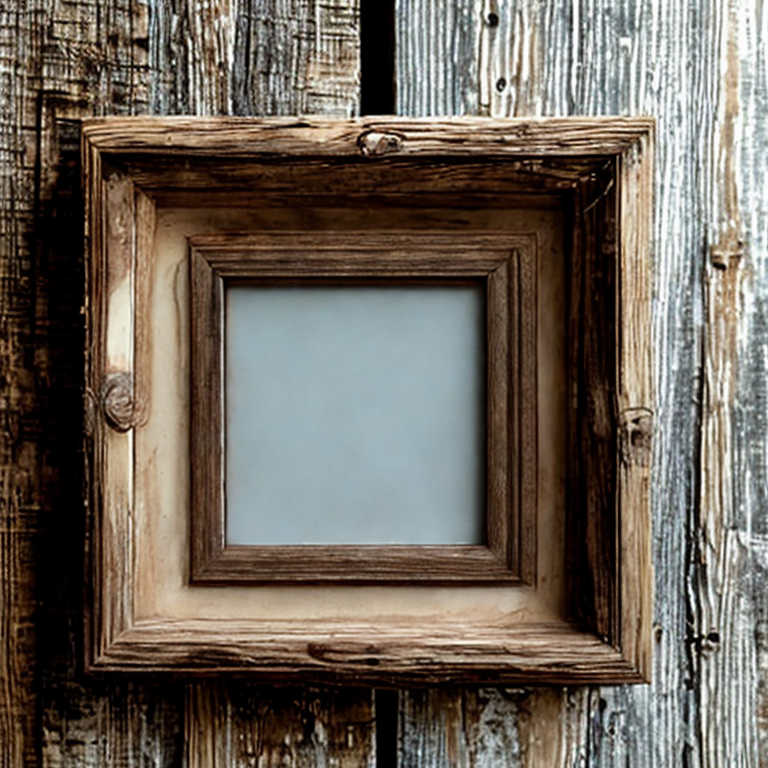}{} & & \topone{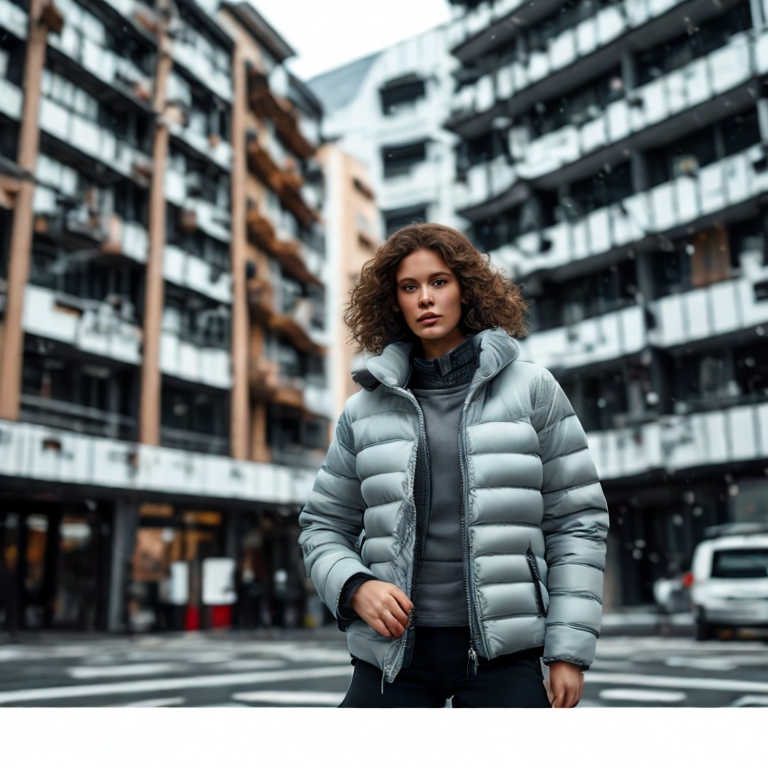}{} & & \topone{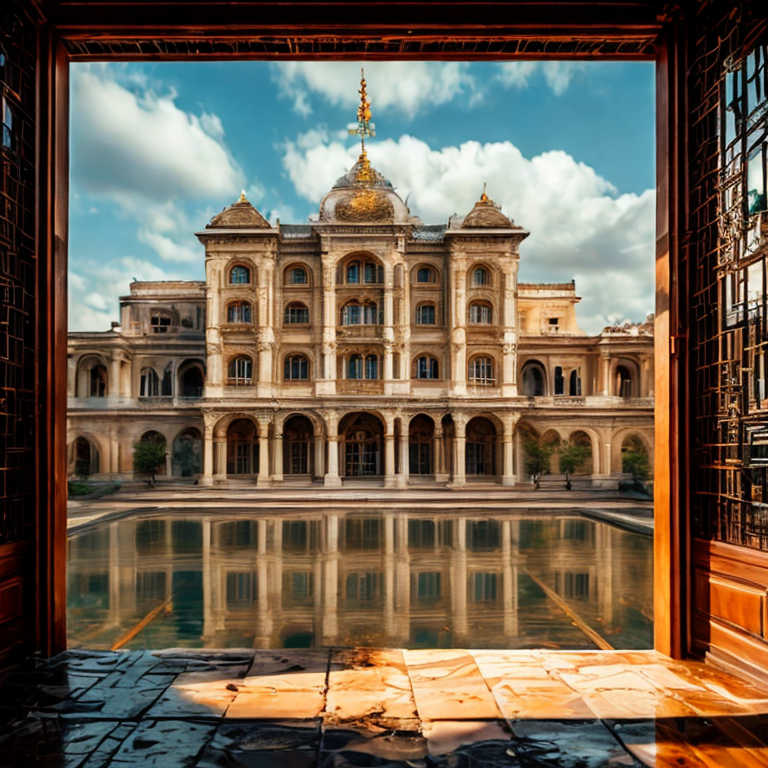}{} & & \topone{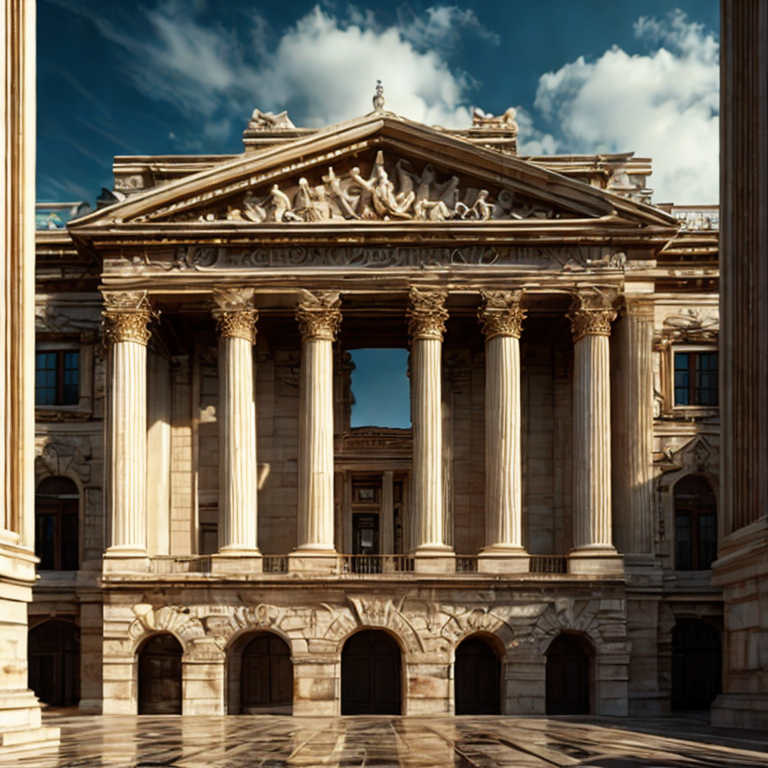}{}\\
\oursplus{} & MET & & \oursplus{} & GLDv2 & & \oursplus{} & R-Oxford & & \oursplus{} & R-Paris \\[4pt]

\multicolumn{2}{c}{similarity (SSCD): $0.23$} & & \multicolumn{2}{c}{similarity (SSCD): $0.38$} & & \multicolumn{2}{c}{similarity (SSCD): $0.29$} & & \multicolumn{2}{c}{similarity (SSCD): $0.28$} \\
\topone{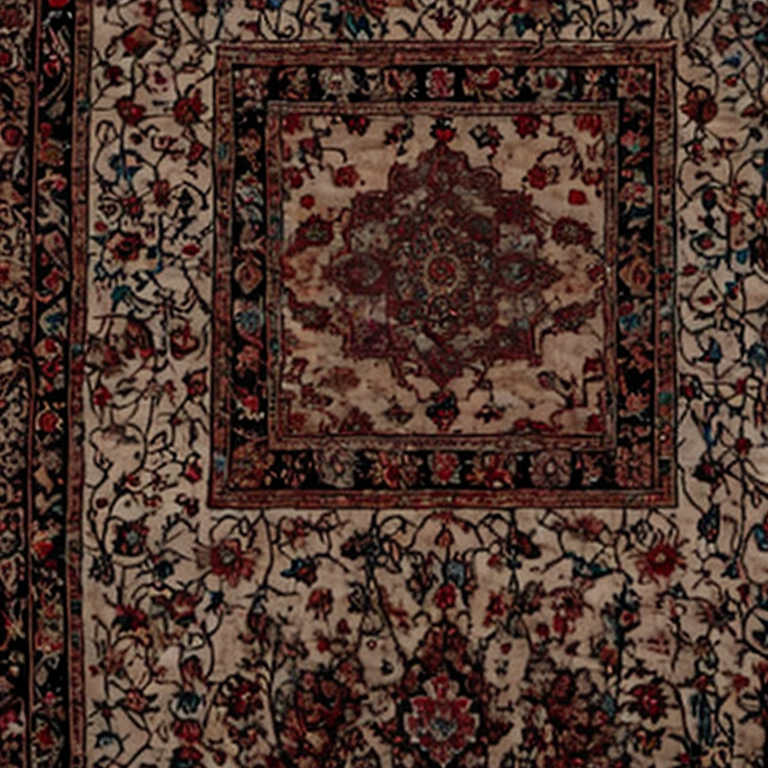}{} & & \topone{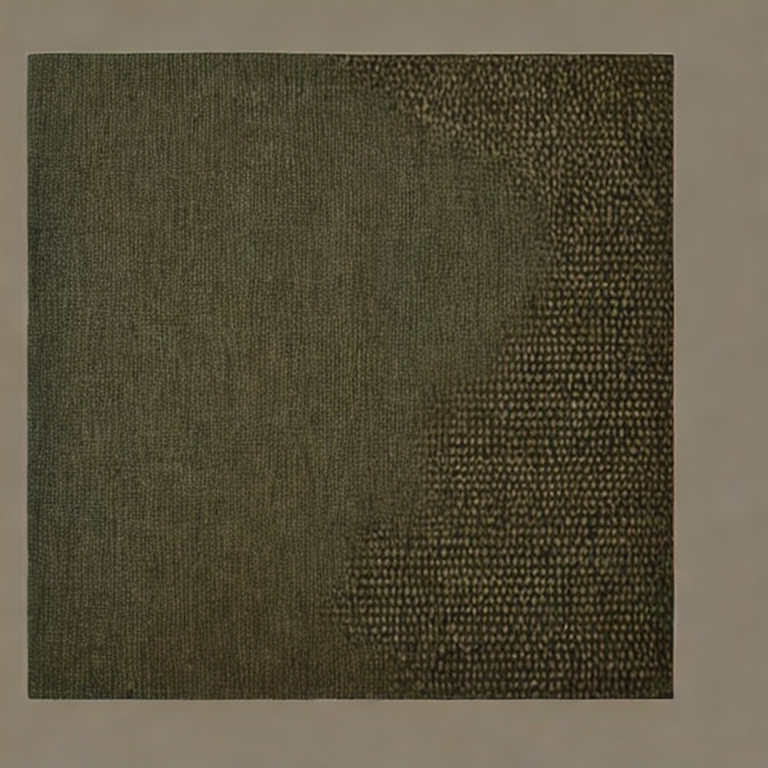}{} & & \topone{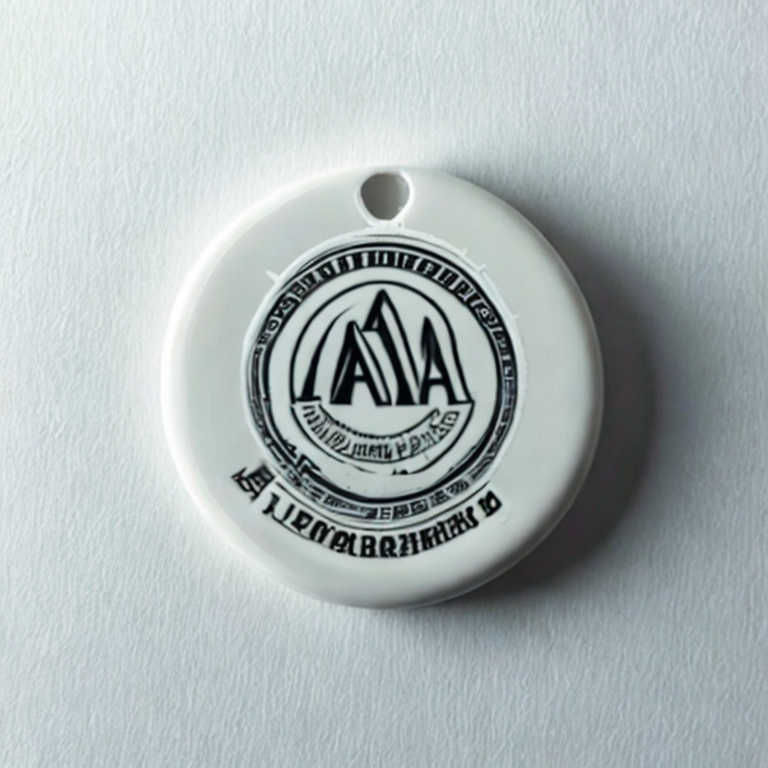}{} & & \toptwo{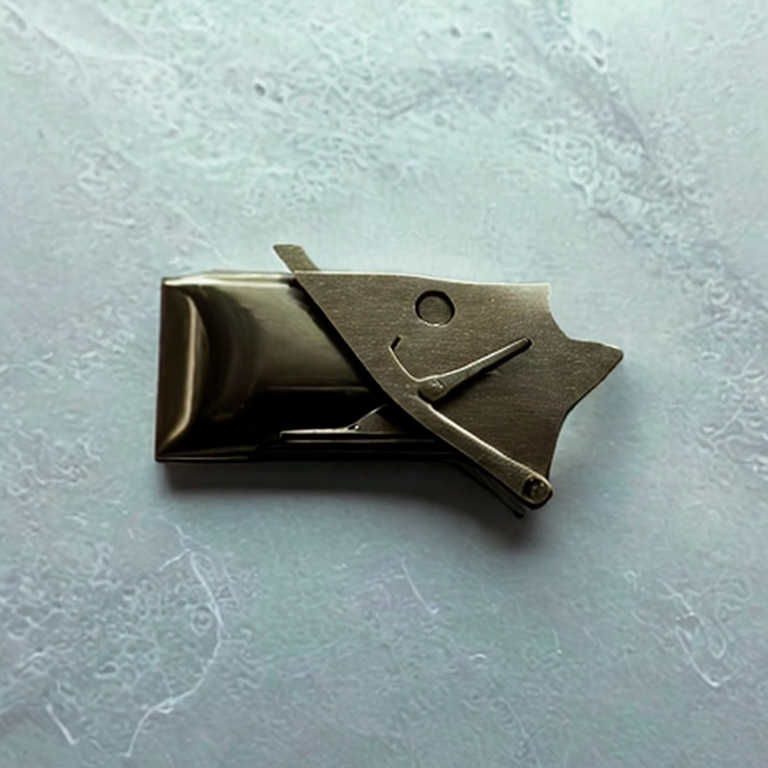}{}\\
\oursplus{} & INSTRE & & \oursplus{} & SOP & & \oursplus{} & mini-ILIAS & & \oursplus{} & mini-ILIAS \\[4pt]

\end{tabular}

\hspace{30pt}
\vspace{-15pt}
\caption{Pairs of \oursplus{} and test sets with the highest similarity score computed using SSCD. The low similarities indicate no copying between the generated and real images.}
 \label{fig:suppl_contamination_sscd}
 \vspace{-8pt}
\end{figure}

\paragraph{Train and test set overlap}
To investigate whether objects from the test sets have leaked into the generated training set, we perform the following mining process. We first use the trained model (ID7) as a descriptor extractor and perform retrieval using the test queries as queries and the generated training set as the database. We visually inspect the results with the highest similarity scores and do not identify any cases of such leakage as shown in \cref{fig:contamination}. The pairs showcase similar characteristics (a strength of our approach), but are not positive from an instance-level point of view.
Moreover, we further use SSCD\footnote{\url{https://github.com/facebookresearch/sscd-copy-detection}} \citep{pizzi2022self}, specifically the \texttt{sscd\_disc\_large} (ResNeXt101) model for copy detection. 
\cref{fig:suppl_contamination_sscd} shows the pairs with highest similarities between \oursplus{} and test datasets. The highest similarity score is 0.38, which is strictly below the copy-detection threshold of 0.5 following \citet{somepalli2023diffusion}, indicating that none of the generated images are copies of the test images.

\section{Conclusion}
This work introduces a novel approach to training ILR models using generative diffusion models to automatically create diverse, instance-specific training images. By eliminating the need for extensive data collection and curation, our method opens up new opportunities to easily train ILR models across various domains. Although foundational representation models are generally considered universal and capable of performing well across a wide range of domains, we show that fine-tuning these models exclusively on synthetic instance-level data results in notable performance improvements. \looseness=-1

\subsubsection*{Acknowledgments}
This work was partly supported by Junior Star GACR GM 21-28830M, Horizon MSCA-PF grant No. 101154126, and JSPS KAKENHI No. 22K12091 and No. 23H00497. We are grateful to Patrick Ramos, Ryan Ramos, and Lu Wei for their time and effort in the human verification. We also thank Pavel Suma for the support on the code.

\bibliography{main}
\bibliographystyle{tmlr}

\appendix
\clearpage
\renewcommand{\thetable}{S\arabic{table}}  
\renewcommand{\thefigure}{S\arabic{figure}}
\setcounter{table}{0}
\setcounter{figure}{0}
\setcounter{page}{1}
\appendix

$\Large{\textbf{Supplementary material}}$
\bigskip

This supplementary material reports further experimental details discussed in the main paper. The document is organized into the following sections:
\begin{itemize}[noitemsep,topsep=0pt]
    \item Section \ref{sec:sup:ethical_consideration}: Ethical considerations.
    \item Section \ref{sec:sup:seed}: Instance identity and diversity.
    \item Section \ref{sec:sup:data_generation}: Data generation.
    \item Section \ref{sec:sup:backbone_details}: Backbone details.
    \item Section \ref{sec:sup:additional_study}: Additional study.
    \item Section \ref{sec:sup:dataset_samples}: Evaluation dataset examples.
\end{itemize}

\section{Ethical considerations}
\label{sec:sup:ethical_consideration}
\subsection{Provenance and licensing of generated data}
Our pipeline relies on pretrained LLMs and text-to-image generative models, both of which are trained on uncurated large-scale datasets that may include copyrighted, trademarked, or restricted content. As a consequence, synthetic images produced for domains such as artworks, landmarks, or products may inherit visual patterns from the training data, \eg, products associated with particular brands. We acknowledge this provenance limitation explicitly and encourage downstream users to (1) verify that their usage complies with the licenses of the generative models, and (2) avoid redistributing generated images with products with recognized brands, artworks, or landmarks. We will release our code under the restrictive non-commercial research license with an explicit prohibition on redistribution as training data for commercial image-generation services.

\subsection{Dual-use considerations}
Instance-level retrieval is inherently dual-use. While it can help museum visitors look up a painting or customers find a specific product, it could also be used for surveillance by tracking distinctive personal belongings. To reduce the risk of such misuse, we target domains that match those of established ILR benchmarks. Therefore, our contribution improves performance on well-established, non-human-centric categories rather than unlocking a new high-risk capability. Second, our prompts do not target faces, license plates, or other personally identifiable attributes. The CC-BY-NC-SA license forbids applications aimed at identifying or tracking individuals, as well as biometric and privacy-invasive use cases. Downstream works must comply with these restrictions, and we will reserve the right to revoke access for violators while maintaining a channel in the repository for reporting misuse. Our aim is to exert every reasonable control available to the users of the training dataset and code while preserving the benefits for research.

\subsection{Inherited biases}
Since our pipeline relies on pretrained LLMs and diffusion models, it inherits the cultural, geographical, and stylistic biases of those models. For example, the prompt ``church'' may produce churches that conflate denominations, and generated products tend to be Western-oriented. These imbalances may lead to discrepancies in retrieval performance across sub-domains and reinforce stereotypes in downstream tasks. Furthermore, although we refrain from using prompts that mention humans or human-related concepts, it is difficult to fully prevent the model from producing images that contain faces. When deploying our pipeline, users should be aware of these risks and apply the generated data to downstream tasks with caution.

\section{Instance identity and diversity}
\label{sec:sup:seed}

We study whether the generated instances are sufficiently distinct and examine the effect of instance diversity. We (1) present qualitative visualizations of GDM outputs per domain; (2) conduct a human verification study in which participants identify the image originating from the same instance after background and lighting variations, validating identity preservation; (3) compute within-class pairwise similarity and remove one of the top 500 near-duplicate pairs for training; and (4) improve instance diversity by enriching the prompts with specific feature descriptions. Finally, we also present failure cases in which instance identity is not preserved well during generation. Overall, the results indicate that generated instances are sufficiently distinguishable and that removing near-duplicate instances and increasing instance diversity yield only marginal performance changes, demonstrating the robustness of our method. Detailed analysis is presented below.

\begin{figure*}[t]
\footnotesize

\newcommand{\generic}[1]{\includegraphics[width=40pt,height=40pt]{fig/seeds/#1_rd_0.jpg}&\includegraphics[width=40pt,height=40pt]{fig/seeds/#1_rd_1.jpg}&\includegraphics[width=40pt,height=40pt]{fig/seeds/#1_rd_2.jpg}&\includegraphics[width=40pt,height=40pt]{fig/seeds/#1_rd_3.jpg}&\includegraphics[width=40pt,height=40pt]{fig/seeds/#1_rd_4.jpg}&\includegraphics[width=40pt,height=40pt]{fig/seeds/#1_rd_5.jpg}&\includegraphics[width=40pt,height=40pt]{fig/seeds/#1_rd_6.jpg}&\includegraphics[width=40pt,height=40pt]{fig/seeds/#1_rd_7.jpg}&\includegraphics[width=40pt,height=40pt]{fig/seeds/#1_rd_8.jpg}&\includegraphics[width=40pt,height=40pt]{fig/seeds/#1_rd_9.jpg}}

\begin{tabular}{@{} r@{\hspace{2pt}} c@{\hspace{2pt}}c@{\hspace{2pt}}c@{\hspace{2pt}}c@{\hspace{2pt}}c@{\hspace{2pt}}c@{\hspace{2pt}}c@{\hspace{2pt}}c@{\hspace{2pt}}c@{\hspace{2pt}}c@{\hspace{2pt}}@{}}

\raisebox{18pt}{glove} & \generic{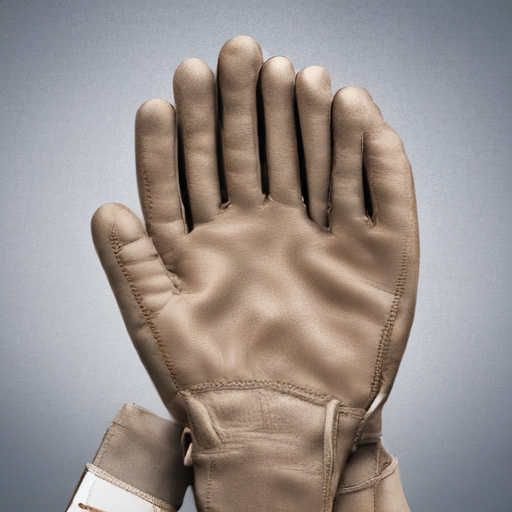} \\
\raisebox{18pt}{table} & \generic{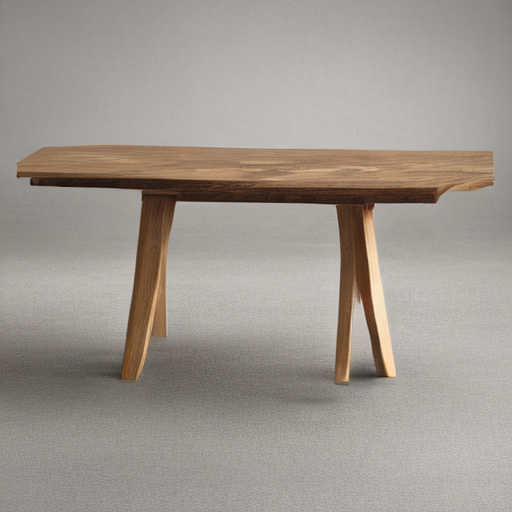} \\
\raisebox{18pt}{chair} & \generic{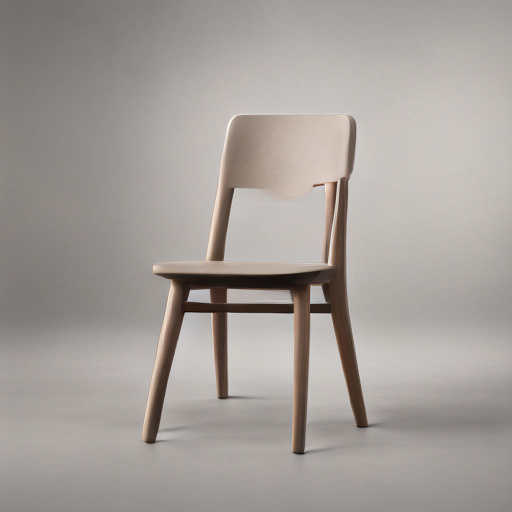} \\
\raisebox{18pt}{hat} & \generic{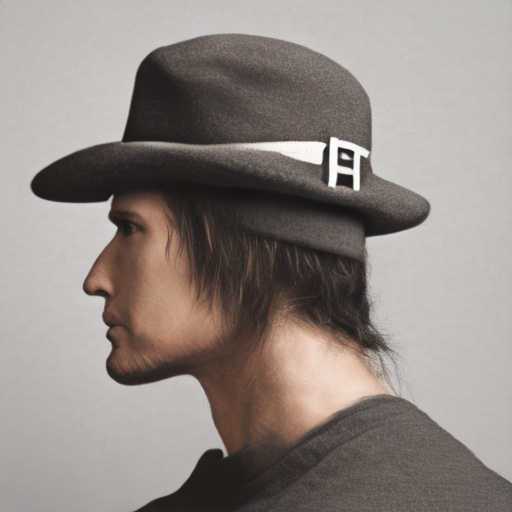} \\

\end{tabular}
\caption{Examples of 10 instances per object generated with different random seeds in the generic domain.}
\label{fig:sup:seed_generic}
\end{figure*}

\begin{figure*}[t]
\footnotesize

\newcommand{\artwork}[1]{\includegraphics[width=28pt,height=28pt]{fig/seeds/#1_rd_0.jpg}&\includegraphics[width=28pt,height=28pt]{fig/seeds/#1_rd_1.jpg}&\includegraphics[width=28pt,height=28pt]{fig/seeds/#1_rd_2.jpg}&\includegraphics[width=28pt,height=28pt]{fig/seeds/#1_rd_3.jpg}&\includegraphics[width=28pt,height=28pt]{fig/seeds/#1_rd_4.jpg}&\includegraphics[width=28pt,height=28pt]{fig/seeds/#1_rd_5.jpg}&\includegraphics[width=28pt,height=28pt]{fig/seeds/#1_rd_6.jpg}&\includegraphics[width=28pt,height=28pt]{fig/seeds/#1_rd_7.jpg}&\includegraphics[width=28pt,height=28pt]{fig/seeds/#1_rd_8.jpg}&\includegraphics[width=28pt,height=28pt]{fig/seeds/#1_rd_9.jpg}&\includegraphics[width=28pt,height=28pt]{fig/seeds/#1_rd_10.jpg}&\includegraphics[width=28pt,height=28pt]{fig/seeds/#1_rd_11.jpg}&\includegraphics[width=28pt,height=28pt]{fig/seeds/#1_rd_12.jpg}&\includegraphics[width=28pt,height=28pt]{fig/seeds/#1_rd_13.jpg}&\includegraphics[width=28pt,height=28pt]{fig/seeds/#1_rd_14.jpg}}

\scalebox{1}{\begin{tabular}{@{} c@{\hspace{2pt}}c@{\hspace{2pt}}c@{\hspace{2pt}}c@{\hspace{2pt}}c@{\hspace{2pt}}c@{\hspace{2pt}}c@{\hspace{2pt}}c@{\hspace{2pt}}c@{\hspace{2pt}}c@{\hspace{2pt}}c@{\hspace{2pt}}c@{\hspace{2pt}}c@{\hspace{2pt}}c@{\hspace{2pt}}c@{\hspace{2pt}}@{}}

\multicolumn{15}{l}{\hspace{-6pt}Byzantine icon painting} \\
\artwork{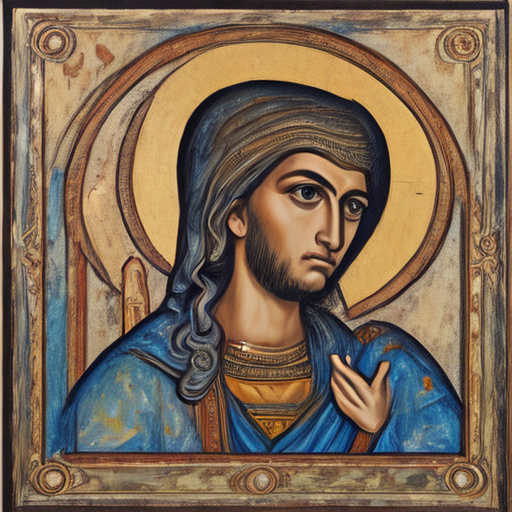} \\
\multicolumn{15}{l}{\hspace{-6pt}Greek krater vase} \\
\artwork{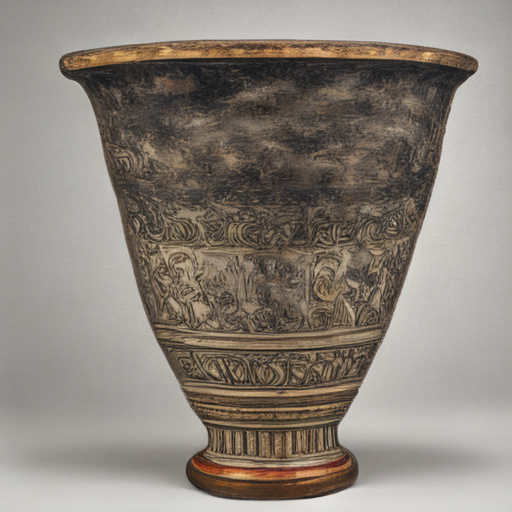} \\
\multicolumn{15}{l}{\hspace{-6pt}Roman gladiator helmet} \\
\artwork{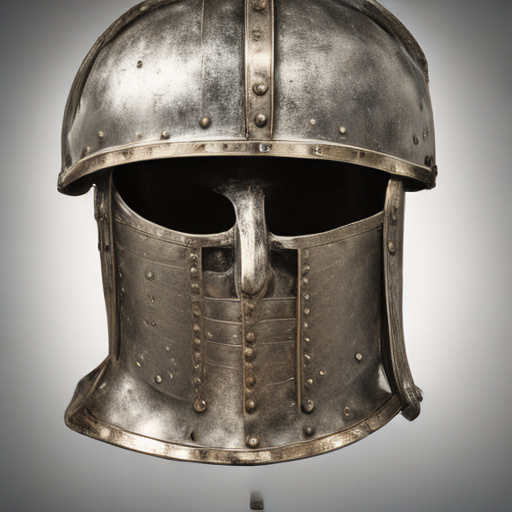} \\
\multicolumn{15}{l}{\hspace{-6pt}Celtic gold bracelet} \\
\artwork{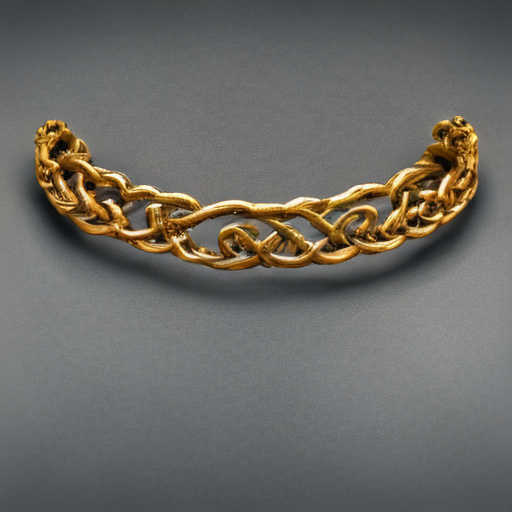} \\

\end{tabular}}
\caption{Examples of 15 instances per object generated with different random seeds in the artwork domain.}
\label{fig:sup:seed_art}
\end{figure*}

\paragraph{Visualization of GDM-generated instances}
To study the effect of random seeds on image generation, 
we select four object names in each of the four domains and generate several instances per object with different random seeds using GDM and present examples in \cref{fig:sup:seed_generic} (generic), \cref{fig:sup:seed_art} (artwork), \cref{fig:sup:seed_landmark} (landmark), and \cref{fig:sup:seed_product} (product). We observe that while certain instances may appear visually similar at first glance, they remain distinguishable through fine-grained details.

\begin{figure*}[t]
\footnotesize

\newcommand{\seedimg}[2]{%
  \includegraphics[width=28pt,height=28pt]{fig/seeds/#1_rd_#2.jpg}%
}

\newcommand{\seedrow}[2]{%
  \seedimg{#1}{\number\numexpr#2\relax} &
  \seedimg{#1}{\number\numexpr#2+1\relax} &
  \seedimg{#1}{\number\numexpr#2+2\relax} &
  \seedimg{#1}{\number\numexpr#2+3\relax} &
  \seedimg{#1}{\number\numexpr#2+4\relax} &
  \seedimg{#1}{\number\numexpr#2+5\relax} &
  \seedimg{#1}{\number\numexpr#2+6\relax} &
  \seedimg{#1}{\number\numexpr#2+7\relax} &
  \seedimg{#1}{\number\numexpr#2+8\relax} &
  \seedimg{#1}{\number\numexpr#2+9\relax} &
  \seedimg{#1}{\number\numexpr#2+10\relax} &
  \seedimg{#1}{\number\numexpr#2+11\relax} &
  \seedimg{#1}{\number\numexpr#2+12\relax} &
  \seedimg{#1}{\number\numexpr#2+13\relax} &
  \seedimg{#1}{\number\numexpr#2+14\relax} \\
}

\begin{tabular}{@{}*{15}{c@{\hspace{2pt}}}@{}}
  \multicolumn{15}{l}{\hspace{-6pt}Bridge} \\
  \seedrow{bridge}{0}
  \multicolumn{15}{l}{\hspace{-6pt}Clock tower} \\
  \seedrow{Clock_tower}{0}
  \multicolumn{15}{l}{\hspace{-6pt}Temple} \\
  \seedrow{Temple}{0}
  \multicolumn{15}{l}{\hspace{-6pt}Skyscraper headset} \\
  \seedrow{Skyscraper}{0}
\end{tabular}
\caption{Examples of 15 instances per object generated with different random seeds in the landmark domain.}
\label{fig:sup:seed_landmark}
\end{figure*}

\begin{figure*}[t]
\footnotesize

\newcommand{\product}[1]{\includegraphics[width=28pt,height=28pt]{fig/seeds/#1_rd_0.jpg}&\includegraphics[width=28pt,height=28pt]{fig/seeds/#1_rd_1.jpg}&\includegraphics[width=28pt,height=28pt]{fig/seeds/#1_rd_2.jpg}&\includegraphics[width=28pt,height=28pt]{fig/seeds/#1_rd_3.jpg}&\includegraphics[width=28pt,height=28pt]{fig/seeds/#1_rd_4.jpg}&\includegraphics[width=28pt,height=28pt]{fig/seeds/#1_rd_5.jpg}&\includegraphics[width=28pt,height=28pt]{fig/seeds/#1_rd_6.jpg}&\includegraphics[width=28pt,height=28pt]{fig/seeds/#1_rd_7.jpg}&\includegraphics[width=28pt,height=28pt]{fig/seeds/#1_rd_8.jpg}&\includegraphics[width=28pt,height=28pt]{fig/seeds/#1_rd_9.jpg}&\includegraphics[width=28pt,height=28pt]{fig/seeds/#1_rd_10.jpg}&\includegraphics[width=28pt,height=28pt]{fig/seeds/#1_rd_11.jpg}&\includegraphics[width=28pt,height=28pt]{fig/seeds/#1_rd_12.jpg}&\includegraphics[width=28pt,height=28pt]{fig/seeds/#1_rd_13.jpg}&\includegraphics[width=28pt,height=28pt]{fig/seeds/#1_rd_14.jpg}}

\begin{tabular}{@{} c@{\hspace{2pt}}c@{\hspace{2pt}}c@{\hspace{2pt}}c@{\hspace{2pt}}c@{\hspace{2pt}}c@{\hspace{2pt}}c@{\hspace{2pt}}c@{\hspace{2pt}}c@{\hspace{2pt}}c@{\hspace{2pt}}c@{\hspace{2pt}}c@{\hspace{2pt}}c@{\hspace{2pt}}c@{\hspace{2pt}}c@{\hspace{2pt}}@{}}

\multicolumn{15}{l}{\hspace{-6pt}Gaming headset} \\
\product{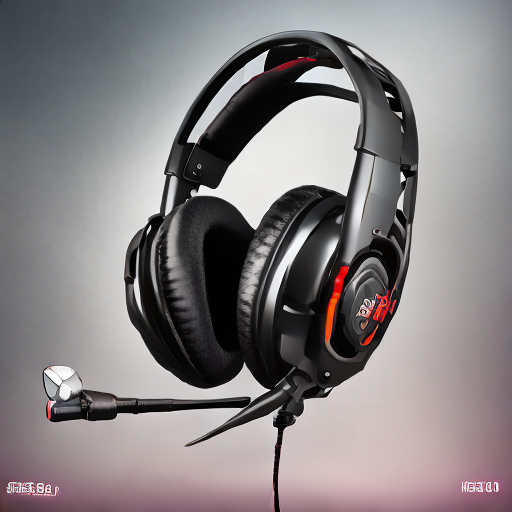} \\
\multicolumn{15}{l}{\hspace{-6pt}Designer scarf} \\
\product{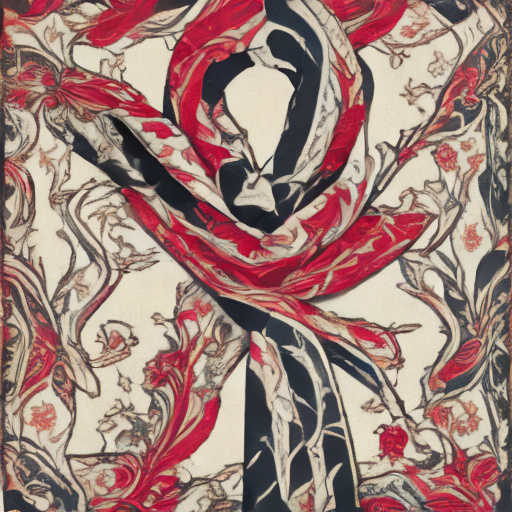} \\
\multicolumn{15}{l}{\hspace{-6pt}Sneakers} \\
\product{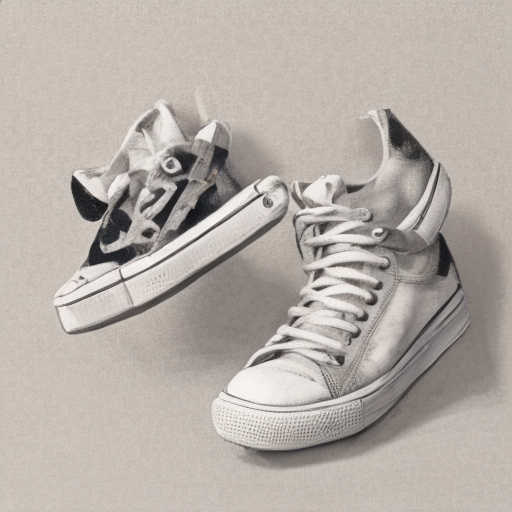} \\
\multicolumn{15}{l}{\hspace{-6pt}Camping tent} \\
\product{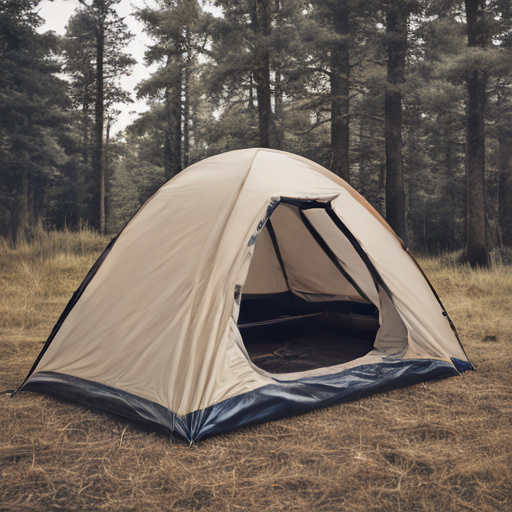} \\

\end{tabular}
\caption{Examples of 15 instances per object generated with different random seeds in the product domain.}
\label{fig:sup:seed_product}
\end{figure*}

\begin{figure*}[t]
\centering
\includegraphics[width=0.87\linewidth]{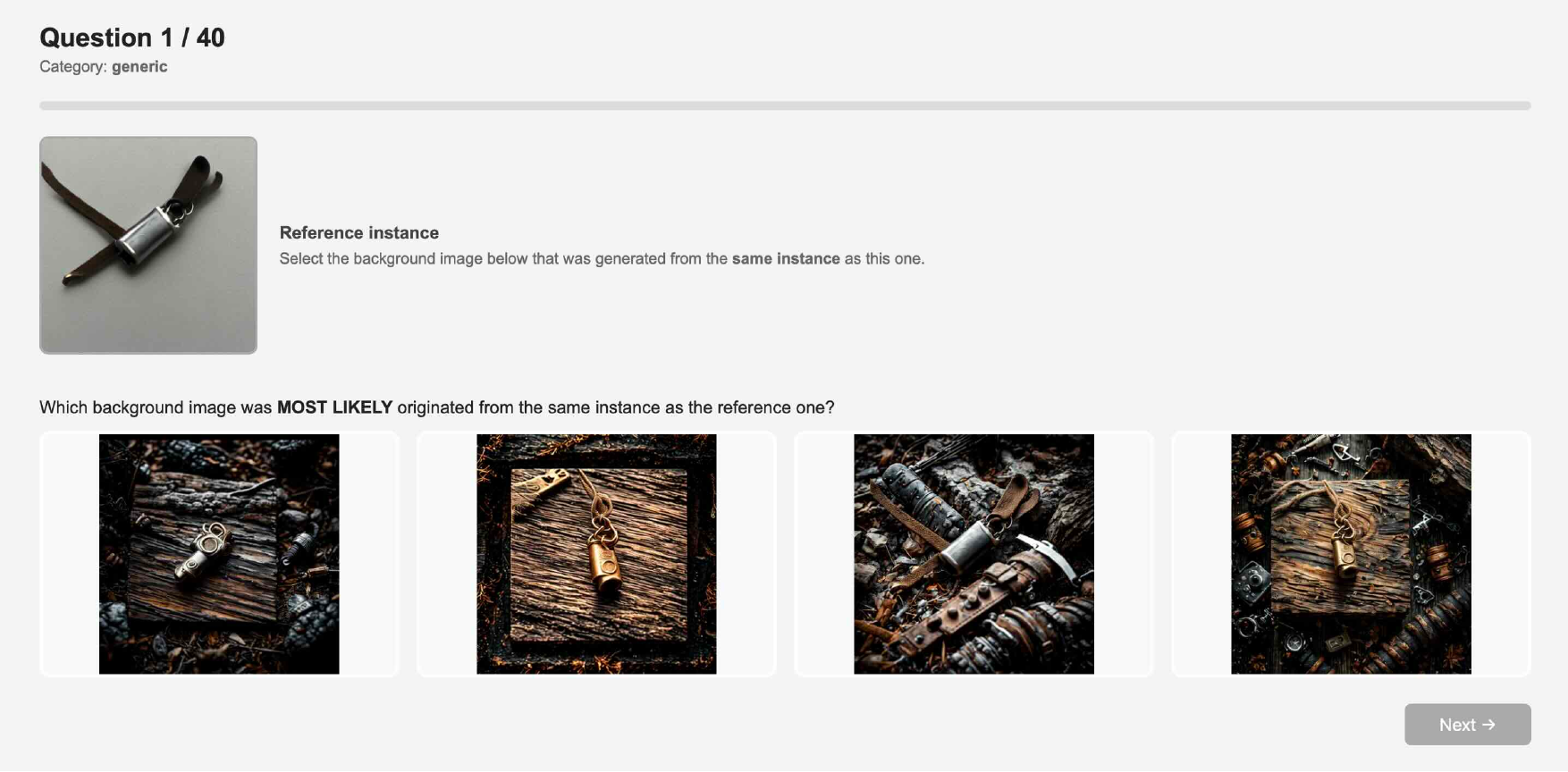}
\caption{The interface of the human verification study.}
\label{fig:sup:example_human_verification}
\end{figure*}

\paragraph{Human verification}
To verify that instance identity is distinguishable and preserved after background and lighting variations, we conduct a human verification study. Each question presents a reference image of an instance generated by GDM, along with four candidate images with background and lighting variations from the same object class, among which one originates from the reference image. The participants are asked to select the candidate image that most likely shares the same identity as the reference. The interface is shown in \cref{fig:sup:example_human_verification}. For each domain, we randomly select 10 instances, resulting in 40 questions in total. Four computer vision researchers participate in the study. The average accuracy of 96.25\% demonstrates that, according to human annotators, generated instances are distinct and retain consistent identity after applying background variation.

\paragraph{Remove near-duplicate}
To measure the impact of very similar instances in the training, we use the trained model (ID7) to compute similarities for all image pairs within each object, and rank pairs across all objects jointly. Top similar pairs ranked from 1st to 20000th are shown in \cref{fig:suppl_seed_similar_pairs}. Pairs within roughly the top 500 are visually very similar, whereas pairs beyond this rank tend to be visually distinguishable. We therefore remove one image from each of the top 500 most similar pairs for training. Note that 500 pairs is a tiny fraction of all 245k pairs (0.02\%). Then, we repeat the training with the slightly reduced amount of data (ID-S12, \cref{tab:sup:seeds}). 
Average performance improves slightly by 0.2, suggesting that near-duplicate instances have only an insignificant negative effect on overall performance.

\begin{figure}[t]
\centering
\small
\newcommand{\toppair}[2]{\includegraphics[width=50pt,height=50pt]{fig/suppl_rank_images/rank_#1_A_#2.jpg}&\includegraphics[width=50pt,height=50pt]{fig/suppl_rank_images/rank_#1_B_#2.jpg}}

\scalebox{0.98}{
\begin{tabular}{c@{\hspace{2pt}}c@{\hspace{2pt}} c@{\hspace{5pt}} c@{\hspace{2pt}}c@{\hspace{2pt}} c@{\hspace{5pt}} c@{\hspace{2pt}}c@{\hspace{2pt}} c@{\hspace{5pt}} c@{\hspace{2pt}}c@{\hspace{2pt}}}

\multicolumn{2}{c}{(top 1) sim: $0.98$} & & \multicolumn{2}{c}{(top 5) sim: $0.97$} & & \multicolumn{2}{c}{(top 10) sim: $0.96$} & & \multicolumn{2}{c}{(top 50) sim: $0.95$}\\
\toppair{00001}{ultra-fine_insect_mesh} & & \toppair{00005}{Venetian_glass_chandelier__} & & \toppair{00010}{rabbit} & & \toppair{00050}{Roman_silver_coin__}\\
\multicolumn{2}{c}{\makebox[102pt]{ultra-fine insect mesh}} & & \multicolumn{2}{c}{Venetian glass chandelier} & & \multicolumn{2}{c}{rabbit} & & \multicolumn{2}{c}{Roman silver coin} \\[4pt]

\multicolumn{2}{c}{(top 100) sim: $0.95$} & & \multicolumn{2}{c}{(top 200) sim: $0.94$} & & \multicolumn{2}{c}{(top 300) sim: $0.94$} & & \multicolumn{2}{c}{(top 400) sim: $0.94$} \\
\toppair{00100}{Egyptian_mummy_portrait__} & & \toppair{00200}{Amphitheater} & & \toppair{00300}{Egyptian_funerary_mask__} & & \toppair{00400}{Ancient_Greek_coin__}\\
\multicolumn{2}{c}{Egyptian mummy portrait} & & \multicolumn{2}{c}{Amphitheater} & & \multicolumn{2}{c}{\makebox[102pt]{Egyptian funerary mask}} & & \multicolumn{2}{c}{Ancient Greek coin} \\[4pt]

\multicolumn{2}{c}{(top 500) sim: $0.94$} & & \multicolumn{2}{c}{(top 1000) sim: $0.93$} & & \multicolumn{2}{c}{(top 2000) sim: $0.92$} & & \multicolumn{2}{c}{(top 3000) sim: $0.92$} \\
\toppair{00500}{High-rise_apartment_building} & & \toppair{01005}{Botanical_garden_pavilion} & & \toppair{02000}{Square} & & \toppair{03004}{Victorian_townhouse}\\
\multicolumn{2}{c}{\makebox[102pt]{High-rise apartment building}} & & \multicolumn{2}{c}{\makebox[102pt]{Botanical garden pavilion}} & & \multicolumn{2}{c}{Square} & & \multicolumn{2}{c}{\makebox[102pt]{Victorian townhouse}} \\[4pt]

\multicolumn{2}{c}{(top 4000) sim: $0.91$} & & \multicolumn{2}{c}{(top 5000) sim: $0.91$} & & \multicolumn{2}{c}{(top 10000) sim: $0.90$} & & \multicolumn{2}{c}{(top 20000) sim: $0.88$} \\
\toppair{04006}{Persian_blue_tile__} & & \toppair{05003}{Residential_house_Modern_European_style} & & \toppair{10002}{Venetian_glass_chandelier__} & & \toppair{20001}{Clock_tower}\\
\multicolumn{2}{c}{\makebox[102pt]{A Persian blue tile}} & & \multicolumn{2}{c}{\makebox[102pt]{Residential house}} & & \multicolumn{2}{c}{\makebox[102pt]{Venetian glass chandelier}} & & \multicolumn{2}{c}{\makebox[102pt]{Clock tower}}

\end{tabular}
}

\caption{Top similar pairs ranked from 1st to 20000th computed by the trained model (ID7). Pairs within roughly the top 500 are visually very similar, whereas pairs beyond this rank tend to be visually distinguishable. We therefore remove one image from each of the top 500 most similar pairs for training.}
 \label{fig:suppl_seed_similar_pairs}
 \vspace{-8pt}
\end{figure}

\begin{table*}[t]
\centering
\caption{Study on instance identity and diversity. ID-S12 removes one image from each of the top 500 most similar pairs. ID-S13 increases instance diversity by appending a feature variation to the prompt for each instance.}
\centering
\scalebox{1}{\begin{tabular}{@{}clc@{\hspace{8pt}}cc cc c@{\hspace{1pt}} cc c@{\hspace{1pt}} c c@{\hspace{1pt}} cc@{}}
\toprule

\multirow{2}{*}{ID} & \multirow{2}{*}{dataset} & \multirow{2}{*}{dedup.} & \multirow{2}{*}{instances} & \multirow{2}{*}{diversity} & \multirow{2}{*}{avg} & \multicolumn{1}{c}{artwork} & & \multicolumn{2}{c}{landmark} & & \multicolumn{1}{c}{product} & & \multicolumn{2}{c}{multi} \\
\cline{7-7}
\cline{9-10}
\cline{12-12}
\cline{14-15}

 & & & & & & MET & & ROP & GLD & & SOP & & INS & mIL \\
\midrule 

1 & pretrained & - & - & - & $47.5$ & $67.3$ & & $45.0$ & $15.7$ & & $55.4$ & & $80.6$ & $21.0$ \\
\hline 
7 & generated & \ding{55} & $20,000$ & - & $52.7$ & $75.1$ & & $48.6$ & $18.7$ & & $55.6$ & & $87.5$ & $30.6$ \\
S12 & generated & \cellcolor{oursrow}\ding{52} & \cellcolor{oursrow}$19,690$ & - & $52.9$ & $75.2$ & & $49.3$ & $18.6$ & & $55.8$ & & $87.7$ & $30.6$ \\

S13 & generated & \ding{55} & $20,000$& \cellcolor{oursrow}\ding{52} & $52.4$ & $75.6$ & & $48.6$ & $18.9$ & & $55.8$ & & $86.5$ & $28.8$ \\

\bottomrule
\end{tabular}}
\label{tab:sup:seeds}
\end{table*}

\begin{figure*}[t]
\footnotesize

\newcommand{\ori}[2]{%
  \includegraphics[width=55pt,height=55pt]{fig/diversity/prev_#1/#2.jpg}
}

\newcommand{\orirow}[2]{%
  \ori{#1}{\number\numexpr#2\relax} &
  \ori{#1}{\number\numexpr#2+1\relax} &
  \ori{#1}{\number\numexpr#2+2\relax} &
  \ori{#1}{\number\numexpr#2+3\relax} &
  \ori{#1}{\number\numexpr#2+4\relax} &
  \ori{#1}{\number\numexpr#2+5\relax} &
  \ori{#1}{\number\numexpr#2+6\relax} &
  \ori{#1}{\number\numexpr#2+7\relax} &
  \ori{#1}{\number\numexpr#2+8\relax} &
  \ori{#1}{\number\numexpr#2+9\relax} \\
}

\newcommand{\diversity}[2]{%
  \includegraphics[width=55pt,height=55pt]{fig/diversity/#1/#2.jpg}
}

\newcommand{\divrow}[2]{%
  \diversity{#1}{\number\numexpr#2\relax} &
  \diversity{#1}{\number\numexpr#2+1\relax} &
  \diversity{#1}{\number\numexpr#2+2\relax} &
  \diversity{#1}{\number\numexpr#2+3\relax} &
  \diversity{#1}{\number\numexpr#2+4\relax} &
  \diversity{#1}{\number\numexpr#2+5\relax} &
  \diversity{#1}{\number\numexpr#2+6\relax} &
  \diversity{#1}{\number\numexpr#2+7\relax} &
  \diversity{#1}{\number\numexpr#2+8\relax} &
  \diversity{#1}{\number\numexpr#2+9\relax} \\
}

\scalebox{0.78}{\begin{tabular}
{@{\hspace{0pt}}p{9pt}p{45pt}p{45pt}p{45pt}p{45pt}p{45pt}p{45pt}p{45pt}p{45pt}p{45pt}p{45pt}@{}}

\raisebox{25pt}{table} & \orirow{0042}{0} \\[-5pt]
& size: large (55 inches+) & top surface material & extension leaf, umbrella hole & weather resistance finishing & umbrella protection & round or square shape & top material & with storage & rustic or modern styles & rectangular or round configuration\\
\raisebox{25pt}{table} & \divrow{0042}{0} \\[-5pt]
\hline
& \\[-3pt]
\raisebox{25pt}{vase} & \orirow{0483}{0} \\[-5pt]
& design details & bottle or bowl shape & miniature & drainage holes & painted designs & gradient pattern & flared base & tapered neck & minimalist & flat bottom \\
\raisebox{25pt}{vase} & \divrow{0483}{0} \\

\end{tabular}}

\vspace{-15pt}
\caption{Examples of generated images using only object name as prompt (rows 1 and 3) \vs object name followed by a specific feature variation, separated by a comma (rows 2 and 4). Prompts enriched with feature variations tend to produce visually more diverse instances.}
\label{fig:sup:diversity}
\end{figure*}

\paragraph{Increase instance diversity}
To further increase instance diversity, we enrich the generation prompts by first producing 10 features per object and then five variations of each feature using ChatGPT. 
The prompt is ``\texttt{You are a helpful assistant. You will be given a list of objects, and for each object, you will generate 10 structural and design features that set objects apart from others in the same category (i.e., the objects are unique). Each feature/type should have 5 variances. The output format should be as follows: <object>$\backslash$t<feature\_or\_type>$\backslash$t<variance 1>$\backslash$t<variance 2>$\backslash$t<variance 3>$\backslash$t<variance 4>$\backslash$t<variance 5>. Only list features and variances that can be visually observed in images. Color and size are not necessary}''. For each feature, one variation is randomly selected to form a prompt of the format ``\textit{<object>, <variation>}''. 
In the generic domain, 10 instances are generated from the 10 feature-based prompts per object. For the artwork and product domain, each feature yields two prompts, from which 15 are randomly sampled. For the landmark domain, each feature provides eight prompts. \cref{fig:sup:diversity} presents examples generated with only the object name and with appended feature variations. Although the resulting instances appear visually more diverse, ID-S13 in \cref{tab:sup:seeds} does not show any performance improvement, but even a minor, insignificant, drop of 0.3 compared to \oursplus{} (ID7), suggesting that our default setting and even lower object diversity is effective enough for instance-level representation learning.

\paragraph{Failure in preserving instance identity}
We present some failure cases in \cref{fig:sup:failure_case} in which instance identity is not preserved well during generation. This is either due to the foreground object appearance getting modified for different backgrounds. Nevertheless, note that these cases are very rare cases that were hard to find. In row 1, the trees painted in ``Japanese screen'' are rendered as real trees after applying background variations, thereby altering the identity of the instance. As shown in rows 2 to 3, while the overall structure of the landmarks is preserved, their appearances vary with the surrounding environment. Unlike other domains, in real-world landmark datasets, such surrounding differences typically correspond to distinct landmarks, making these variations difficult to train. This discrepancy limits the performance of our method in the landmark domain, suggesting that a more moderate augmentation strategy may be better for further exploration.

\begin{figure*}[t]
\centering
\hspace{-20pt}
\footnotesize

\newcommand{\failimg}[2]{%
  \raisebox{-.5\height}{\includegraphics[width=63pt,height=63pt]{fig/failure_case/#1/#2.jpg}}%
}

\newcommand{\failure}[3]{%
  \failimg{#1}{#2}&%
  \failimg{#1}{#3}&%
  \failimg{#1}{\number\numexpr#3+1\relax}&%
  \failimg{#1}{\number\numexpr#3+2\relax}&%
  \failimg{#1}{\number\numexpr#3+3\relax}%
}

\newcommand{\rowlabel}[1]{%
  \parbox[c]{60pt}{\raggedleft #1}%
}

\begin{tabular}{@{} c c@{\hspace{8pt}}c@{\hspace{3pt}}c@{\hspace{3pt}}c@{\hspace{3pt}}c @{}}

\rowlabel{18th-century Japanese screen} & \failure{10676}{01}{42704} \\[35pt]
\rowlabel{Train station}                & \failure{16195}{35}{64780} \\[35pt]
\rowlabel{Public restroom building}     & \failure{19559}{39}{78236} \\[35pt]

\end{tabular}
\caption{Examples of failure cases in preserving the object identity. Column 1 presents the instances generated by GDM, and columns 2 to 5 show the corresponding images with background variations.}
\label{fig:sup:failure_case}
\end{figure*}

\section{Data generation}
\label{sec:sup:data_generation}

\paragraph{Prompts on object categories generation}

\cref{tab:sup:prompts} provides the designed and template prompts used to generate object categories for each domain.
Designed prompts are used in all cases except for ID-3 in \cref{tab:ablation}, where we use template prompts for the ablation study. 
The template is ``\textit{Provide a raw list of} $2000$ \textit{object names from the domain of} \texttt{<domain>} \textit{objects. Here are some examples, which you can include in your list too but also get inspired by} \texttt{<examples>}.'' 
For each domain, we substitute the \texttt{<domain>} and \texttt{<examples>} accordingly. We randomly sample images to match the number of instances per domain in \cref{tab:gene_details}.

\begin{table*}[h]
    \centering
    \caption{Designed and template prompts for object categories generation.}
    \begin{tabular}{p{0.1\textwidth} p{0.43\textwidth} p{0.43\textwidth}}
    \toprule
    \textbf{domains} & \textbf{Designed prompts} & \textbf{Template prompts} \\
    \midrule
    generic       &   Provide a raw list of 2000 objects names from the domain of everyday objects, such as household items, retail products, electronics, collectibles, vehicles, buildings, outdoor objects, etc.
    Here are some examples, which you can include in your list too but also get inspired by sandal, mug, laptop, chair, bottle, temple, house, dress, teapot, dog, toy, rabbit, teddybear, car, toy, car, bowl, church, skyscraper hotel. & Provide a raw list of 2000 objects names from the domain of \textit{everyday} objects. Here are some examples, which you can include in your list too but also get inspired by \textit{sandal, mug, laptop, chair, house, dress, teddybear, car, toy, church.} \\
    \hline
    art           &   Provide a raw list of 200 object names from the domain of museum items. Consider a encyclopedic art museum that is home to collections classic art such as paintings, graphic work, jewelry, vases, sculptures, but also of musical instruments, costumes, and decorative arts and textiles, as well as antique weapons and armor from around the world. & Provide a raw list of 2000 objects names from the domain of \textit{museum} objects. Here are some examples, which you can include in your list too but also get inspired by \textit{Renaissance oil painting, Baroque tapestry, Egyptian faience amulet, Medieval longsword, Japanese samurai armor, Greek krater vase, Rococo gilded mirror, Ancient Roman cameo ring, Venetian glass chandelier, 19th-century concert grand piano.} \\
    \hline
    landmark     &   Provide a raw list of 50 object names from the domain of buildings, landmarks, urban structures, outdoor constructions, such as church, neoclassical building, train station, temple, cathedral, tower building, square. It can be fine-grained too, \eg, catholic church. Please name the most common things, such as house in standard modern European architecture. & Provide a raw list of 2000 objects names from the domain of \textit{landmarks} objects. Here are some examples, which you can include in your list too but also get inspired by \textit{catholic church, neoclassical building, train station, temple, cathedral, tower building, square, Mosque, Skyscraper, Castle.} \\
    \hline
    product      &   Provide a raw list of 200 object names from the domain of retail products, supermarket products, e-shop electronics, clothes, fashion items, shoes, anything that someone would sell in a second hand online market. & Provide a raw list of 2000 objects names from the domain of \textit{retail products} objects. Here are some examples, which you can include in your list too but also get inspired by \textit{Leather jacket, Smartphone, Gaming console, Bluetooth headphones, Smartwatch, Designer handbag, Running shoes, Vintage dress, DSLR camera.} \\
    \bottomrule
\end{tabular}
    \label{tab:sup:prompts}
\end{table*}

\begin{figure*}[!h]
\centering
\hspace{-2cm}
\newcommand{\objstep}[2]{\includegraphics[width=48pt,height=48pt]{fig/suppl_hq/noise_0_#1_inf_step_#2.jpg}}

\begin{tabular}
{@{\hspace{0pt}}c@{\hspace{2pt}}c@{\hspace{2pt}}c@{\hspace{2pt}}c@{\hspace{2pt}}c@{\hspace{2pt}}c@{\hspace{2pt}}c@{\hspace{2pt}}c@{\hspace{2pt}}c@{\hspace{2pt}}c@{\hspace{2pt}}c@{\hspace{2pt}}}

& step $1$ & step $2$ & step $3$ & step $4$ & step $5$ & step $6$ & step $7$ & step $8$ & step $9$ \\
\raisebox{20pt}{chair} & \objstep{chair}{1} & \objstep{chair}{2} & \objstep{chair}{3} & \objstep{chair}{4} & \objstep{chair}{5} & \objstep{chair}{6} & \objstep{chair}{7} & \objstep{chair}{8} & \objstep{chair}{9} \\

\raisebox{20pt}{table} & \objstep{table}{1} & \objstep{table}{2} & \objstep{table}{3} & \objstep{table}{4} & \objstep{table}{5} & \objstep{table}{6} & \objstep{table}{7} & \objstep{table}{8} & \objstep{table}{9} \\

\end{tabular}

\caption{Images generated by GDM with inference steps ranging from $1$ to $9$. To minimize artifacts in the object while preventing a detailed background, we select $5$ steps for the ablation study.
 \label{fig:sup:hq}}
\end{figure*}

\paragraph{Examples with different inference steps}
Examples generated by Stable Diffusion Turbo with different inference steps are shown in \cref{fig:sup:hq}. Increasing the number of steps enhances the quality of the objects, but also results in more detailed backgrounds. To balance the higher quality of the object and a less detailed background, we choose images with $5$ inference steps for the ablation study.

\section{Backbone details}
\label{sec:sup:backbone_details} 
\cref{tab:sup:backbones} includes further details of the three backbones.

\begin{table}[t]
    \centering
    \caption{Details of backbones.}
    \begin{tabular}{lrr}
    \toprule
    \textbf{backbone} & \textbf{params} & \textbf{dim} \\
    \midrule
    SigLIP  &   $316$M &  $1,024$ \\
    CLIP    &   $304$M &  $1,024$ \\
    ViT-B     &   $86$M &  $768$ \\
    \bottomrule
\end{tabular}
    \label{tab:sup:backbones}
\end{table}

\section{Additional study}
\label{sec:sup:additional_study}

\paragraph{Image generation hyperparameters}
To validate the robustness of the data generation hyperparameters, we evaluate several combinations of the number of objects $C$ and the number of instances for each object $K$. Results are reported in \cref{tab:sup:hyperparameter}. For the generic domain (ID-S14 to ID-S18), we keep the same total number of training images ($C \times K$). The best configuration (ID-S15) surpasses the default setting (ID5) by only 0.4 on average, indicating low sensitivity to these parameters. As $C$ decreases and $K$ increases, MET and INSTRE remain stable, while the landmark datasets (ROP and GLD) drop. Interestingly, ILIAS improves and reaches its best performance when $C=10$ and $K=2,000$ (ID-S18), presumably benefiting from a large number of instances per object, which encourages the model to capture fine-grained intra-class differences. Mixing both extremes in ID-S19 (combining ID-S18 and ID5) yields only average performance. Replacing the generic domain in \oursplus{} with ID-S18 does not improve overall performance (ID-S24). For the artwork and product domain (ID-S20, ID-S22), reducing $C$ while increasing $K$ drops on both domain-specific and average performance. For the landmark domain (ID-S21), adopting the same $C$ and $K$ as artwork and product does not improve standalone performance. However, when this setting substitutes the landmark domain in \oursplus{}, it achieves a minor gain of 0.1 (ID-S23 \textit{v.s.} ID7). Overall, these results demonstrate that our methodology is robust to a wide range of hyperparameter choices.
\begin{table*}[t]

\centering
    \caption{Ablation on data generation hyperparameters (number of objects $C$ and number of instances per object $K$). ID-S14 to ID-S22 train on a single domain, while ID-S23 and ID-S24 train on all domains with one domain replaced relative to \oursplus{}. Changes with respect to the default setting — \oursgeneric{} (ID5), \oursspecific{} (ID9, ID11, ID13), and \oursplus{} (ID7) — are \colorbox{oursrow}{highlighted}.}
\centering
\scalebox{1}{
\begin{tabular}{@{}c r@{\hspace{6pt}}r@{\hspace{4pt}} cr@{\hspace{6pt}}r@{\hspace{4pt}} cr@{\hspace{6pt}}r@{\hspace{4pt}}  cr@{\hspace{6pt}}r@{\hspace{4pt}} c@{\hspace{8pt}}c@{\hspace{8pt}}c@{\hspace{8pt}}c@{\hspace{8pt}}c@{\hspace{8pt}}c@{\hspace{8pt}}c@{\hspace{8pt}}c@{\hspace{8pt}}c@{\hspace{8pt}}c@{}}
\toprule

\multirow{2}{*}{\textbf{ID}} & \multicolumn{2}{c}{\textbf{generic}} & & \multicolumn{2}{c}{\textbf{artwork}} & & \multicolumn{2}{c}{\textbf{landmark}} & & \multicolumn{2}{c}{\textbf{product}} & & \multicolumn{7}{c}{\textbf{results}} \\
 \cline{2-3}
 \cline{5-6}
 \cline{8-9}
 \cline{11-12}
 \cline{14-20}

& $C$ & $K$ & & $C$ & $K$  & & $C$ & $K$  & & $C$ & $K$  & & avg & MET & ROP & GLD & SOP & INS & mIL \\
\midrule 

$1$ & - & - & & - & - & & - & - & & - & - & & $47.5$ & $67.3$ & $45.0$ & $15.7$ & $55.4$ & $80.6$ & $21.0$ \\
\midrule

S14 & \cellcolor{oursrow}$4,000$ & \cellcolor{oursrow}$5$ & & - & - & & - & - & & - & - & & $50.8$ & $72.8$ & $46.1$ & $17.7$ & $56.3$ & $85.4$ & $26.2$ \\
5 & $2,000$ & $10$ & & - & - & & - & - & & - & - & & $50.8$ & $72.4$ & $46.4$ & $17.4$ & $55.9$ & $85.4$ & $27.4$ \\
S15 & \cellcolor{oursrow}$1,000$ & \cellcolor{oursrow}$20$ & & - & - & & - & - & & - & - & & $51.2$ & $72.9$ & $45.7$ & $17.5$ & $55.8$ & $86.5$ & $28.8$ \\
S16 & \cellcolor{oursrow}$500$ & \cellcolor{oursrow}$40$ & & - & - & & - & - & & - & - & & $50.5$ & $72.5$ & $45.0$ & $17.1$ & $54.7$ & $85.4$ & $28.4$ \\
S17 & \cellcolor{oursrow}$100$ & \cellcolor{oursrow}$200$ & & - & - & & - & - & & - & - & & $49.9$ & $73.2$ & $41.5$ & $15.3$ & $52.6$ & $85.4$ & $31.4$ \\
S18 & \cellcolor{oursrow}$10$ & \cellcolor{oursrow}$2,000$ & & - & - & & - & - & & - & - & & $50.0$ & $72.7$ & $40.0$ & $14.4$ & $51.2$ & $86.4$ & $35.3$ \\

\multirow{2}{*}{S19} & \cellcolor{oursrow}$2,000$ & \cellcolor{oursrow}$10$ & & \multirow{2}{*}{-} & \multirow{2}{*}{-} & & \multirow{2}{*}{-} & \multirow{2}{*}{-} & & \multirow{2}{*}{-} & \multirow{2}{*}{-} & & \multirow{2}{*}{$50.3$} & \multirow{2}{*}{$72.9$} & \multirow{2}{*}{$43.1$} & \multirow{2}{*}{$15.8$} & \multirow{2}{*}{$54.3$} & \multirow{2}{*}{$85.7$} & \multirow{2}{*}{$30.3$} \\
 & \cellcolor{oursrow}$10$ & \cellcolor{oursrow}$2,000$ & &  &  & &  &  & &  &  & & & & & & & & \\
\midrule

$9$ & - & - & & $200$ & $15$ & & - & - & & - & - & & $51.2$ & $73.7$ & $47.0$ & $17.2$ & $55.6$ & $85.4$ & $28.3$ \\
S20 & - & - & & \cellcolor{oursrow}$10$ & \cellcolor{oursrow}$300$ & & - & - & & - & - & & $50.7$ & $72.6$ & $46.7$ & $17.2$ & $54.4$ & $85.3$ & $27.9$ \\
\midrule

$11$ & - & - & & - & - & & $50$ & $80$ & & - & - & & $51.0$ & $72.5$ & $50.7$ & $19.7$ & $54.6$ & $84.4$ & $24.2$ \\
S21 & - & - & & - & - & & \cellcolor{oursrow}$200$ & \cellcolor{oursrow}$15$ & & - & - & & $50.3$ & $71.9$ & $49.9$ & $19.1$ & $55.2$ & $82.7$ & $22.9$ \\
\midrule

$13$ & - & - & & - & - & & - & - & & $200$ & $15$ & & $50.5$ & $71.6$ & $46.4$ & $17.0$ & $55.9$ & $85.0$ & $27.1$ \\
S22 & - & - & & - & - & & - & - & & \cellcolor{oursrow}$50$ & \cellcolor{oursrow}$60$ & & $50.4$ & $71.5$ & $45.6$ & $17.1$ & $55.6$ & $85.0$ & $27.6$ \\
\midrule

$7$ & $2,000$ & $10$ & & $200$ & $15$ & & $50$ & $80$ & & $200$ & $15$ & & $52.7$ & $75.1$ & $48.6$ & $18.7$ & $55.6$ & $87.5$ & $30.6$ \\
S23 & $2,000$ & $10$ & & $200$ & $15$ & & \cellcolor{oursrow}$200$ & \cellcolor{oursrow}$15$ & & $200$ & $15$ & & $52.8$ & $74.9$ & $49.0$ & $19.0$ & $56.2$ & $87.0$ & $30.9$ \\
S24 & \cellcolor{oursrow}$10$ & \cellcolor{oursrow}$2,000$ & & $200$ & $15$ & & $50$ & $80$ & & $200$ & $15$ & & $52.1$ & $74.4$ & $46.3$ & $17.3$ & $53.3$ & $87.5$ & $33.7$ \\

\bottomrule
\end{tabular}
}
\label{tab:sup:hyperparameter}
\vspace{-5pt}
\end{table*}

\paragraph{Rendered viewpoints}
We use one of the cutting-edge image-to-3D model TRELLIS~\citep{xiang2025structured} to synthesize viewpoints from different angles. Given instances generated by GDM as input, TRELLIS renders synthetic viewpoints at seven angles (90, 120, 150, 180, 210, 240, and 270 degrees). To maintain the same number of images per instance, we randomly sample four views and then apply background generation. Examples are shown in \cref{fig:sup:views_v3}. Results shown as ID-S31 in \cref{tab:sup:views} indicate that the rendered viewpoints degrade performance on most datasets, despite producing visually plausible generation. This may be attributed to 3D reconstruction artifacts, especially for landmark objects.

\begin{figure*}[t]
\centering
\hspace{-6cm}
\footnotesize
\newcommand{\ori}[2]{\includegraphics[width=40pt,height=40pt]{fig/multi_views/#1/#2.jpg}}

\newcommand{\views}[5]{\includegraphics[width=40pt,height=40pt]{fig/multi_views/#1/views/angle_#2.png}&\includegraphics[width=40pt,height=40pt]{fig/multi_views/#1/views/angle_#3.png}&\includegraphics[width=40pt,height=40pt]{fig/multi_views/#1/views/angle_#4.png}&\includegraphics[width=40pt,height=40pt]{fig/multi_views/#1/views/angle_#5.png}}

\newcommand{\bg}[5]{\includegraphics[width=40pt,height=40pt]{fig/multi_views/#1/bg/jpg/angle_#2.jpg}&\includegraphics[width=40pt,height=40pt]{fig/multi_views/#1/bg/jpg/angle_#3.jpg}&\includegraphics[width=40pt,height=40pt]{fig/multi_views/#1/bg/jpg/angle_#4.jpg}&\includegraphics[width=40pt,height=40pt]{fig/multi_views/#1/bg/jpg/angle_#5.jpg}}

\begin{tabular}{@{} c@{\hspace{2pt}} c@{\hspace{2pt}}c@{\hspace{2pt}}c@{\hspace{2pt}}c@{\hspace{2pt}} c@{\hspace{5pt}} c@{\hspace{2pt}} c@{\hspace{2pt}}c@{\hspace{2pt}}c@{\hspace{2pt}}c@{\hspace{2pt}} l@{\hspace{2pt}} @{}}

\multirow{2}{*}[-19pt]{backpack} & degree=90 & 150 & 180 & 210 & & \multirow{2}{*}[-19pt]{bust} & degree=90 & 180 & 210 & 270 \\
\multirow{2}{*}[9pt]{\ori{backpack}{3}} & \views{backpack}{90}{150}{180}{210} & & \multirow{2}{*}[9pt]{\ori{bust}{04}} & \views{bust}{90}{180}{210}{270} & \raisebox{20pt}{+ \textit{views}} \\
& & & & & & & & & & & \raisebox{-10pt}{+ \textit{views}} \\ [-18pt]
& \bg{backpack}{90}{150}{180}{210} & & & \bg{bust}{90}{180}{210}{270} & \raisebox{14pt}{+ \textit{background}} \\[8pt]

\multirow{2}{*}[-19pt]{car} & degree=210 & 240 & 90 & 270 & & \multirow{2}{*}[-19pt]{church} & degree=180 & 210 & 240 & 270 \\
\multirow{2}{*}[9pt]{\ori{car}{3}} & \views{car}{210}{240}{90}{270} & & \multirow{2}{*}[9pt]{\ori{church}{07}} & \views{church}{180}{210}{240}{270} & \raisebox{20pt}{+ \textit{views}} \\
& & & & & & & & & & & \raisebox{-10pt}{+ \textit{views}} \\ [-18pt]
 & \bg{car}{210}{240}{90}{270} & & & \bg{church}{180}{210}{240}{270} & \raisebox{14pt}{+ \textit{background}} \\

\end{tabular}
\caption{Examples of synthetic viewpoints rendered by an image-to-3D model, TRELLIS, followed by background generation. To maintain the same number of instances, four views are randomly sampled from seven angles (90, 120, 150, 180, 210, 240, and 270 degrees).}
 \label{fig:sup:views_v3}
\end{figure*}

\begin{table*}[t]
\centering
\caption{Study on rendered viewpoints. ID-S31 renders four viewpoints per instance before applying background generation.}
\centering
\begin{tabular}{@{}clc@{\hspace{8pt}}cc c@{\hspace{1pt}} cc c@{\hspace{1pt}} c c@{\hspace{1pt}} cc@{}}
\toprule

\multirow{2}{*}{ID} & \multirow{2}{*}{dataset} & \multirow{2}{*}{views} & \multirow{2}{*}{avg} & \multicolumn{1}{c}{artwork} & & \multicolumn{2}{c}{landmark} & & \multicolumn{1}{c}{product} & & \multicolumn{2}{c}{multi} \\
\cline{5-5}
\cline{7-8}
\cline{10-10}
\cline{12-13}

 & & & & MET & & ROP & GLD & & SOP & & INS & mIL \\
\midrule 

1 & pretrained & - & $47.5$ & $67.3$ & & $45.0$ & $15.7$ & & $55.4$ & & $80.6$ & $21.0$ \\
\hline 
7 & generated & $1$ & $52.7$ & $75.1$ & & $48.6$ & $18.7$ & & $55.6$ & & $87.5$ & $30.6$ \\
S31 & generated & \cellcolor{oursrow}$4$ & $51.8$ & $74.3$ & & $47.8$ & $18.1$ & & $56.0$ & & $86.7$ & $28.0$ \\

\bottomrule
\end{tabular}
\label{tab:sup:views}
\end{table*}

\paragraph{Additional study on Objaverse}

When training on Objavese, we match the same dataset sizes ($20$K) as \oursplus{} to ensure a fair comparison (ID2). 
As shown in \cref{tab:sup:objaverse}, increasing the number of Objaverse images to $40$K performs a bit worse (ID-S10). Therefore, we did not increase further.
We additionally tested the impact of real viewpoint variations from Objaverse by switching from 4 views of our main experiment to one view, where only the background differs (ID2 vs ID-S11). There is a small impact.

\begin{table*}[!h]
\centering
\caption{Additional study on training on Objaverse. Views refer to viewpoints before background generation.}
\centering
\begin{tabular}{@{}clcc cc c@{\hspace{1pt}} cc c@{\hspace{1pt}} c c@{\hspace{1pt}} cc@{}}
\toprule

\multirow{2}{*}{ID} & \multirow{2}{*}{dataset} & \multirow{2}{*}{instances} & \multirow{2}{*}{views} & \multirow{2}{*}{avg} & \multicolumn{1}{c}{artwork} & & \multicolumn{2}{c}{landmark} & & \multicolumn{1}{c}{product} & & \multicolumn{2}{c}{multi} \\
\cline{6-6}
\cline{8-9}
\cline{11-11}
\cline{13-14}
     
 & & & & & MET & & ROP & GLD & & SOP & & INS & mIL \\
\midrule 

1 & pretrained & - & - & $47.5$ & $67.3$ & & $\textbf{45.0}$ & $15.7$ & & $55.4$ & & $80.6$ & $21.0$ \\
\hline 
2 & Objaverse-background & $20$K & $4$ & $\textbf{50.9}$ & $\textbf{74.1}$ & & $42.6$ & $15.87$ & & $57.6$ & & $\textbf{86.8}$ & $\textbf{28.6}$ \\
S10 & Objaverse-background & \cellcolor{oursrow}$40$K & $4$ & $50.1$ & $72.6$ & & $40.6$ & $14.6$ & & $\textbf{58.3}$ & & $86.4$ & $27.9$ \\
S11 & Objaverse-background & $20$K & \cellcolor{oursrow}$1$ & $50.3$ & $73.4$ & & $42.5$ & $\textbf{15.92}$ & & $57.4$ & & $85.8$ & $27.1$ \\

\bottomrule
\end{tabular}
\label{tab:sup:objaverse}
\end{table*}

\begin{table*}[h]
    \centering
    \caption{Training on real-labeled images using SigLIP with softmax margin loss.}
\begin{tabular}{@{}lcc c@{\hspace{1pt}} cc c@{\hspace{1pt}} c c@{\hspace{1pt}} cc@{}}
    \toprule
     \multirow{2}{*}{dataset} & \multirow{2}{*}{avg} & \multicolumn{1}{c}{artwork} & & \multicolumn{2}{c}{landmark} & & \multicolumn{1}{c}{product} & & \multicolumn{2}{c}{multi} \\
     \cline{3-3}
     \cline{5-6}
     \cline{8-8}
     \cline{10-11}

     & & MET & & ROP & GLD & & SOP & & INS & mIL \\
    \midrule 
    pretrained & $47.5$ & $67.3$ & & $45.0$ & $15.7$ & & $55.4$ & & $80.6$ & $21.0$ \\
    
    \midrule
    Real-S (artwork)   & $51.5$ & $\textbf{77.4}$ & & $47.8$ & $17.6$ & & $58.7$ & & $82.2$ & $\textbf{25.1}$ \\

    Real-S (landmark)  & $51.8$ & $73.3$ & & $56.8$ & $21.4$ & & $57.8$ & & $80.3$ & $21.2$ \\

    Real-S (product)  & $52.0$ & $72.1$ & & $47.1$ & $17.5$ & & $68.7$ & & $83.1$ & $23.2$ \\
    
    Real-ALL  & $\textbf{56.1}$ & $76.8$ & & $\textbf{58.3}$ & $\textbf{21.9}$ & & $\textbf{71.0}$ & & $\textbf{83.5}$ & $25.0$ \\
    \bottomrule
\end{tabular}

\label{tab:sup:real_smm}
\end{table*}

\paragraph{Training real-labeled images with softmax margin loss}
\cref{tab:sup:real_smm} presents the results of training real-labeled images with softmax margin loss on four domains using SigLIP. We observe a consistent improvement over the pre-trained model on average. Real-ALL achieves the best overall performance, particularly on ROP and SOP. Similar to the trend observed with recall@k loss, training on the artwork domain yields the most improvement on MET and mini-ILIAS.

\begin{figure*}[t]
\hspace{-30pt}
\begin{tabular} 
{@{\hspace{3pt}}r@{\hspace{3pt}}c@{\hspace{3pt}}c@{\hspace{3pt}}c@{\hspace{3pt}}c@{\hspace{10pt}}c@{\hspace{3pt}}c@{\hspace{3pt}}c@{\hspace{3pt}}c}
     &  query  & positive 1 &  positive 2 &  positive 3 &  query  & positive 1 &  positive 2 &  positive 3 \\

    \raisebox{23pt}{MET} &
    \includegraphics[width=52pt,height=52pt]{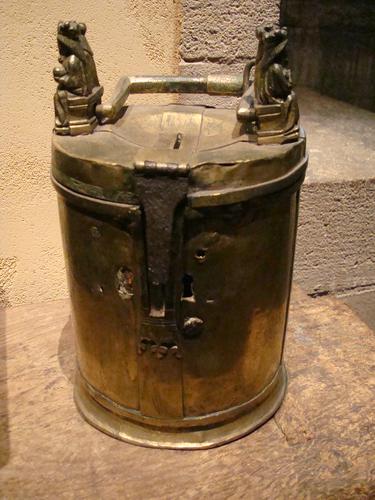} & 
    \includegraphics[width=52pt,height=52pt]{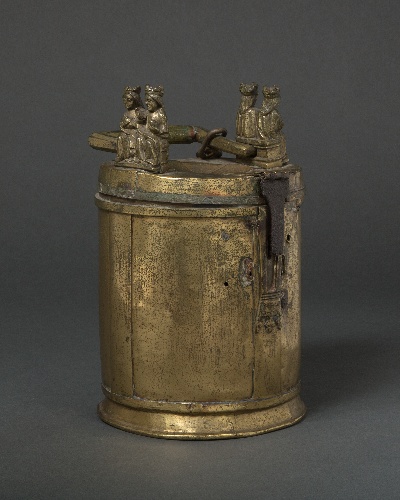} & 
    \includegraphics[width=52pt,height=52pt]{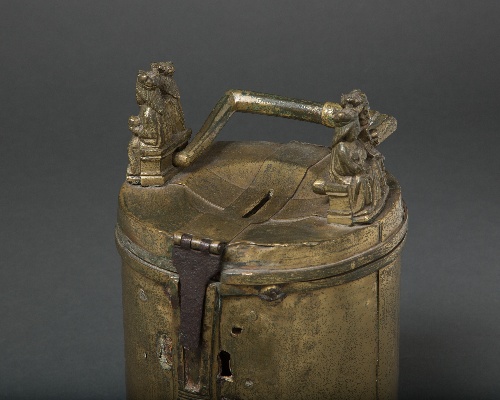} & 
    \includegraphics[width=52pt,height=52pt]{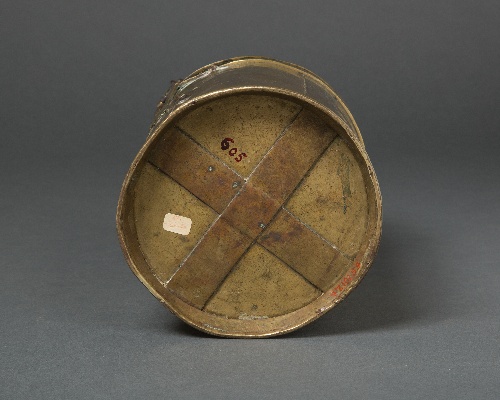}  &
    \includegraphics[width=52pt,height=52pt]{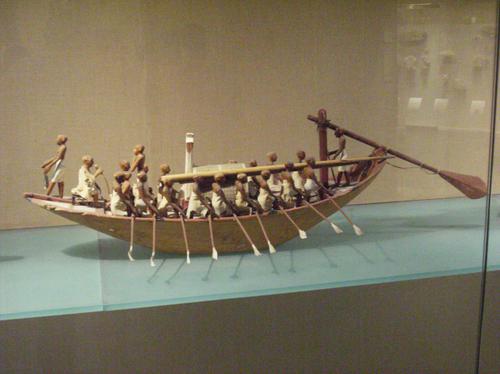} & 
    \includegraphics[width=52pt,height=52pt]{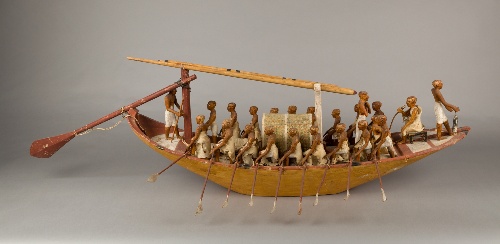} & 
    \includegraphics[width=52pt,height=52pt]{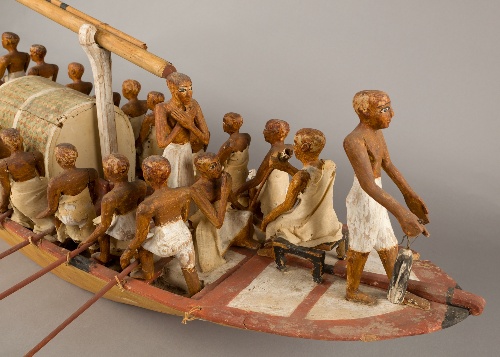} & 
    \includegraphics[width=52pt,height=52pt]{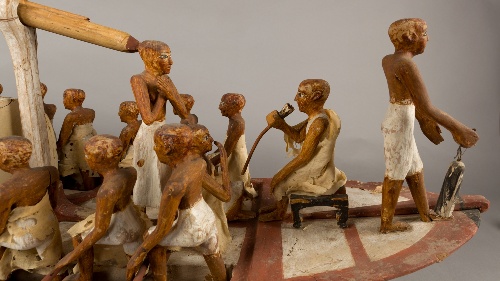} \\

    \raisebox{23pt}{R-Oxford} &
    \includegraphics[width=52pt,height=52pt]{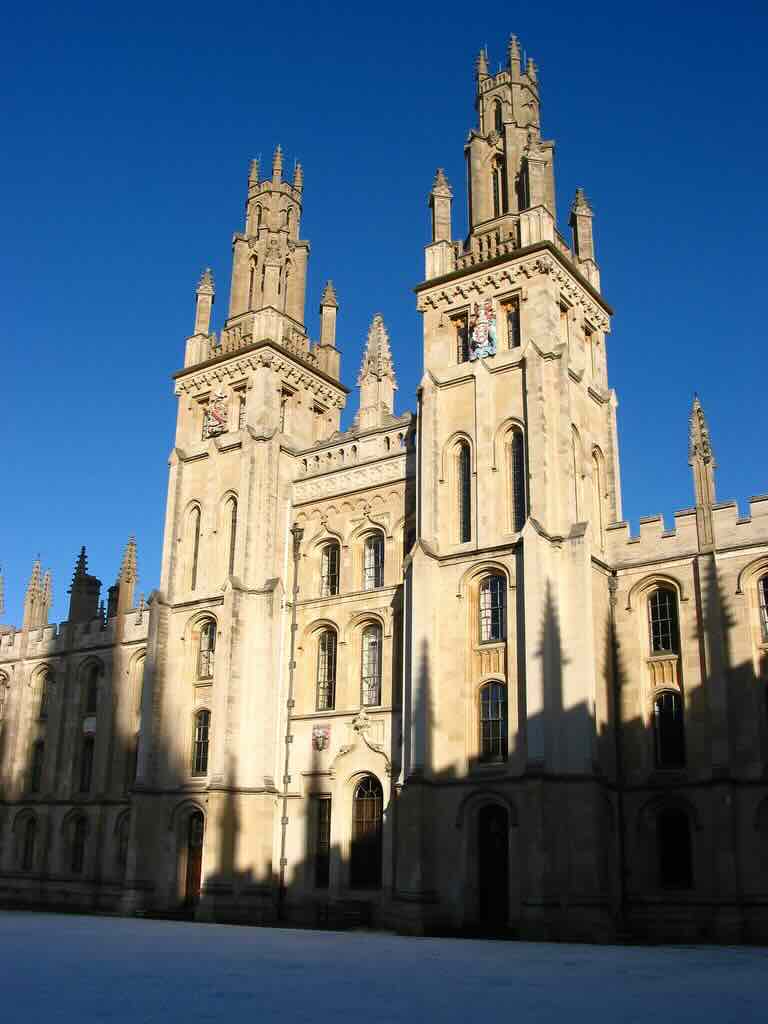} & 
    \includegraphics[width=52pt,height=52pt]{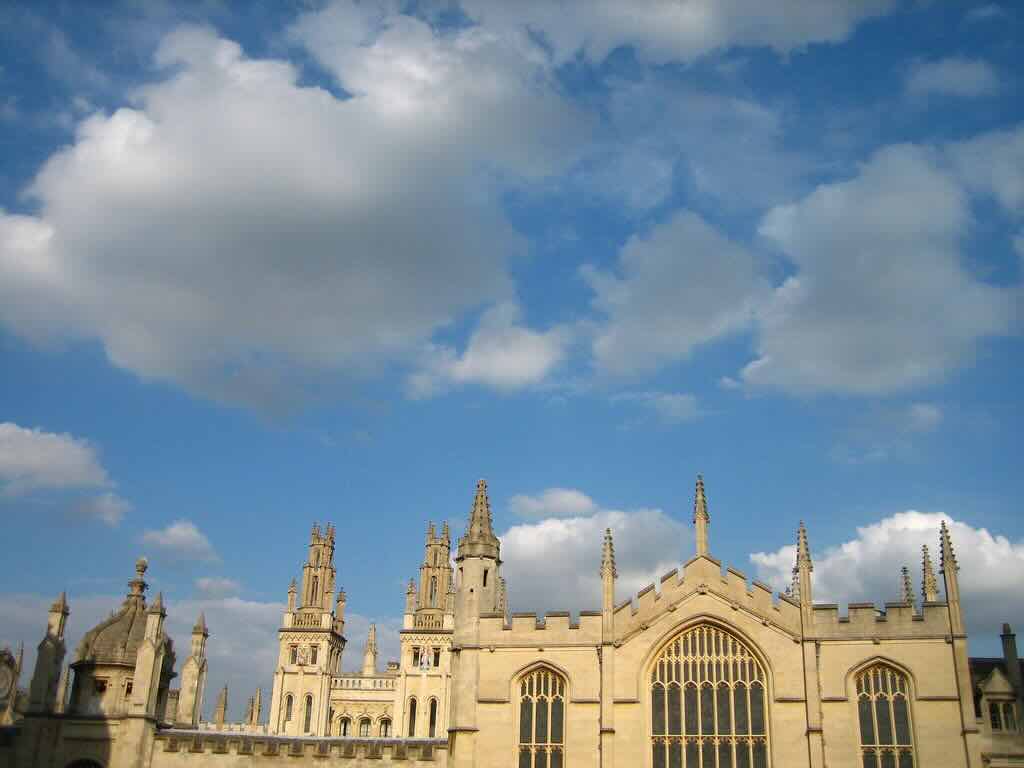} & 
    \includegraphics[width=52pt,height=52pt]{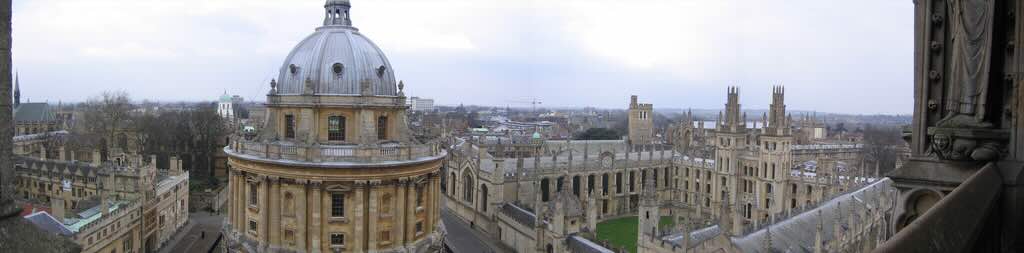} & 
    \includegraphics[width=52pt,height=52pt]{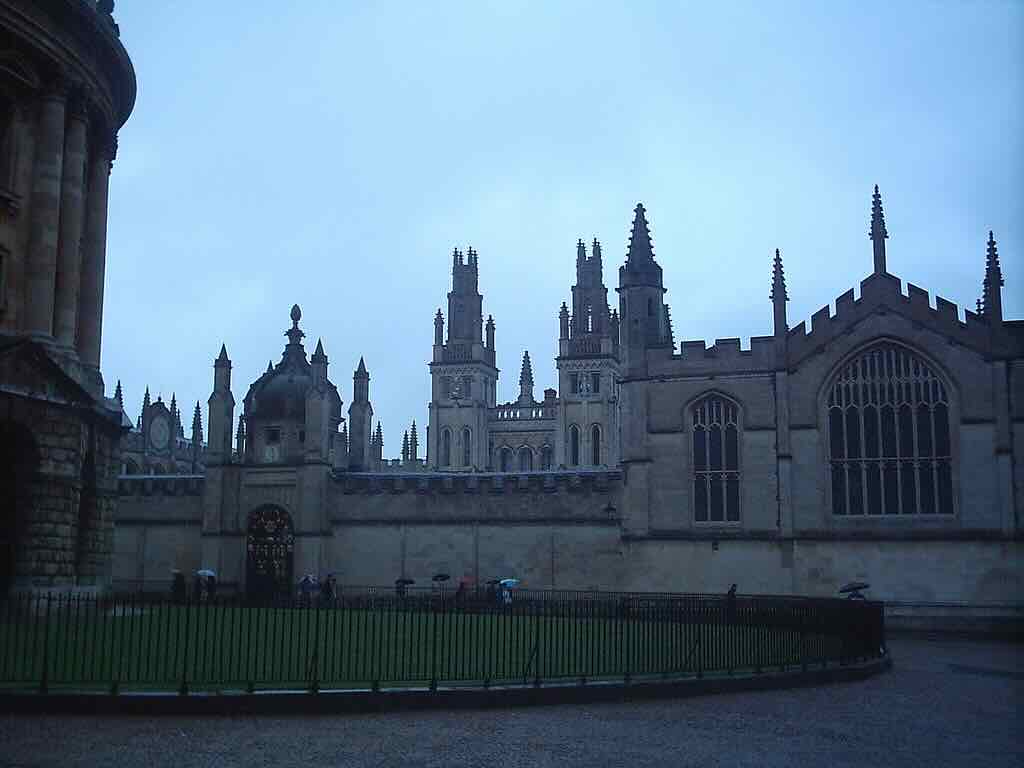} &
    \includegraphics[width=52pt,height=52pt]{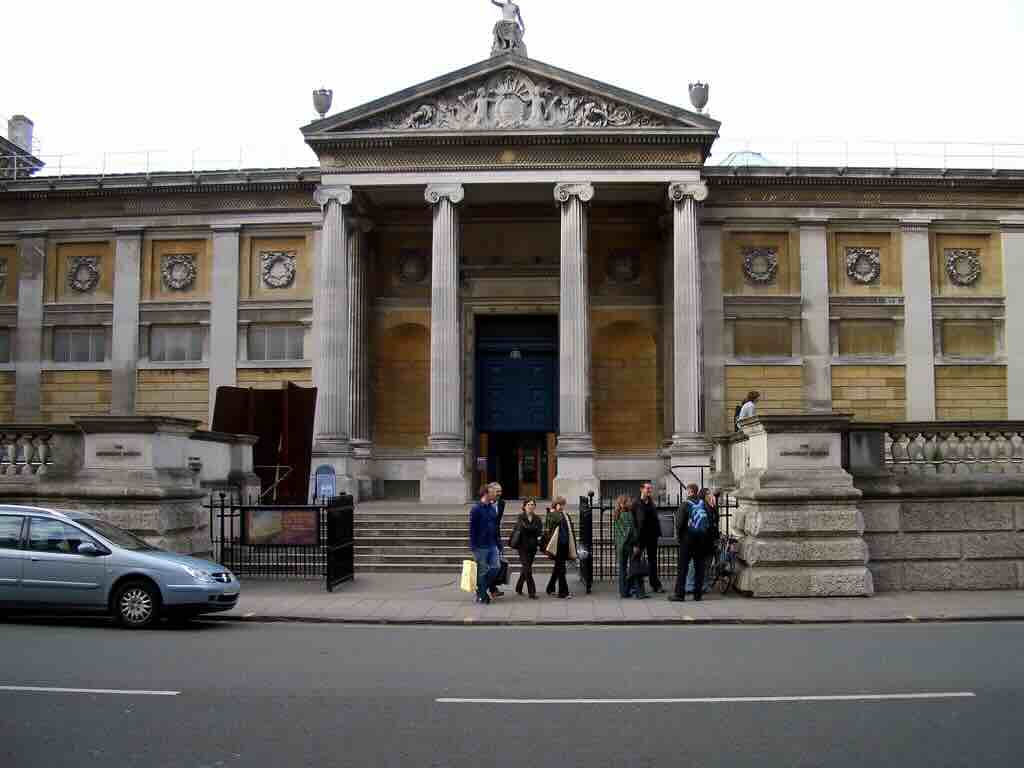} & 
    \includegraphics[width=52pt,height=52pt]{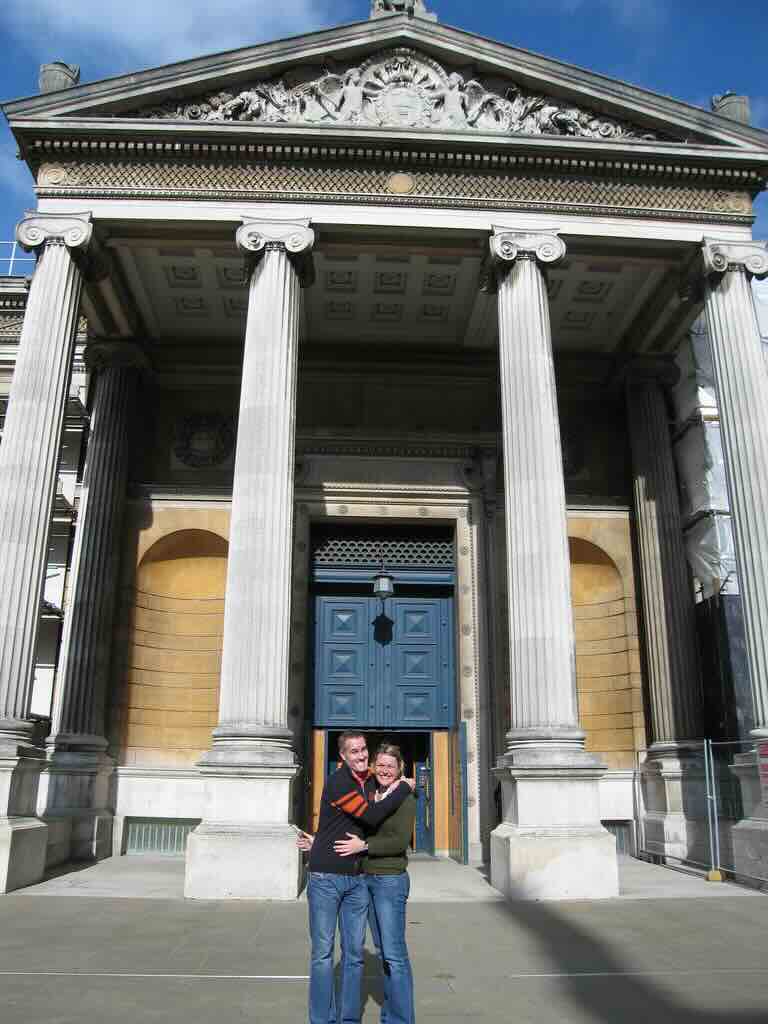} & 
    \includegraphics[width=52pt,height=52pt]{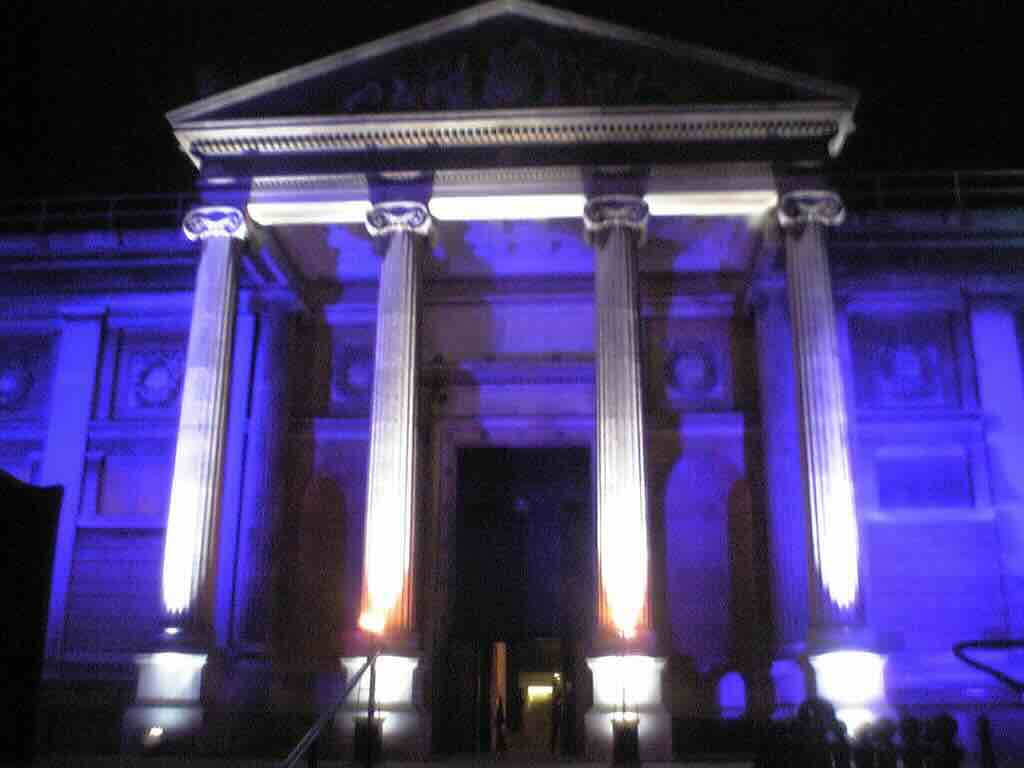} & 
    \includegraphics[width=52pt,height=52pt]{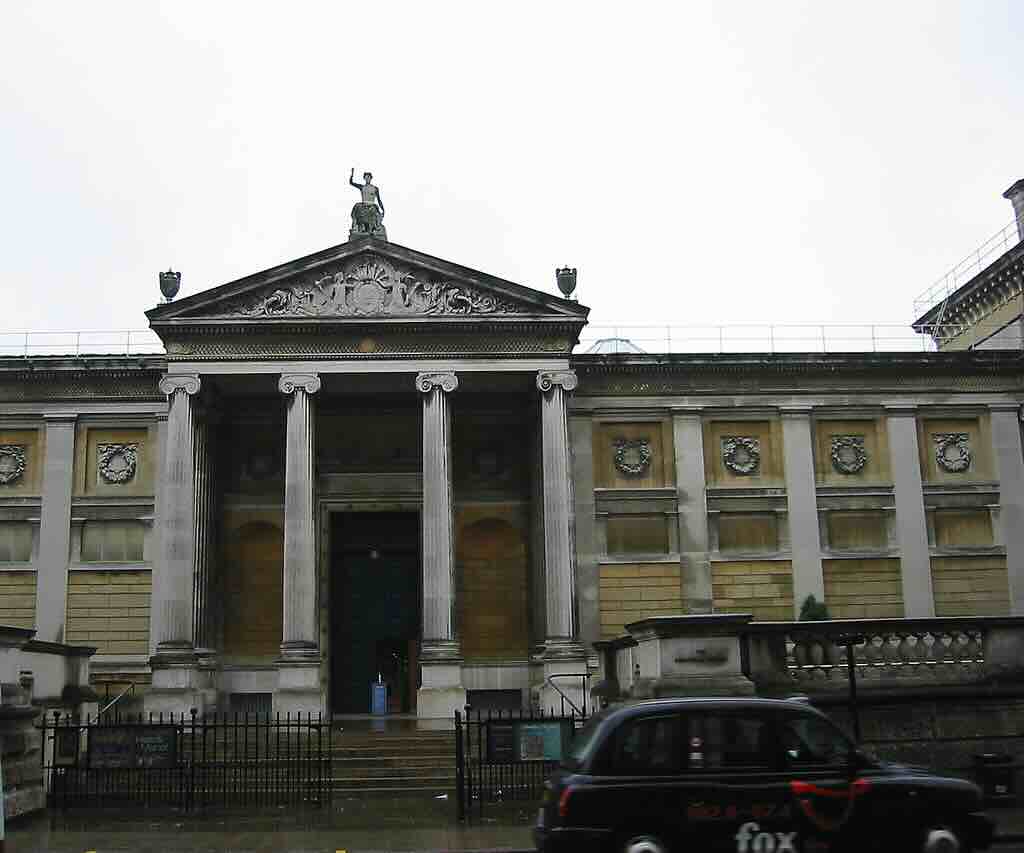} \\

    \raisebox{23pt}{R-Paris} &
    \includegraphics[width=52pt,height=52pt]{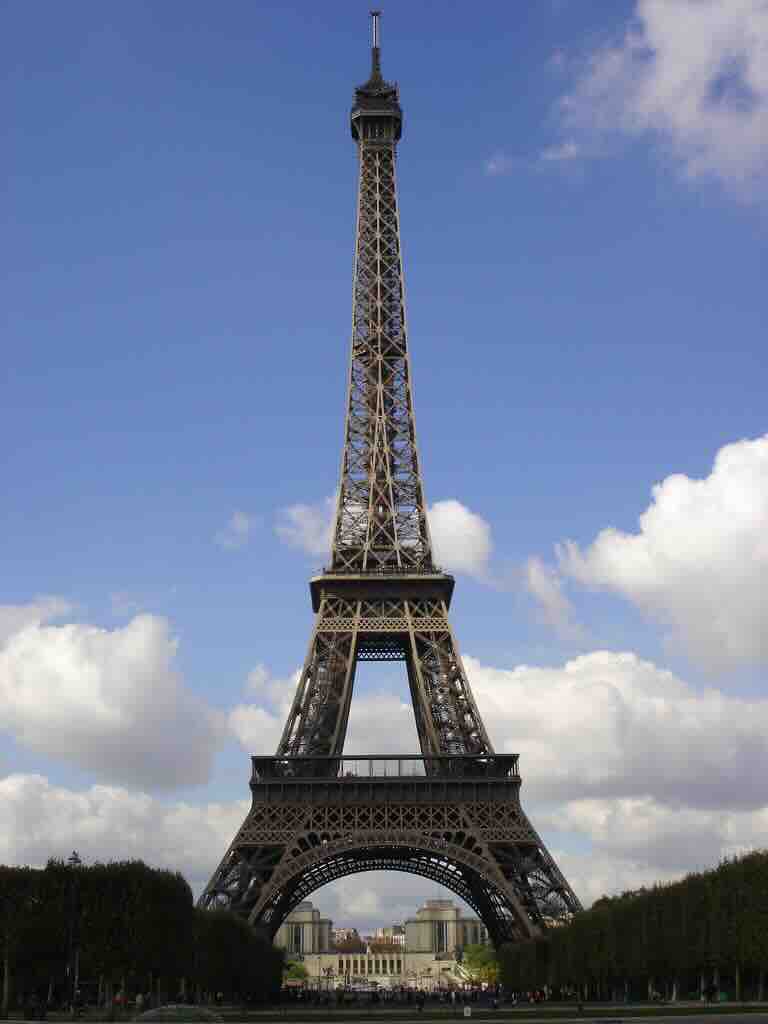} & 
    \includegraphics[width=52pt,height=52pt]{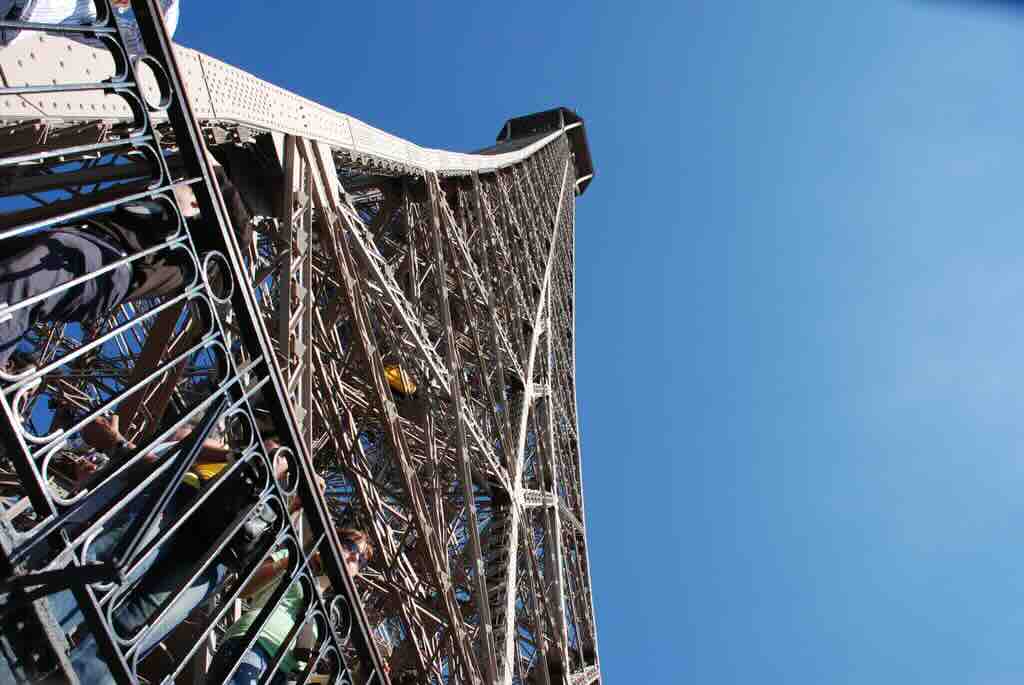} & 
    \includegraphics[width=52pt,height=52pt]{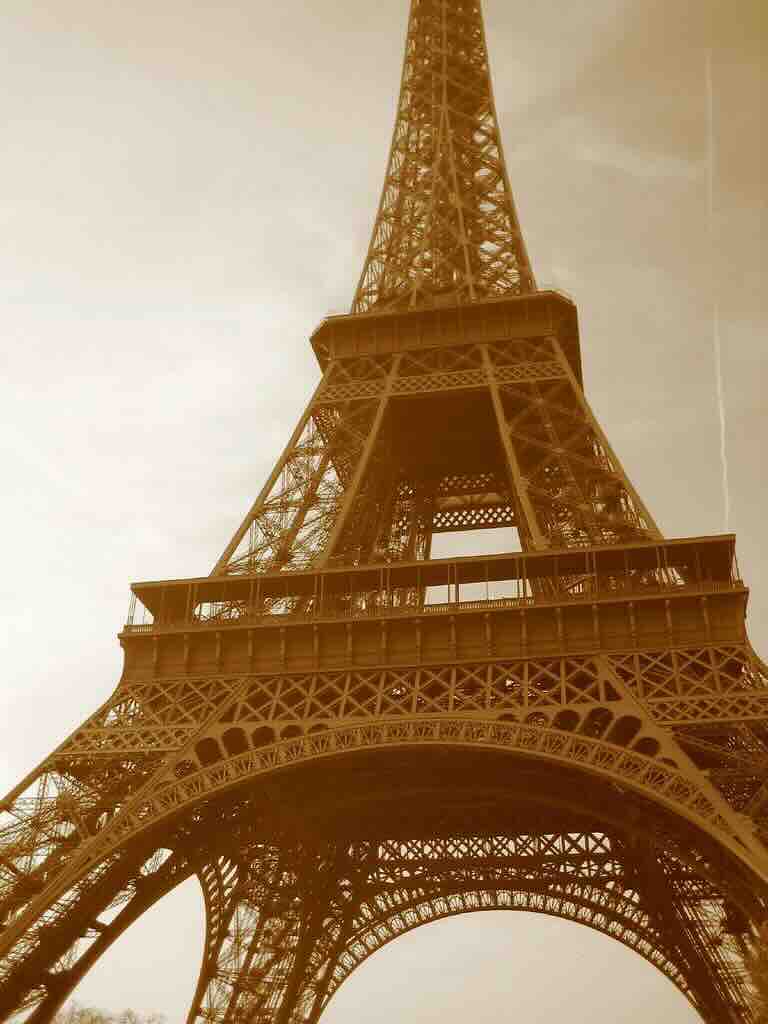} & 
    \includegraphics[width=52pt,height=52pt]{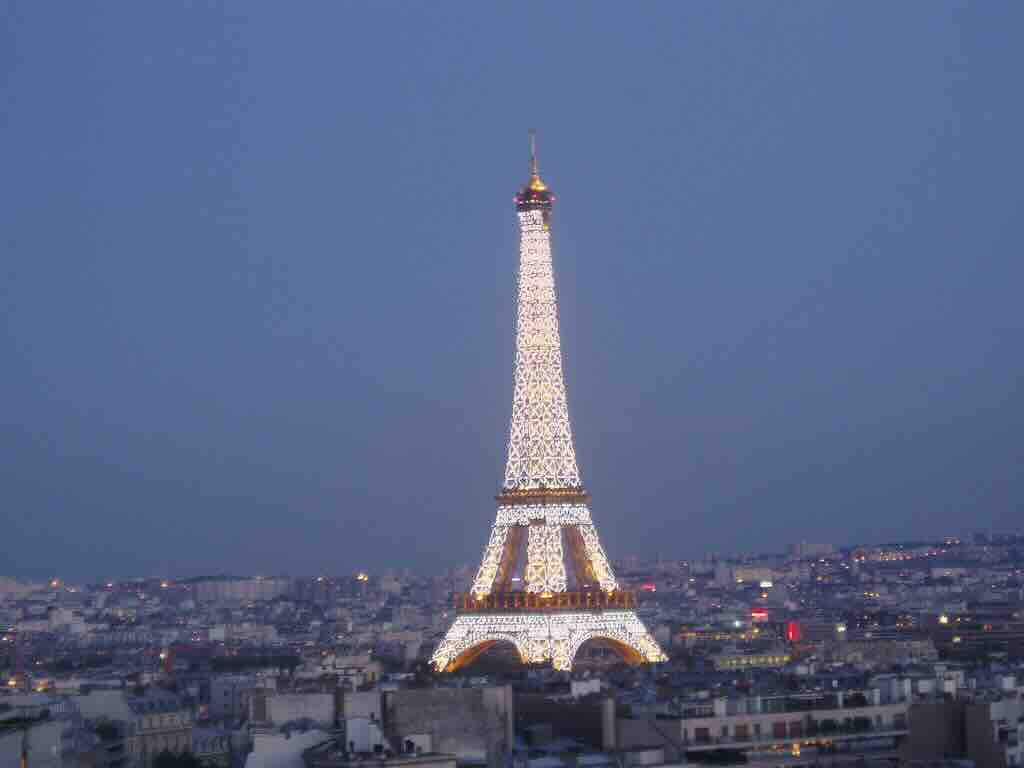} &
    \includegraphics[width=52pt,height=52pt]{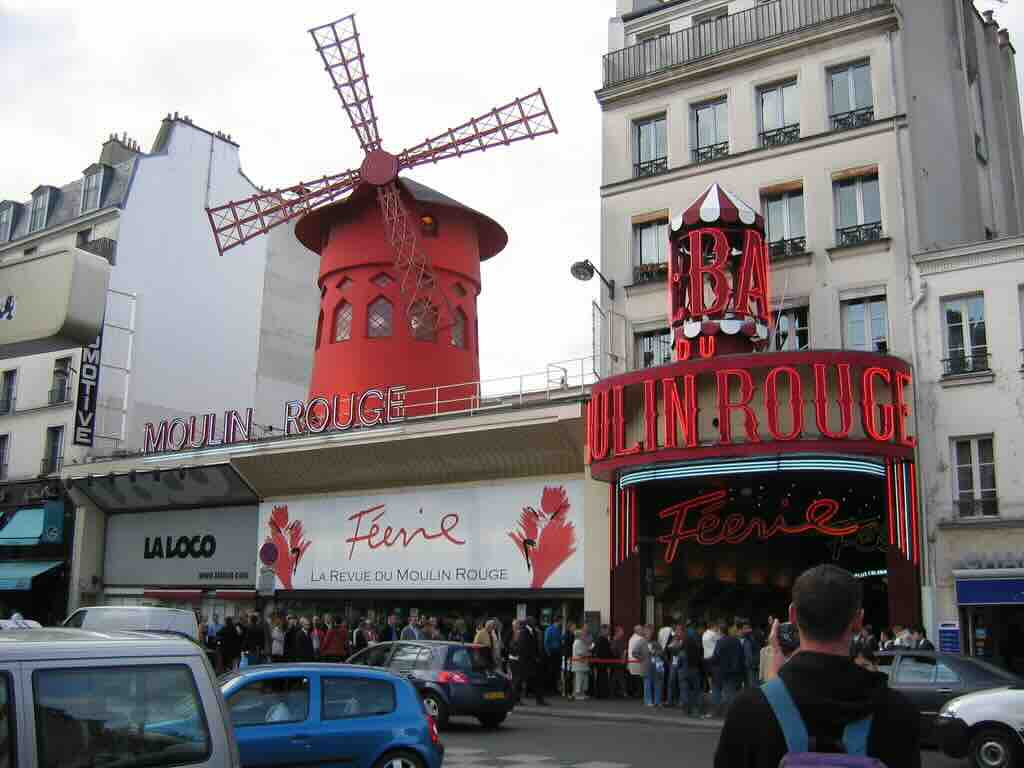} & 
    \includegraphics[width=52pt,height=52pt]{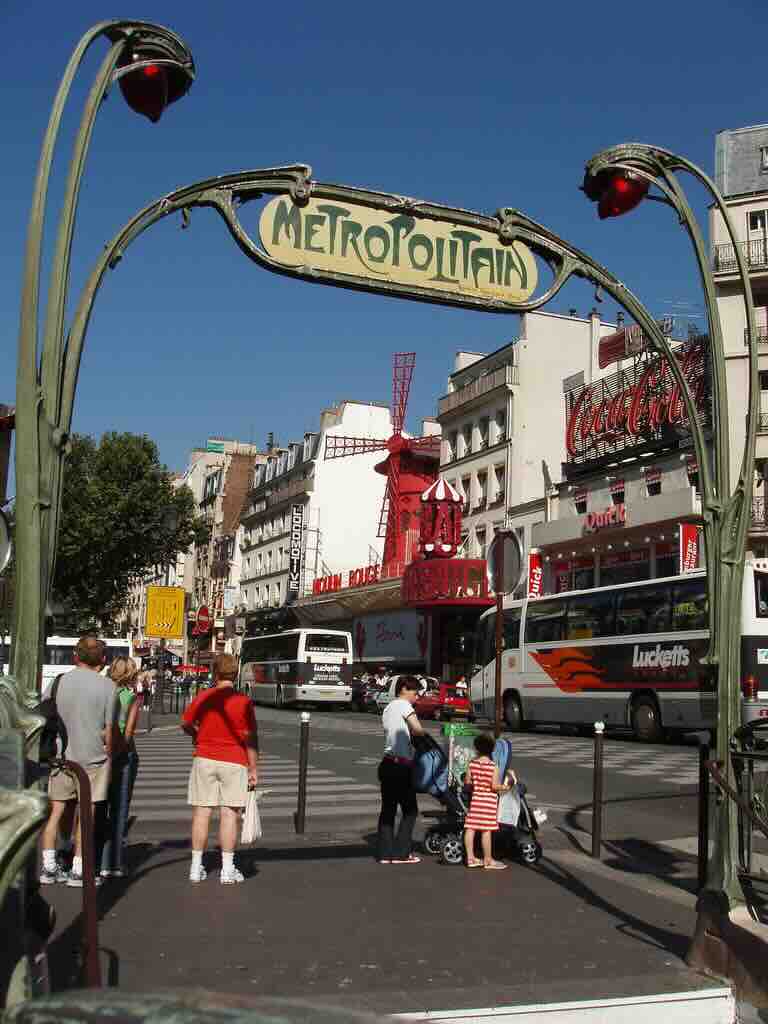} & 
    \includegraphics[width=52pt,height=52pt]{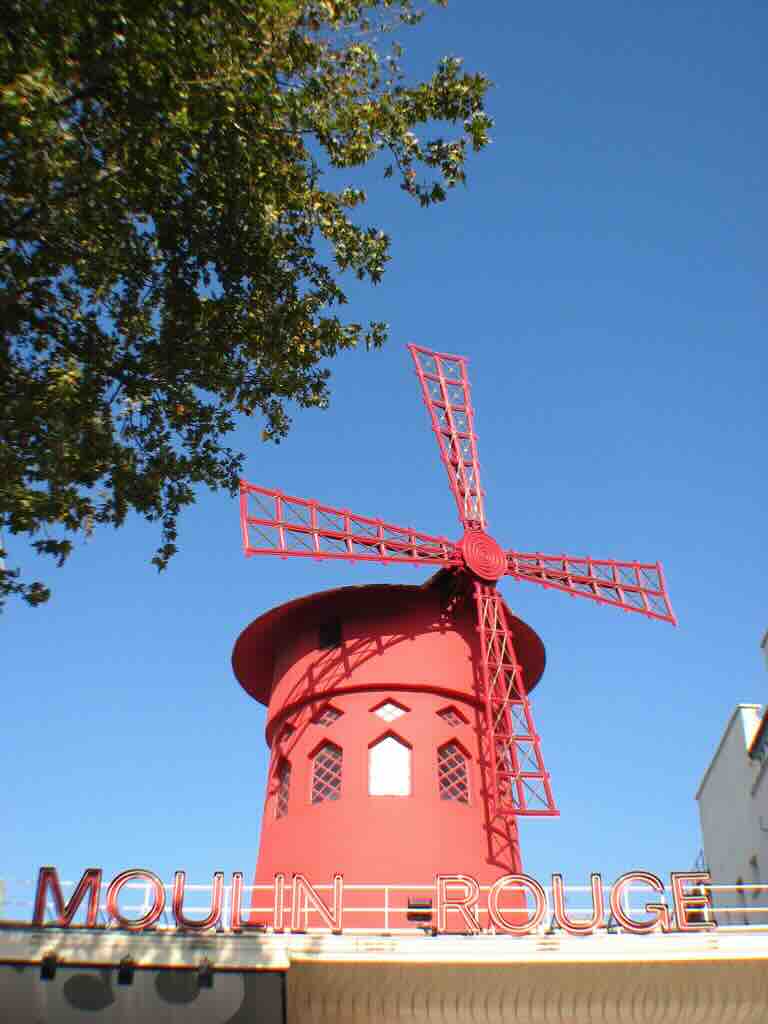} & 
    \includegraphics[width=52pt,height=52pt]{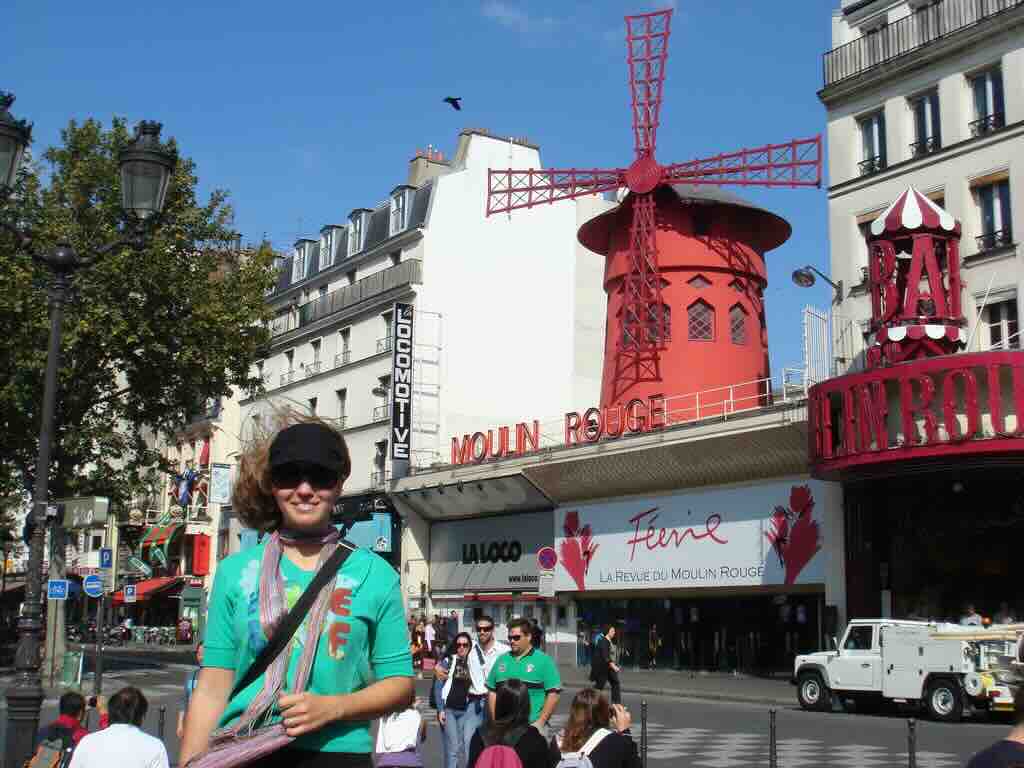} \\

    \raisebox{23pt}{GLDv2} &
    \includegraphics[width=52pt,height=52pt]{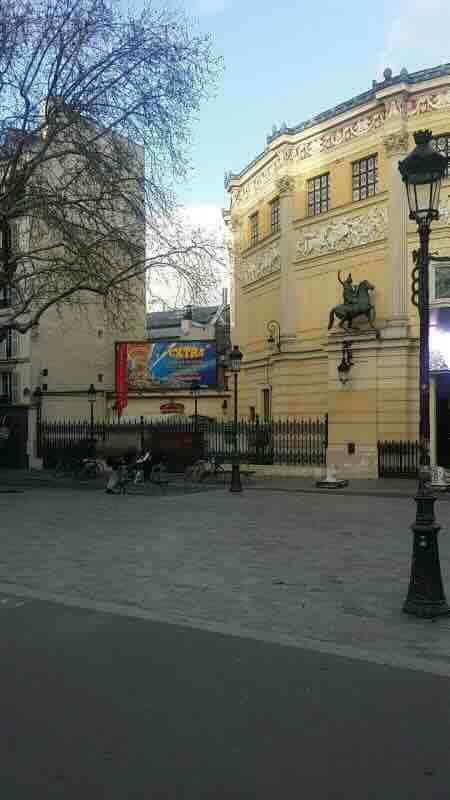} & 
    \includegraphics[width=52pt,height=52pt]{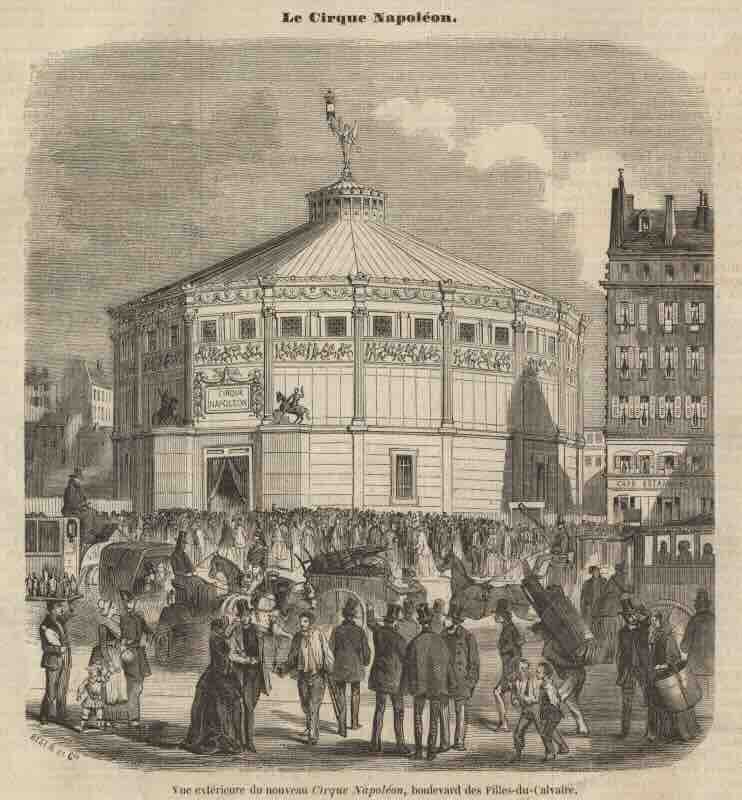} & 
    \includegraphics[width=52pt,height=52pt]{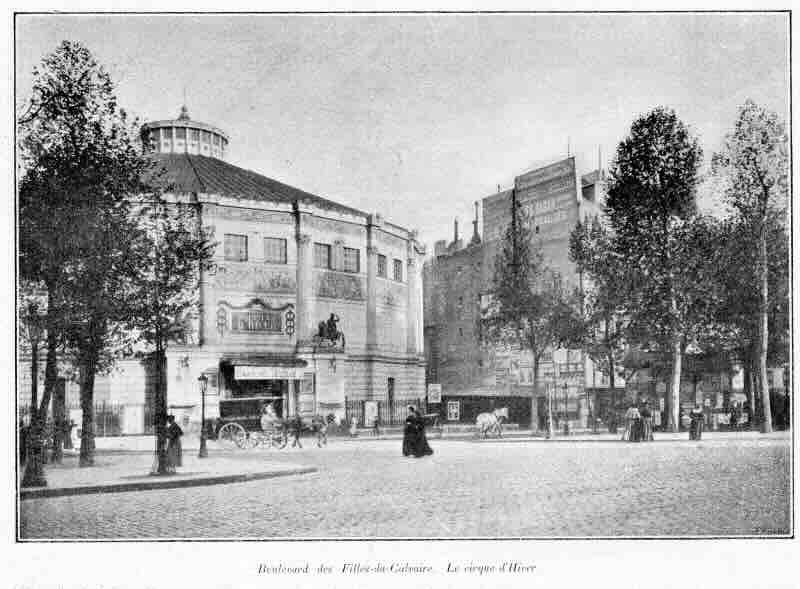} & 
    \includegraphics[width=52pt,height=52pt]{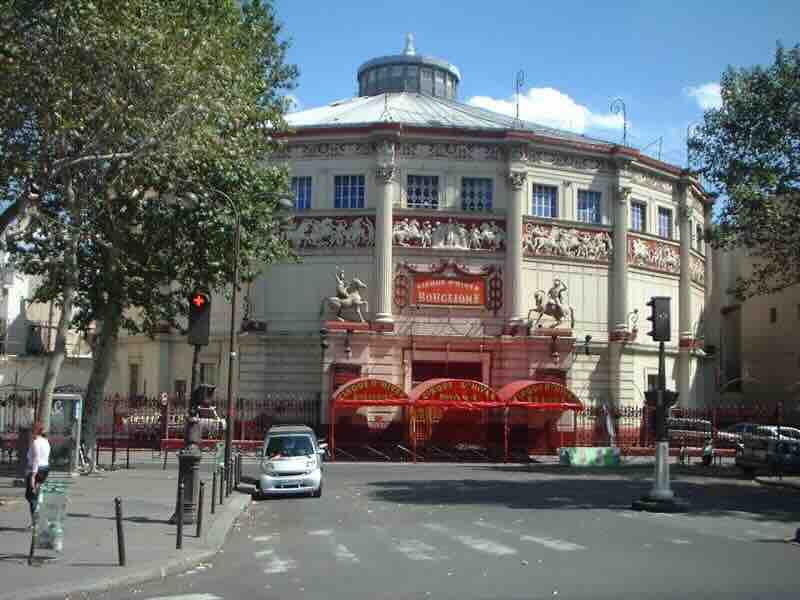} &
    \includegraphics[width=52pt,height=52pt]{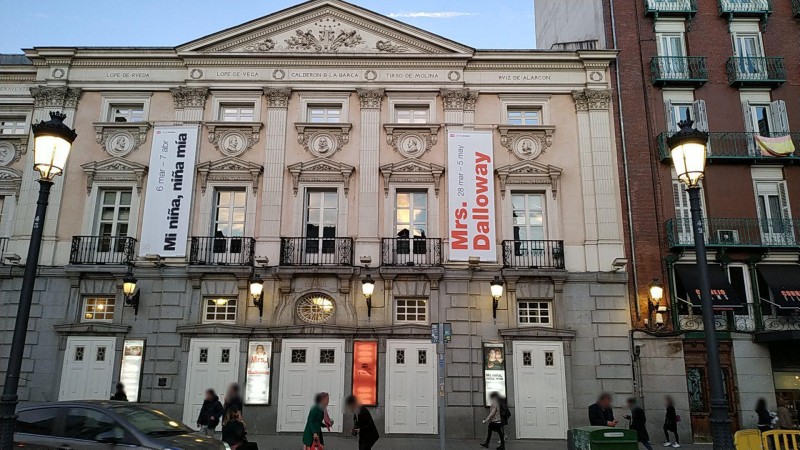} & 
    \includegraphics[width=52pt,height=52pt]{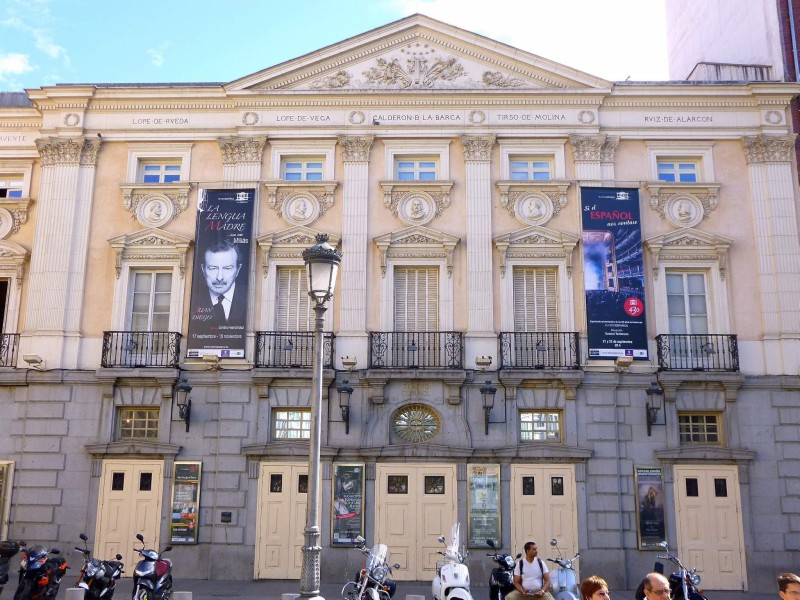} & 
    \includegraphics[width=52pt,height=52pt]{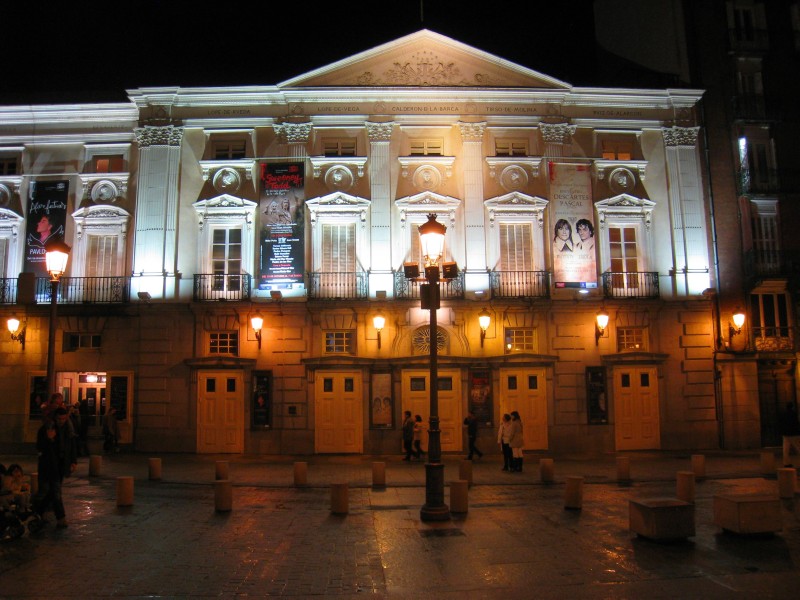} & 
    \includegraphics[width=52pt,height=52pt]{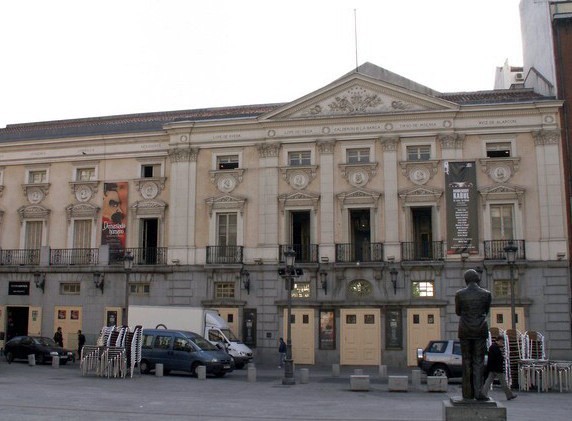} \\

    \raisebox{23pt}{SOP} &
    \includegraphics[width=52pt,height=52pt]{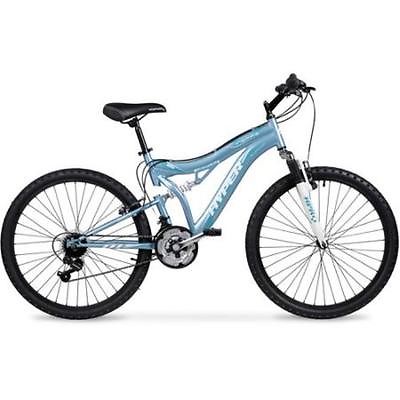} & 
    \includegraphics[width=52pt,height=52pt]{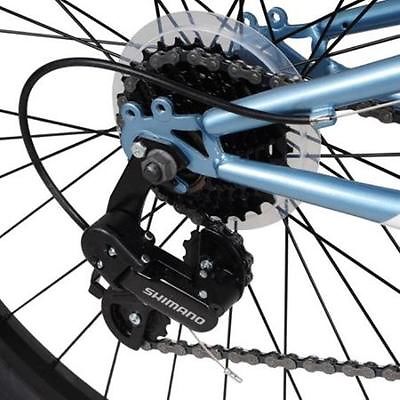} & 
    \includegraphics[width=52pt,height=52pt]{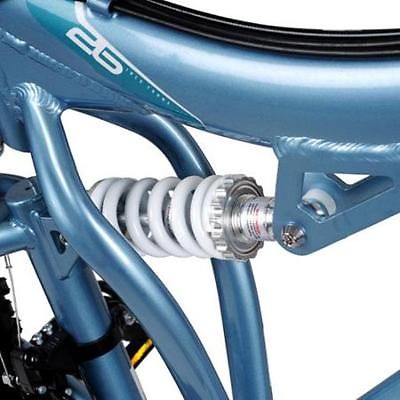} & 
    \includegraphics[width=52pt,height=52pt]{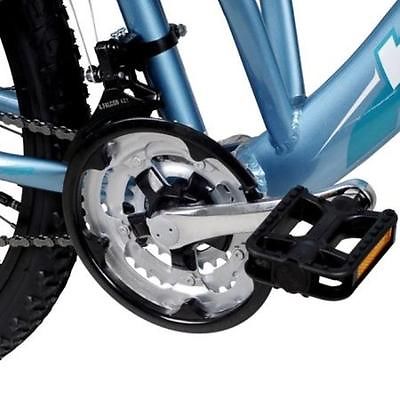} &
    \includegraphics[width=52pt,height=52pt]{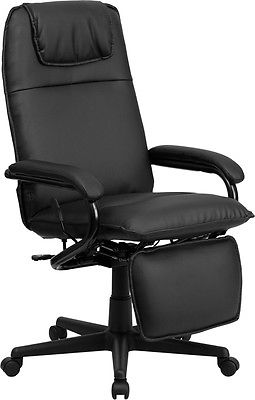} & 
    \includegraphics[width=52pt,height=52pt]{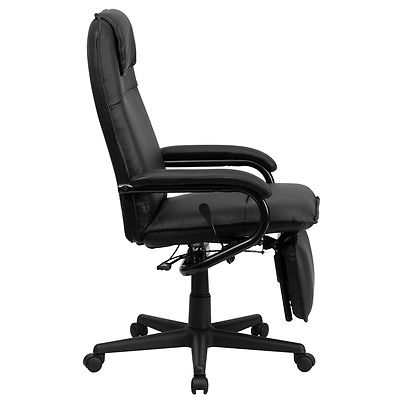} & 
    \includegraphics[width=52pt,height=52pt]{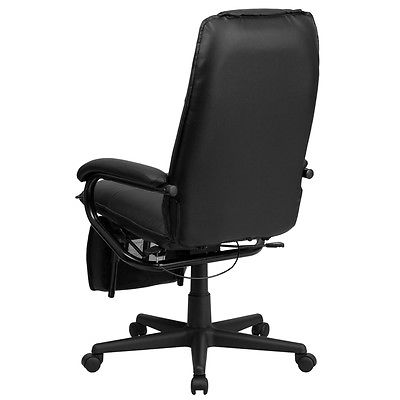} & 
    \includegraphics[width=52pt,height=52pt]{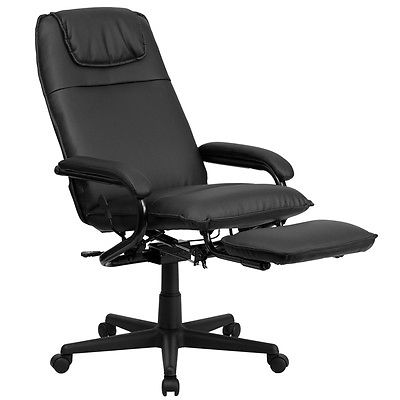} \\

    \raisebox{23pt}{INSTRE} &
    \includegraphics[width=52pt,height=52pt]{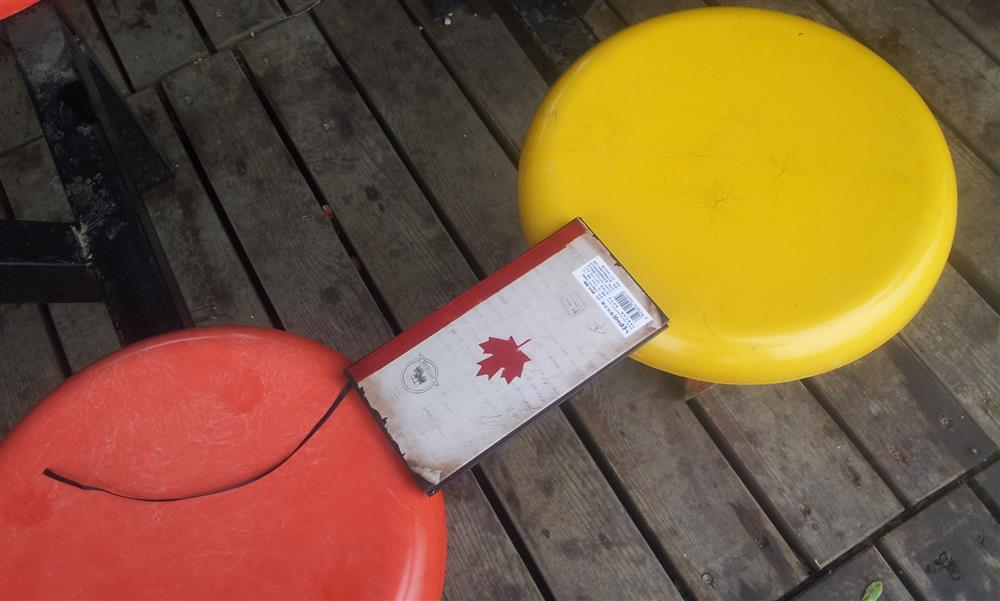} & 
    \includegraphics[width=52pt,height=52pt]{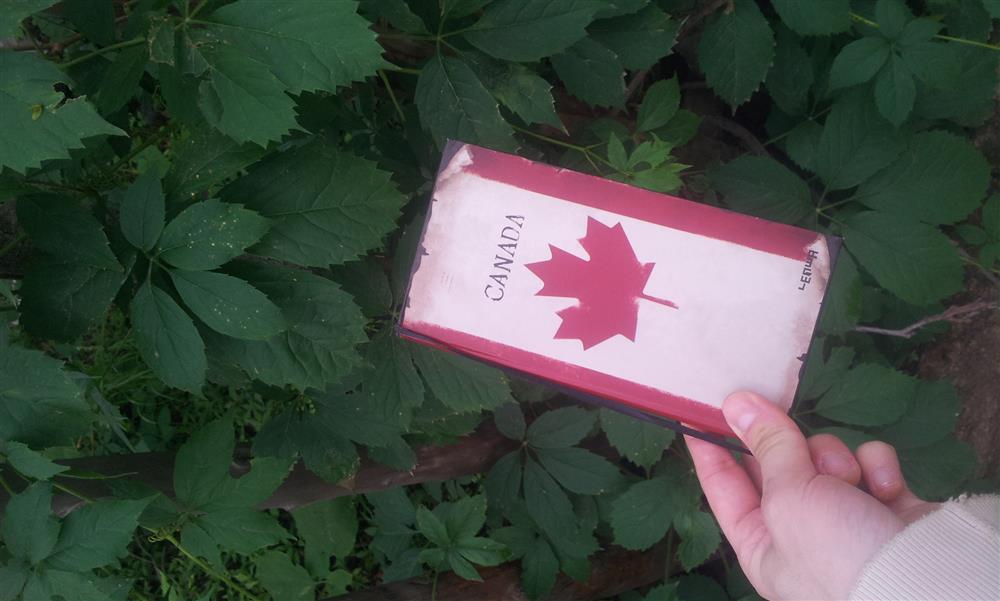} & 
    \includegraphics[width=52pt,height=52pt]{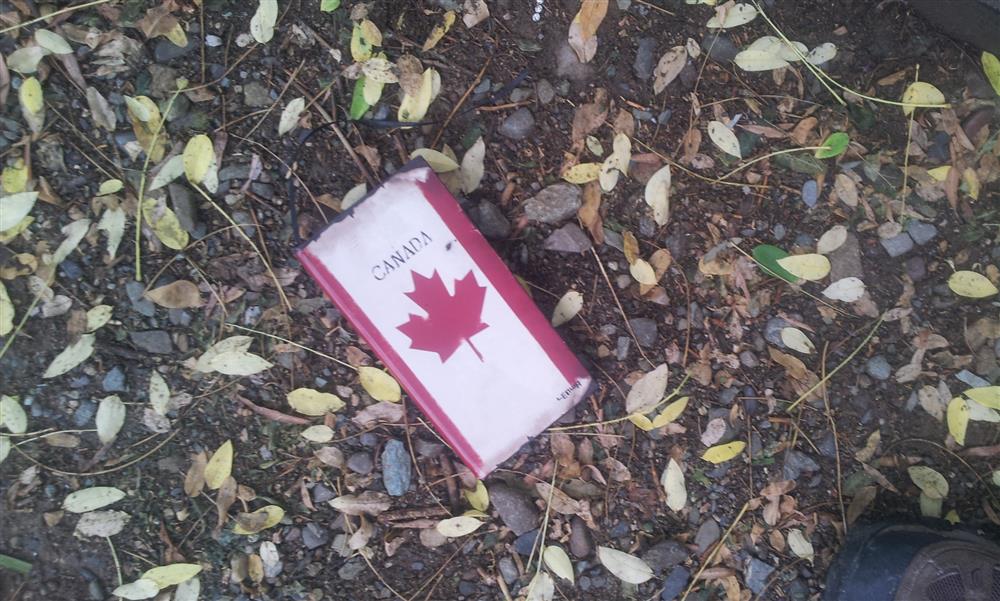} & 
    \includegraphics[width=52pt,height=52pt]{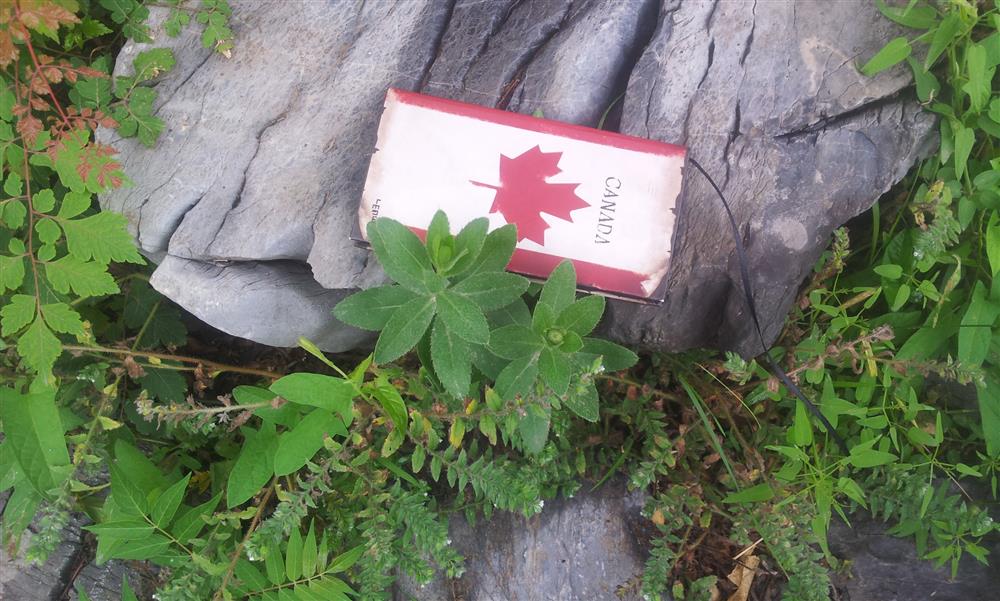} &
    \includegraphics[width=52pt,height=52pt]{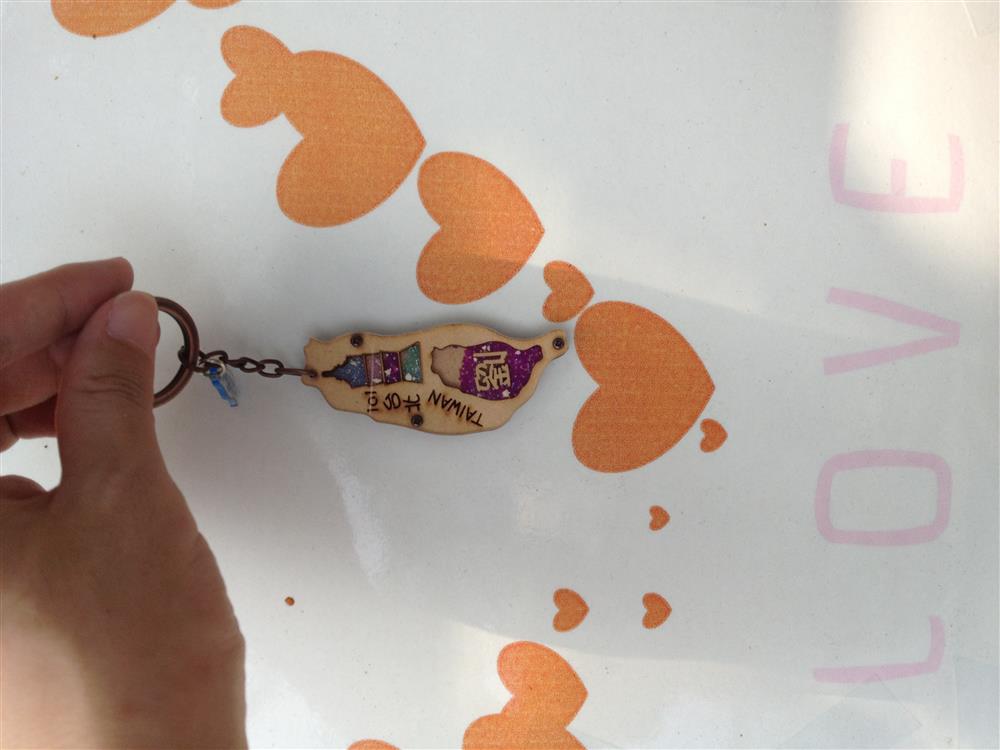} & 
    \includegraphics[width=52pt,height=52pt]{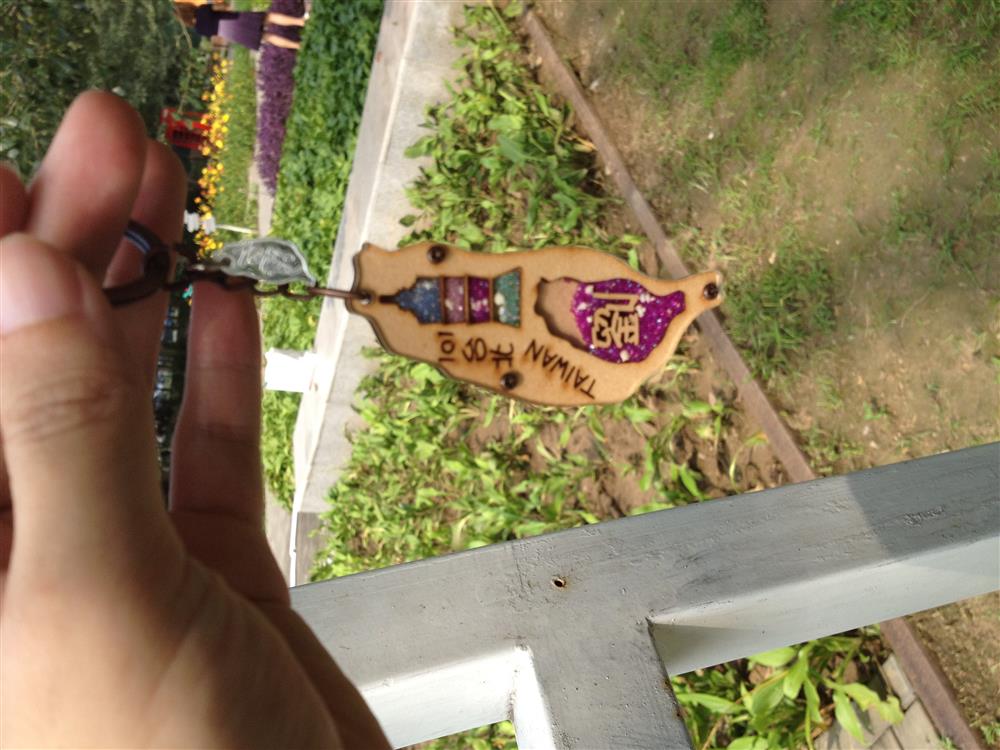} & 
    \includegraphics[width=52pt,height=52pt]{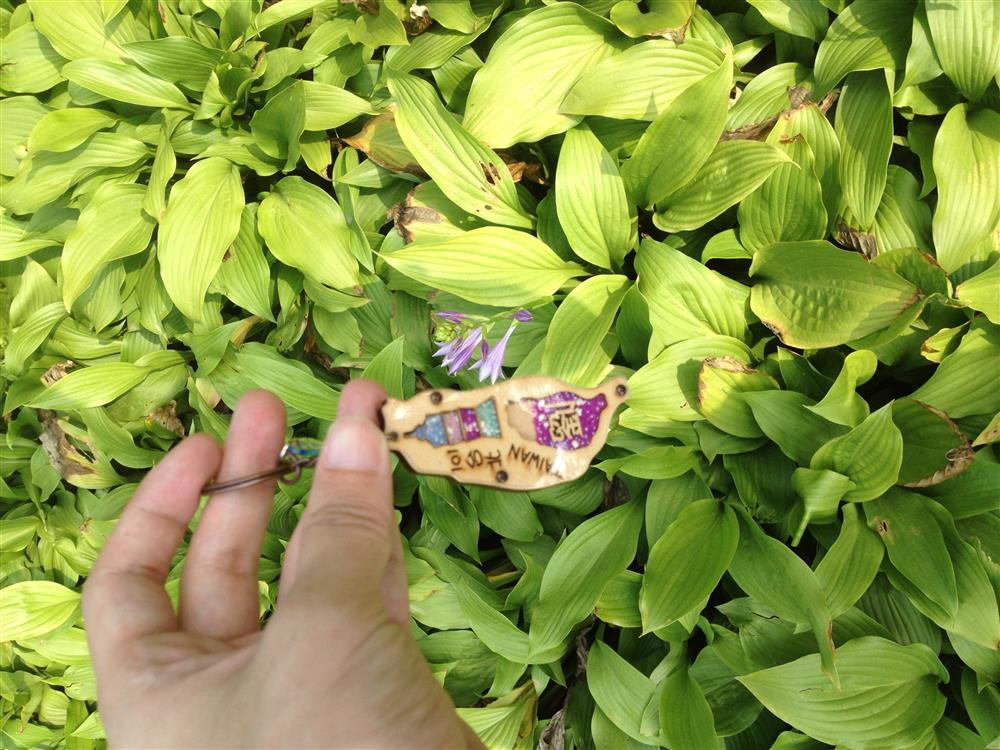} & 
    \includegraphics[width=52pt,height=52pt]{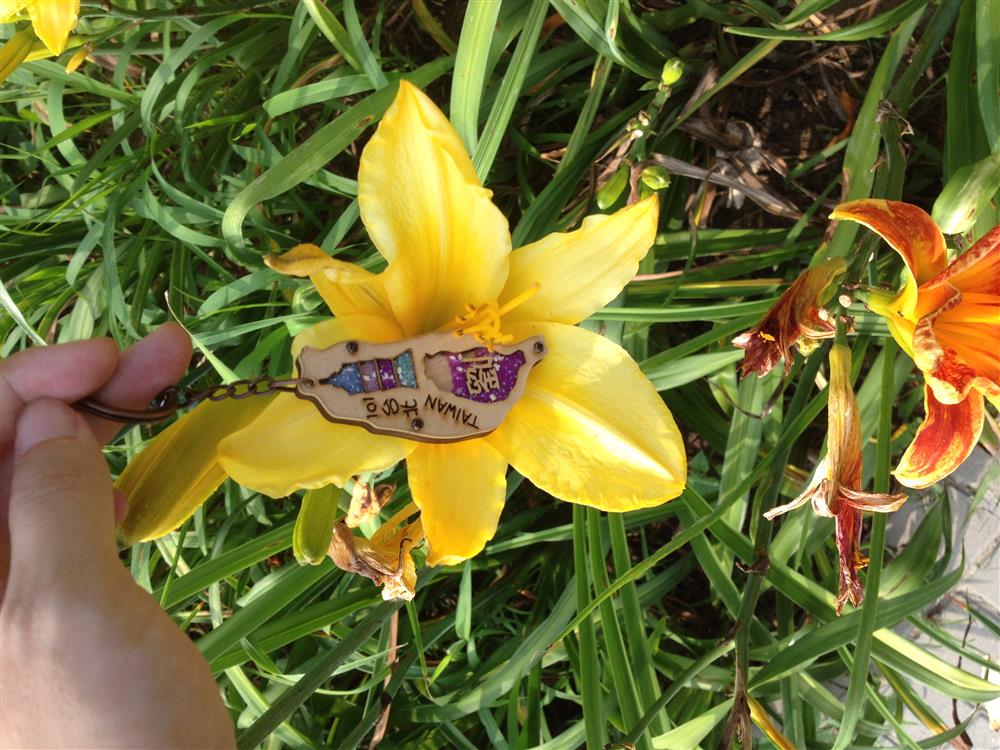} \\
    
    \raisebox{23pt}{mini-ILIAS} &
    \includegraphics[width=52pt,height=52pt]{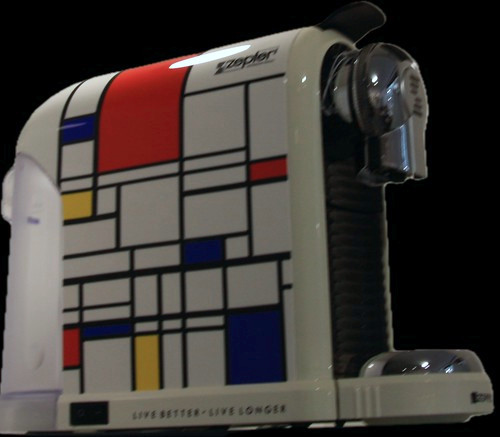} & 
    \includegraphics[width=52pt,height=52pt]{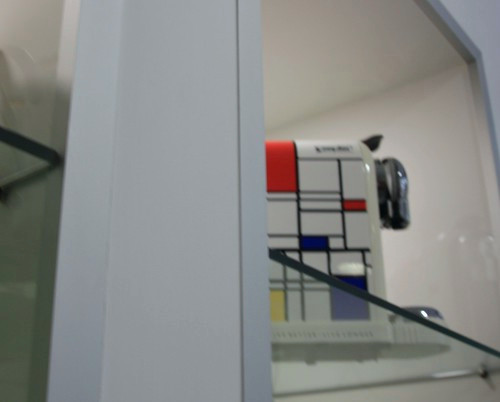} & 
    \includegraphics[width=52pt,height=52pt]{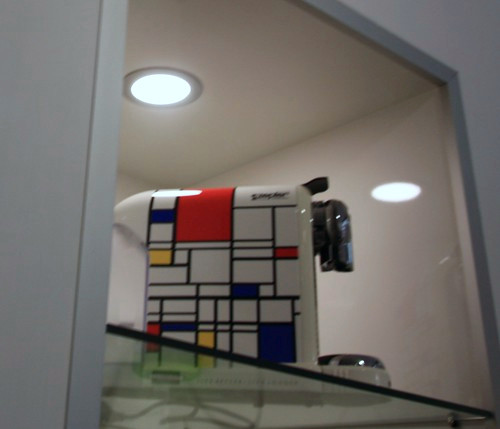} & 
    \includegraphics[width=52pt,height=52pt]{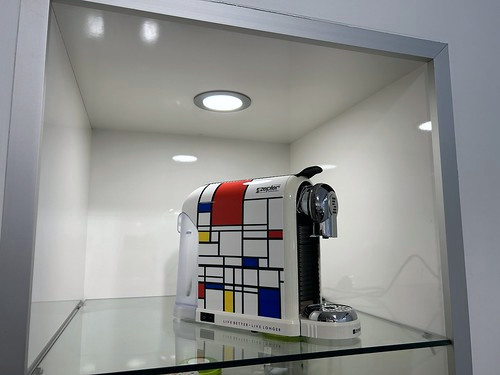} &
    \includegraphics[width=52pt,height=52pt]{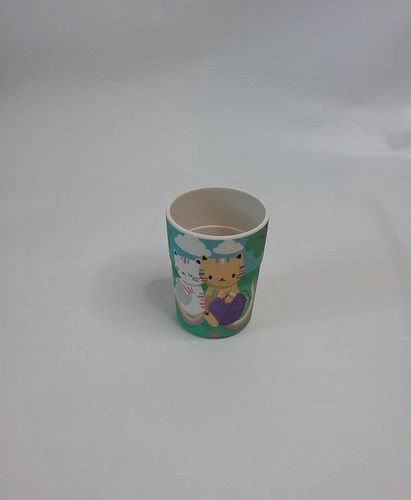} & 
    \includegraphics[width=52pt,height=52pt]{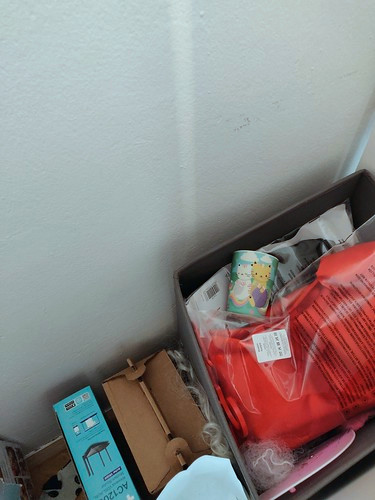} & 
    \includegraphics[width=52pt,height=52pt]{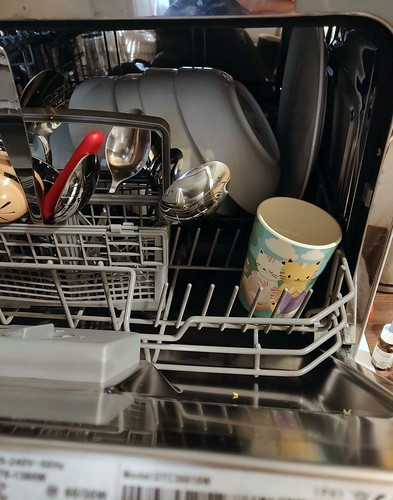} & 
    \includegraphics[width=52pt,height=52pt]{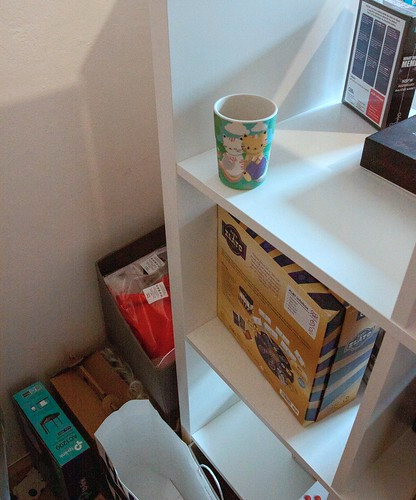} \\
\end{tabular}
\caption{Examples of queries (column 1) and positives (columns 2 $\sim$ 4) from all the test sets.
}
\label{fig:sup:eval_examples}
\end{figure*}

\section{Evaluation dataset examples}
\label{sec:sup:dataset_samples}

\cref{fig:sup:eval_examples} shows some examples of the queries and positives from the seven test sets.
The examples illustrate the diversity within the image domains and highlight the challenges posed by variations in viewpoint and background between queries and positive matches.
In detail, in the MET dataset, queries consist of photos taken by visitors inside the museum, often have complex backgrounds, whereas the database images, collected from the museum's website, typically have clean backgrounds. 
R-Oxford, R-Paris, and GLDv2 are landmark datasets where both the query and database images have complex backgrounds.
In the SOP dataset, retail product images are collected from e-commerce websites, with both query and database having clean backgrounds.
Lastly, INSTRE and mini-ILIAS include multi-domain objects. In INSTRE, both query and database have diverse backgrounds, while in mini-ILIAS, queries have clean backgrounds.

\end{document}